\documentclass{article}

\newif\ificlrstyle \iclrstylefalse
\newif\ifanon      \anonfalse

\usepackage[utf8]{inputenc}
\usepackage[T1]{fontenc}
\usepackage{caption}
\usepackage[dvipsnames]{xcolor}
\definecolor{linkblue}{RGB}{0,0,139}
\usepackage{xltabular}
\usepackage{booktabs}
\usepackage{array}
\usepackage{url}
\usepackage{textcomp}
\usepackage{setspace}
\usepackage{pdfpages}
\usepackage{amsmath}
\usepackage{mathtools}
\usepackage{amssymb}
\usepackage{algorithm}
\usepackage{algpseudocode}
\usepackage{graphicx}
\usepackage{float}
\usepackage{afterpage}
\usepackage{tcolorbox}
\usepackage{fvextra}
\tcbuselibrary{breakable,skins}

\definecolor{promptsystem}{RGB}{84,106,138}
\definecolor{promptuser}{RGB}{58,122,92}
\definecolor{promptmodel}{RGB}{176,112,38}

\DefineVerbatimEnvironment{PromptText}{Verbatim}{%
  fontsize=\scriptsize,
  fontfamily=zi4,
  breaklines=true,
  breakanywhere=true,
  breakindent=1.2em,
  breaksymbolleft={\tiny\ensuremath{\hookrightarrow}},
  breaksymbolsepleft=3pt}

\newtcolorbox{promptbox}[2][promptsystem]{%
  enhanced, breakable,
  colback=#1!4, colframe=#1!60, colbacktitle=#1!16, coltitle=black,
  boxrule=0.4pt, arc=1.5pt,
  fonttitle=\footnotesize\bfseries, title={#2},
  left=4pt, right=4pt, top=2pt, bottom=2pt, toptitle=1.5pt, bottomtitle=1.5pt,
  before skip=6pt plus 2pt, after skip=6pt plus 2pt}

\newenvironment{prosecompletion}{\small\setlength{\parskip}{3pt}}{}

\tcbuselibrary{skins}

\usepackage{enumitem}
\AtBeginDocument{%
  \setlist[itemize]{leftmargin=1.25em}%
  \setlist[enumerate]{leftmargin=1.75em}%
}

\usepackage{placeins}
\renewcommand{\topfraction}{0.92}

\renewcommand{\textfraction}{0.06}
\renewcommand{\floatpagefraction}{0.7}

\newcommand{\ourbenchmark}{EDB-Bio}

\newcommand{\linkpending}[1]{{\color{red}[LINK: #1]}} 

\newcommand{\modelname}{Ibis0}

\newcommand{\hfdataset}{EdisonScientific/EDB-Bio}

\newcommand{\urldataset}{huggingface.co/datasets/\hfdataset}
\newcommand{\urltestset}{huggingface.co/datasets/\hfdataset/tree/v1.0.0/testsets}
\newcommand{\urlweights}{huggingface.co/collections/EdisonScientific/ibis0}
\newcommand{\urlcode}{github.com/EdisonScientific/ibis0}
\newcommand{\urllogs}{huggingface.co/datasets/EdisonScientific/ibis0-paper-artifacts}

\DeclareRobustCommand{\artifacthref}[2]{%
  \begingroup
  \edef\artifactaddress{#1}%
  \ifx\artifactaddress\empty
    \linkpending{#2}%
  \else
    \edef\artifacttarget{https://\artifactaddress}%
    \expandafter\href\expandafter{\artifacttarget}{#2}%
  \fi
  \endgroup}

\DeclareRobustCommand{\artifacturl}[2]{%
  \begingroup
  \edef\artifactaddress{#1}%
  \ifx\artifactaddress\empty
    \linkpending{#2}%
  \else
    \edef\artifacttarget{https://\artifactaddress}%
    \expandafter\href\expandafter{\artifacttarget}%
      {\expandafter\nolinkurl\expandafter{\artifactaddress}}%
  \fi
  \endgroup}

\usepackage{lmodern}

\usepackage{sectsty}

\definecolor{edisongray}{HTML}{f0f0f0}
\usepackage[left=0.9in, right=0.9in, top=0.85in, bottom=0.9in]{geometry}

\newenvironment{headerbox}{
  \begin{tcolorbox}[
    colback=edisongray,
    colframe=edisongray,
    frame hidden,
    arc=4mm,
    boxrule=0pt,
    boxsep=0pt,
    left=15pt, right=15pt, top=10pt, bottom=15pt,
    nobeforeafter
  ]
}{
  \end{tcolorbox}
}

\usepackage[numbers,sort&compress]{natbib}

\usepackage{hyperref}
\hypersetup{
  colorlinks=true,
  linkcolor=linkblue,
  citecolor=linkblue,
  urlcolor=linkblue
}
\usepackage[nameinlink]{cleveref}

\crefname{section}{\S}{\S\S}
\Crefname{section}{Section}{Sections}
\crefname{subsection}{\S}{\S\S}
\Crefname{subsection}{Section}{Sections}
\crefname{subsubsection}{\S}{\S\S}
\Crefname{subsubsection}{Section}{Sections}
\crefformat{section}{\S#2#1#3}
\crefformat{subsection}{\S#2#1#3}
\crefformat{subsubsection}{\S#2#1#3}
\crefrangeformat{section}{\S\S#3#1#4--#5#2#6}
\crefrangeformat{subsection}{\S\S#3#1#4--#5#2#6}
\crefmultiformat{section}{\S\S#2#1#3}{ and #2#1#3}{, #2#1#3}{, and #2#1#3}
\crefmultiformat{subsection}{\S\S#2#1#3}{ and #2#1#3}{, #2#1#3}{, and #2#1#3}

\crefname{appendix}{App.}{Apps.}
\Crefname{appendix}{App.}{Apps.}
\crefname{subappendix}{App.}{Apps.}
\Crefname{subappendix}{App.}{Apps.}
\crefname{subsubappendix}{App.}{Apps.}
\Crefname{subsubappendix}{App.}{Apps.}

\crefname{figure}{Fig.}{Figs.}
\Crefname{figure}{Fig.}{Figs.}
\crefname{table}{Table}{Tables}
\Crefname{table}{Table}{Tables}
\crefname{equation}{Eq.}{Eqs.}
\Crefname{equation}{Eq.}{Eqs.}

\sectionfont{\sffamily}

\newcommand{\titlebreak}{\\[7pt]}
\newcommand{\paperbibliography}{%
  \bibliographystyle{unsrtnat}%
  \bibliography{references}%
}

\ifanon

  \newcommand{\repohandle}[1]{{\color{black!55}[repository withheld for anonymous review]}}

  \hypersetup{pdfauthor={}, pdftitle={}, pdfsubject={}, pdfkeywords={},
              pdfcreator={}, pdfproducer={}}
\else
  
  \DeclareUrlCommand\repohandlepath{\urlstyle{tt}%
    \mathchardef\UrlBreakPenalty=5000\relax}
  \DeclareRobustCommand{\repohandle}[1]{%
    \begingroup\edef\repohandlename{#1}%
    \expandafter\repohandlepath\expandafter{\repohandlename}\endgroup}
\fi

\begin{document}

%

\newcommand{\papertitle}{Do Language Models Know What Compelling\titlebreak Scientific Evidence Looks Like?}
\newcommand{\papersubtitle}{Calibrating Hypothesis Judgment Across Multiple Biological Domains}

\newcommand{\abstracttext}{%
Expert scientists excel at weighing empirical findings from multiple sources that vary in both statistical reliability and epistemic relevance. To what extent are language models, which are increasingly driving autonomous discovery systems, capable of such \textit{evidence deliberation}? We introduce \textbf{\ourbenchmark}, a scientific reasoning benchmark curated from real-world data in three biological domains: drug-perturbation transcriptomics, CRISPR screens, and human genome-wide association studies (GWAS). Each of \ourbenchmark's 96k episodes presents a sequence of experimental results from published studies and asks the model to predict the probability that a target hypothesis is true. Out-of-the-box, open-weight and frontier models alike demonstrate systematic underconfidence and attain Brier scores comparable to a trivial base-rate forecaster. Post-training against strictly proper scoring rules significantly improves both calibration and discrimination. This calibration pays off downstream: in a sequential experiment selection task, our Ibis0 model is $2.5\times$ more efficient than the best frontier model. Throughout, quantifying evidence weights as \textit{Bayes factors} reveals how models update their beliefs over time, whether they agree with expert biologists on relevance, and where they exhibit inconsistencies and biases. Our work extends calibration training and uncertainty analysis to a realistic scientific forecasting setting with applications to human cancer biology.
We release all \artifacthref{\urlcode}{code}, \artifacthref{\urldataset}{data}, and \artifacthref{\urlweights}{trained models}.
}

\ificlrstyle
\title{\papertitle}
\author{%
Conor Igoe$^{1}$, Gabriel Grand$^{1,2}$, Zane Koch$^{1}$, \\
Albert Bou$^{1}$, Eli Moss$^{1}$, Michaela M. Hinks$^{1}$, 
Samuel G. Rodriques$^{1}$, Andrew White$^{1}$, 
Siddharth Narayanan$^{1}$ \\
$^{1}$Edison Scientific Inc., San Francisco, CA, USA \\
$^{2}$MIT CSAIL, Cambridge, MA, USA \\
\texttt{cigoe@edisonscientific.com}
}
\maketitle
\begin{abstract}
\abstracttext
\end{abstract}
\else

\begin{center}
\begin{headerbox}
  \vspace*{1em}
    {\sffamily \textbf{\huge \papertitle}}

    \vspace{0.6em}
    {\sffamily \large \papersubtitle}

    \vspace{1em}
    {Conor Igoe$^{1}$},
    {Gabriel Grand$^{1,2}$},
    {Zane Koch$^{1}$},
    {Albert Bou$^{1}$},
    {Eli Moss$^{1}$},
    {Michaela M. Hinks$^{1}$},
    {Samuel G. Rodriques$^{1}$}, 
    {Andrew White$^{1}$},
    {Siddharth Narayanan$^{1}$}

    \vspace{1em}

    \sffamily \textbf{\large Abstract}

    \vspace{0.2em}

    \abstracttext
    \\

    \vspace{1em}

    {\footnotesize
    \raggedright
    $^{1}$Edison Scientific Inc., San Francisco, CA, USA \quad $^{2}$MIT CSAIL, Cambridge, MA, USA\\
    Correspondence to \texttt{\{cigoe,sid\}$@$edisonscientific.com}\\
    }
    \vspace{-1.5cm}\hfill\raisebox{0cm}{\includegraphics[width=1.3cm]{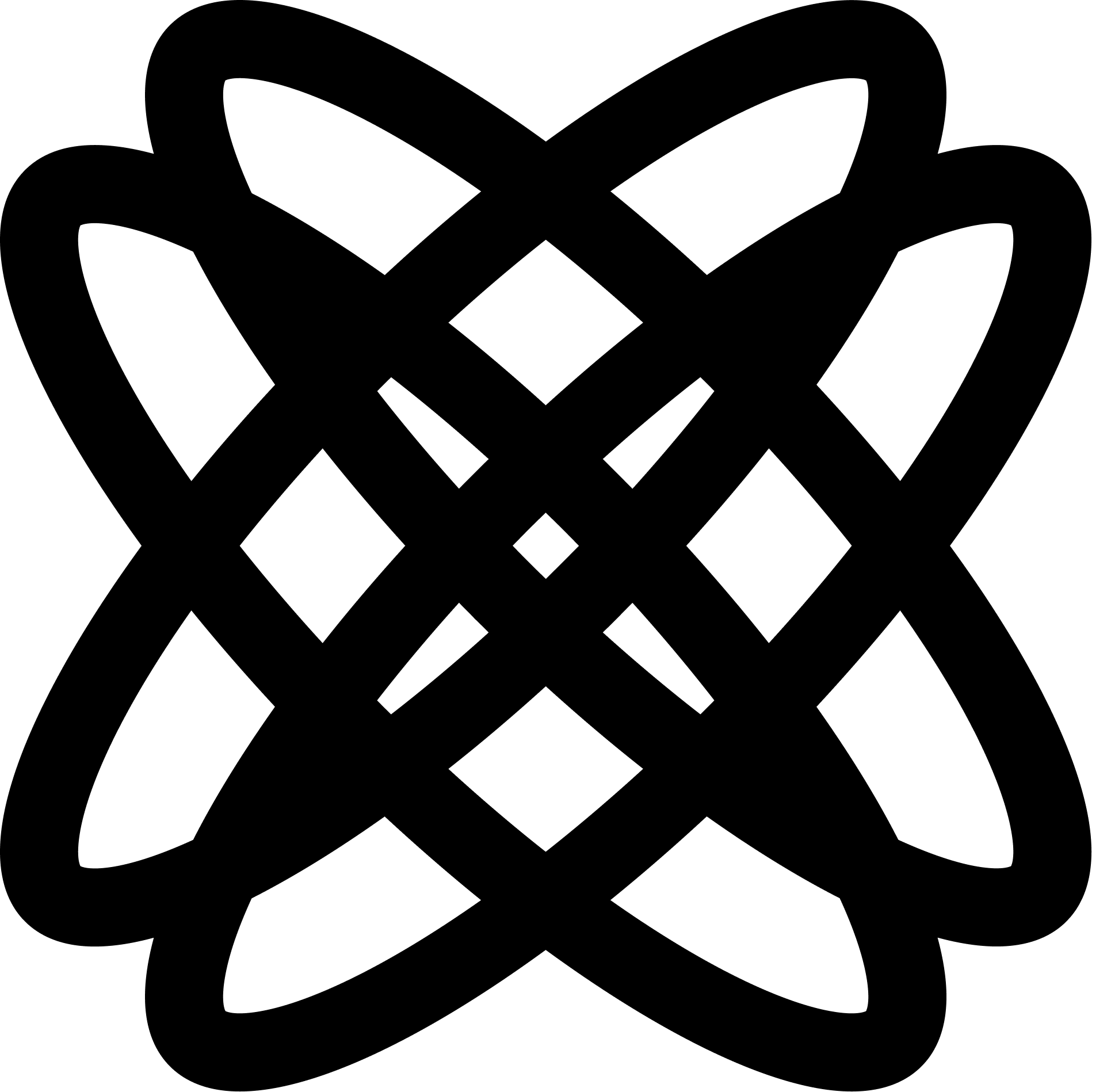}}\hspace{0cm}

\end{headerbox}
\end{center}
\fi


\normalsize

\addtocontents{toc}{\protect\setcounter{tocdepth}{-1}}

\section{Introduction}
\label{sec:intro}

Experienced scientists are good at weighing evidence. Shown the result of an experiment, they can often judge whether it is compelling and likely to generalize; or whether it is weak, confounded, or irrelevant to the question being studied. Yet today, research is increasingly driven by language models~\citep{boiko2023coscientist,lu2024aiscientist,yamada2025aiscientistv2,kosmos,gottweis2026coscientist}. How reliably can these models exercise epistemic judgment? This question is particularly important for autonomous workflows in which a model's output feeds decisions such as which hypothesis to pursue, what experiments to run, or how confidently to report conclusions.

Prior calibration work has concentrated on general-purpose domains such as mathematics and commonsense reasoning rather than on empirical scientific evidence \citep{damani2026beyond, puri2026reaching, xu2024sayself}. Agentic research systems weigh large and heterogeneous bodies of evidence of variable quality, which can be very different from observations in other domains. To this end, we introduce EvidenceDeliberationBench-Biology, \ourbenchmark{}, a benchmark of real biological hypotheses, each paired with a controlled, relevance-graded pool of experimental evidence. The model is tasked to predict whether the hypothesis is true or not, conditioned on progressively more evidence from the pool. Because the evidence is controlled, we can measure both whether a model is right and how its belief tracks the strength and relevance of the evidence.

On \ourbenchmark{}, language models can often differentiate more likely hypotheses from less likely ones, yet in all three domains their stated probabilities are poorly calibrated. Training an open model against a strictly proper scoring rule closes the calibration gap and matches or exceeds frontier discrimination. A model with reliable probabilistic judgments can also \textit{choose} the experiment whose result would most change its belief. In a sequential Bayesian experimental design~\citep{rainforth2024modern} setting, our trained \modelname{} model extracts about $2.5\times$ more information per experiment than leading frontier models.

\begin{figure}[t]\centering
\includegraphics[width=\textwidth]{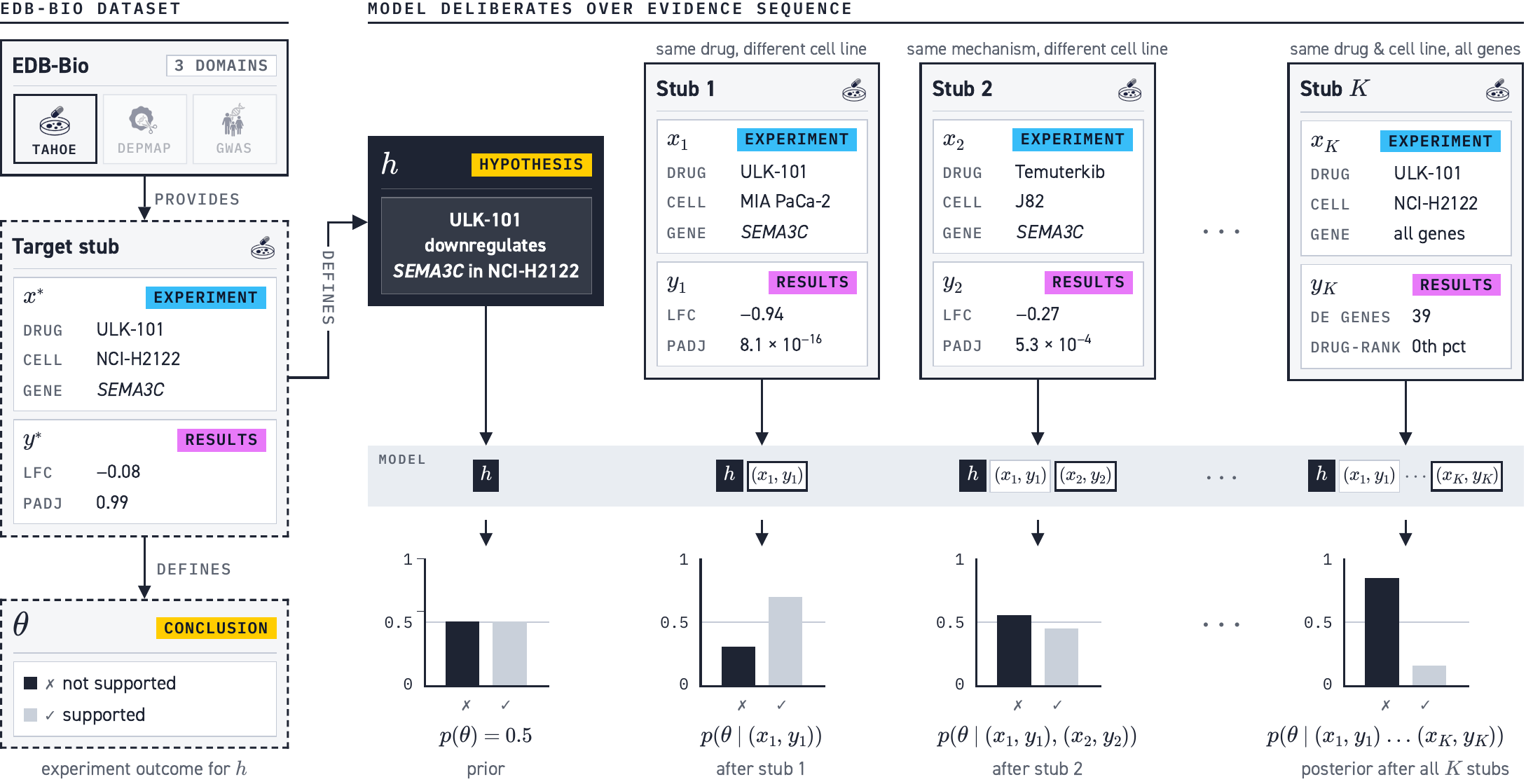}
\caption{\textbf{Evidence deliberation task.} \ourbenchmark{} asks models to reason about a scientific hypothesis $h$ given a sequence of empirical evidence. Each \textit{evidence stub} is derived from published data in one of three biological domains (\cref{tab:domains-principles}). In the example here, each experiment $x$ is an RNA-Seq study from Tahoe-100M \citep{tahoe100m} in which a drug perturbs a gene in a particular cell line. The result $y$ reports a DESeq2 assay readout, including an \textit{effect} (log$_2$ fold-change) and its \textit{significance} (adjusted $p$-value). \textit{Left:} In each episode, the hypothesis $h$ and binary support label $\theta$ are defined by a held-out \textit{target stub}, which is not shown to the model. \textit{Right:} Given $h$, the model sees a sequence of supporting stubs from studies on related drugs, cell lines, and genes and must verbalize its belief $p(\theta)$ after each. By comparing sequential belief states, we can measure which stubs models find compelling based on the strength of their updates.}
\label{fig:schematic}
\end{figure}

\subsection{Contributions}
\begin{enumerate}
  \item \textbf{A novel scientific forecasting task.} We formalize evidence deliberation as forecasting whether a target hypothesis is true or not, given a pool of prior \textit{evidence stubs} revealed one at a time. The task structure uses a domain-agnostic format, is scored with proper scoring rules, and enables principled analysis of belief revision and sequential experiment selection.

  \item \textbf{\ourbenchmark{} dataset.} We instantiate the task as $96{,}000$ directly tested biological hypotheses across three domains: drug-perturbation transcriptomics (Tahoe), CRISPR gene-dependency screens (DepMap), and genome-wide association studies (GWAS). Each domain is built using five design principles that ensure class balance/diversity and provide evidence relevance labels based on a tiered expert hierarchy.

  \item \textbf{A calibration study of frontier and open models.} Frontier models rank supported hypotheses above unsupported ones, but their probabilities are poorly calibrated. Post-training Qwen3-30B-A3B against a strictly proper scoring rule yields \modelname{}, which closes the calibration gap while outperforming frontier discrimination.

  \item \textbf{Sequential analysis of belief revision in LMs.} We quantify the \textit{weight of evidence} using log Bayes factors, which measure how each new stub causally influences the model's belief about the target hypothesis. This allows us to test if updates move in the right direction, track expert relevance, have the right magnitude, and are invariant to order. Our analysis reveals interesting biases; for instance, models show ordering effects and overreact to evidence, making updates larger than the outcomes would justify.

  \item \textbf{An EIG estimator for sequential experiment selection.} We use \ourbenchmark{} to instantiate a Bayesian experimental design setting and introduce a novel factored estimator of expected information gain (EIG) that is more sample-efficient than standard Monte Carlo methods. Under a well-calibrated belief model (\modelname{}), it extracts roughly $2.5\times$ more information per experiment than a prompted frontier model. We also find that training an LLM to \textit{simulate} experimental outcomes outperforms empirical samples when used for experiment selection with our EIG estimator (\cref{sec:bed}).

  \item \textbf{Open release.} We release the \artifacthref{\urldataset}{\ourbenchmark{} datasets}, \artifacthref{\urlweights}{weights for all \modelname{} variants}, our \artifacthref{\urlcode}{training and evaluation code}, and the \artifacthref{\urllogs}{logs from our training runs}. The reproducibility statement lists each artifact in full.
\end{enumerate}

We introduce \ourbenchmark{} in \cref{sec:benchmark} and the models and metrics in \cref{sec:methods}. We then evaluate the models' ability to forecast target hypotheses' validity on \ourbenchmark{} (\cref{sec:main-results}). Next, we examine how model beliefs update as evidence is progressively disclosed; in particular focusing on prior-independent analyses (\cref{sec:woe}). Finally, we use the models to select which experiment to run next and compare their efficiency against prompted frontier models (\cref{sec:bed}).

\section{Evidence Deliberation}
\label{sec:benchmark}
\label{sec:task}

Our benchmark models the task of \textit{evidence deliberation}, in which a scientist is trying to forecast the likelihood of a \textit{target hypothesis} given a mixed pool of empirical evidence from prior studies. Evidence is organized into episodes (\cref{fig:schematic}), which each contain:
\begin{itemize}
    \item \textbf{Target hypothesis} $h$: A statement of a scientific hypothesis, which for each domain follows a consistent format; e.g., \textit{Drug $D$ downregulates gene $G$ in cell line $C$.}
    \item \textbf{Label} $\theta \in \{0, 1\}$: Whether direct scientific evidence supports $h$ ($\theta = 1$) or not ($\theta = 0$), as determined by a domain-specific criterion (\cref{tab:domains-principles}).
    \item \textbf{Evidence pool} $I_k = \{(x_1, y_1), \ldots, (x_k, y_k)\}$: A variable-length sequence of $k \in \{1, \dots, 16 \}$ tuples each comprised of an experiment $x_j$ and the outcome $y_j$ of conducting the experiment. Each entry in the sequence is formatted consistently and referred to as a stub.
\end{itemize}

\paragraph{Model beliefs \& weights of evidence.}
The model's task is to predict the probability that the target hypothesis $h$ is true conditioned on the available evidence, i.e. $p(\theta=1 | I_k)$. For notational convenience, we define $p_k = p(\theta=1 \mid I_k)$. The evolution of $p_k$ as a function of $k$ is a useful setting in which to study evidence that language models deem persuasive in favor of or against the target hypothesis. We also study belief updates using prior-invariant metrics that isolate the effect of scientific evidence on belief revisions, independent of model priors on EDB-Bio tasks (\cref{app:dose}). In particular, we leverage the \emph{weight of evidence}~\citep{good1950probability} of stub $k$, $w_k=\operatorname{logit}(p_k)-\operatorname{logit}(p_{k-1})$, where $\operatorname{logit}(p)=\log\tfrac{p}{1-p}$. We report it in \emph{decibans} (dB), $10\log_{10}$ of the Bayes factor.

\subsection{EDB-Bio Dataset}
\label{sec:EDB-Bio}

\paragraph{Domains.}
The same task format is instantiated in three biological domains: transcriptional responses to drug perturbation in cancer cell lines (Tahoe-100M;~\citealp{tahoe100m}), CRISPR gene-dependency screens \citep{depmap}, and published genetic associations from the GWAS Catalog \citep{gwascatalog}. The domains differ in what is measured, in the support criterion, and in how related evidence is defined, allowing us to test one form of evidence deliberation across interventional and observational data (\cref{tab:domains-principles}A).

\paragraph{Hypothesis and label construction.}
Each hypothesis in EDB-Bio is derived from an experimental stub ($x^*$, $y^*$) from the source dataset. Specifically, the target hypothesis is defined to be the hypothesis that $x^*$ directly tests. In stating the hypothesis $h$, we include details about $x^*$ and how the truth label $\theta$ is determined as a function of $y^*$~(\autoref{tab:domains-principles}). This minimizes ambiguity about the formulation of the hypothesis and its label.

\paragraph{Design principles.}
Because the source datasets are imbalanced and contain assay-specific artifacts, every domain is built under five design principles (\cref{tab:domains-principles}B) that control for these issues: targets with artifact or trivial labels are excluded (P1); targets are diversified and balanced $50{:}50$ across claim axes (P2); evidence pools follow a graded relevance hierarchy defined by domain experts (P3) and vary in sufficiency from evidence-rich to evidence-poor (P4); and the target experiment and outcome are never part of the pool for the corresponding hypothesis (P5). \Cref{app:benchmark} reports domain-specific construction details, and \cref{app:prompts} gives a complete prompt from each domain.

\begin{table}[tp]\centering
\includegraphics[width=\textwidth]{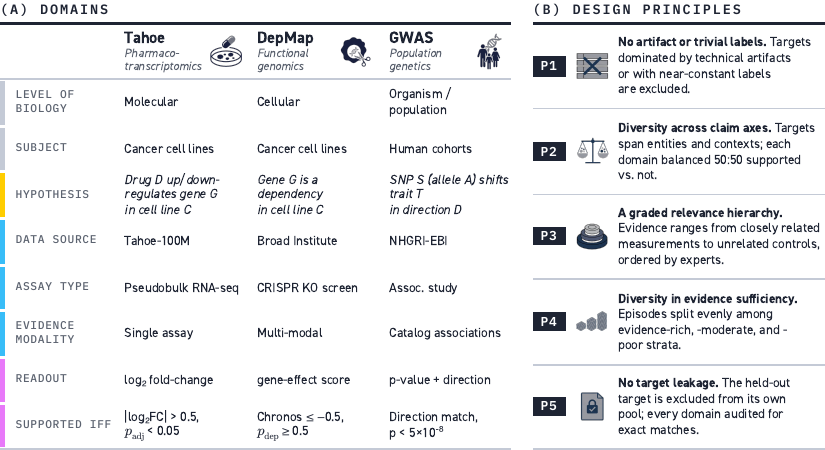}
\caption{\textbf{(A)} Comparison of the three \ourbenchmark{} domains: each instantiates the same evidence-deliberation task with a distinct assay, hypothesis, support criterion, and evidence modality. \textbf{(B)} Five overarching design principles, which are applied to each domain.}
\label{tab:domains-principles}
\end{table}

\paragraph{Dataset.}
\label{sec:dataset}
Each domain contains $32{,}000$ training, $256$ validation, and $2{,}000$ test episodes, with evidence pools of up to $k=16$ stubs. We evaluate two test configurations: the \emph{primary set} shows every test episode once at a single context length, and the \emph{sensitivity set} shows a $256$-episode subset at every $k\in\{0,\dots,16\}$ for the belief revision analyses (\cref{sec:revision}). \Cref{app:testset} gives the full configuration and provenance, together with domain-specific details.

\section{Methods}
\label{sec:methods}
\label{sec:experiments}

\subsection{Models and training}
\label{sec:models}

\paragraph{Models and baselines.}
We consider three kinds of LLMs: an open-weight model evaluated as released, our post-trained variants of it, and frontier models. Our starting checkpoint is \texttt{Qwen3-30B-A3B}, which we refer to throughout as \textbf{Qwen3}. We evaluate Qwen3 in two modes: \emph{non-thinking} prompts the model to state a probability immediately, while \emph{thinking} generates a reasoning trace before stating the probability.

\paragraph{Calibrating LLM uncertainty.}
We post-train our models against two objective functions: Brier score and binary cross-entropy, which are both strictly proper scoring rules~\citep{gneiting2007strictly}. This approach is a promising method for calibrating LLM uncertainty~\citep{damani2026beyond,puri2026reaching,baniharouni2026rewarding,leng2025taming,xu2024sayself}, but prior work has applied it mainly to factual question answering and mathematics rather than to scientific judgment (\cref{sec:related_work}). In total, we present three post-trained model variants, which we refer to collectively as \textbf{\modelname{}}:
\begin{itemize}
    \item \textbf{\modelname{} (non-thinking)} trains with a group-relative policy-gradient objective (\cref{sec:training-details}) on a negative-Brier reward $r=1-(p-\theta)^2$, where the model directly states a probability $p$ that the target hypothesis is supported, with no intermediate reasoning.
    \item \textbf{\modelname{} (thinking)} trains using the same objective but reasons in text before stating the probability.
    \item \textbf{\modelname{} (classifier)} replaces the language-model output head with a scalar linear head, applies it to the final-token representation, and minimizes binary cross-entropy (BCE).
\end{itemize}
All models were trained on a multi-node cluster with 8x NVIDIA B200 GPUs per node using \texttt{prime-rl} \citep{primerl}. Training details are provided in \cref{sec:training-details}.

\paragraph{Frontier models.}
As a strong zero-shot baseline, we evaluate two frontier models: \texttt{Claude Opus 5} (Anthropic) and \texttt{GPT-5.6-Sol} (OpenAI). Both models were fed identical prompts and evaluated at high reasoning effort with a $24{,}000$ token budget. Further details are provided in \cref{app:frontier}.

\subsection{Evaluation metrics}
\label{sec:metrics}

\paragraph{Aggregate forecast quality.}
For beliefs $p_i = p(\theta_i=1\mid I_i)$ and binary labels $\theta_i \in\{0,1\}$ we report the \emph{Brier score} $\frac1N\sum_i(p_i-\theta_i)^2$~\citep{brier1950verification}; lower is better, and because \ourbenchmark{} is balanced $50{:}50$ by construction (P2), a constant forecast at the base rate scores $0.25$.

\paragraph{Calibration vs.\ discrimination.}
We report \emph{sMECE}, a binning-free calibration error in the smooth-ECE family of \citet{blasiok2024smooth}: the density-weighted mean gap between a stated probability and the empirical frequency of support at that probability (\cref{app:evalprotocol}). \emph{AUROC} measures whether supported targets are ranked above not-supported ones, independent of calibration.

\paragraph{Calibration of weights of evidence.}
If a model's probabilities are calibrated, its weights of evidence $w$ should correlate with the label $\theta$. Specifically, across episodes in which the model starts from belief $p_{k-1}$, an update of $w$~dB should change the odds that the target hypothesis turns out to be supported by $w$~dB, compared with near-zero updates. To test this, we bin each model's updates into five tiers by $w$. Each tier's \emph{odds ratio of support} $\psi$ is its odds of support divided by the odds after near-zero updates, in decibans like $w$. It is computed within bins of starting belief, so it does not depend on the model's prior (\cref{app:woe-estimator}). Plotting $\psi$ against $w$ gives a reliability diagram for belief updates (\cref{fig:woe_reliability}, left), which we summarize by its slope $\beta$: the change in the log-odds of support per deciban of weight of evidence. Calibrated updates give $\beta=1$. When $\beta<1$, the model \emph{overreacts}: its updates are larger than the outcomes justify, and multiplying each update by $\beta$ would calibrate it on average. When $\beta>1$, the model \emph{underreacts}, and its updates are too small. The quantity $\beta-1$ is the log-odds analog of the coefficient that economists use to measure over- and underreaction to news, obtained by regressing forecast errors on forecast revisions~\citep{coibion2015information,bordalo2020overreaction}.

All statistics are equal-weight macro-averages over the three domains, with $95\%$ intervals from a bootstrap that resamples whole episodes within each domain (\cref{app:evalprotocol}).

\section{Calibration and discrimination}
\label{sec:main-results}
\label{sec:primary}

\begin{figure}[t]\centering
\includegraphics[width=\textwidth]{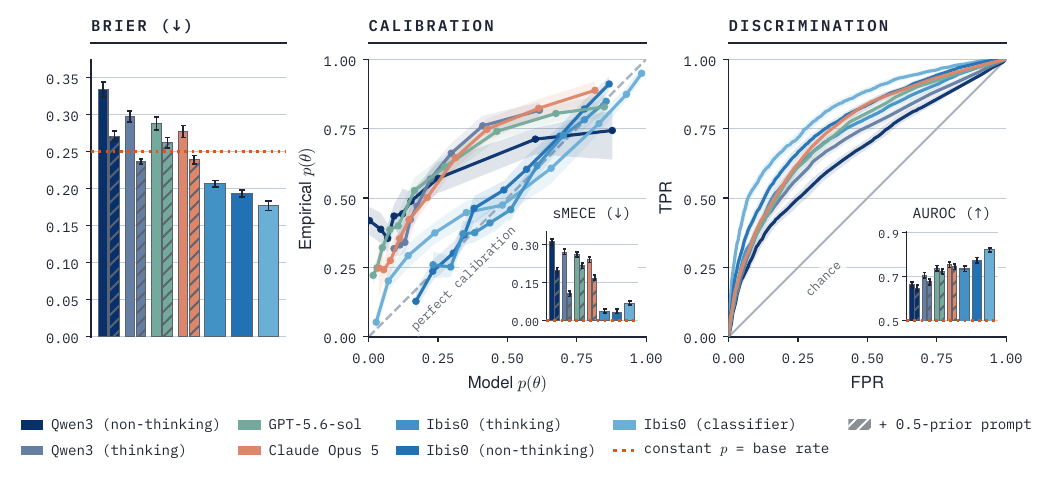}
\caption{\textbf{Do language models produce calibrated \& accurate beliefs when presented with evidence?} Zero-shot LLMs (Qwen3, GPT-5.6-sol, and Opus 5) underperform a constant $p=0.5$ forecast on Brier score (dotted line, \textit{left}). This is primarily due to poor calibration: Models that sit above the diagonal (\textit{middle}) are systematically \textit{underconfident}, despite demonstrating above-chance discrimination (\textit{right}). Explicitly instructing models to use a prior of 0.5, matching the base rate, helps somewhat (dashed bars), but post-training on proper-score objectives (\modelname{} variants) improves both calibration and discrimination substantially.}
\label{fig:headline}
\end{figure}

\begin{figure}[t]\centering
\includegraphics[width=\textwidth]{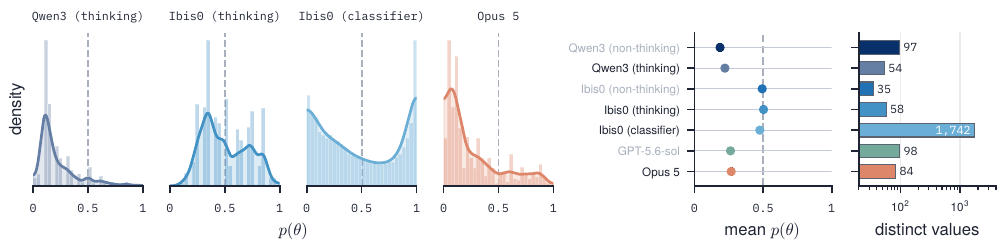}
\caption{\textbf{How do language models verbalize beliefs?} Histograms of every stated $p(\theta)$ produced as evidence accrues on the sensitivity set ($N=13{,}056$ per model). Out of the box, Qwen3 (thinking) and the frontier models are underconfident; post-training recenters the mean near the $0.5$ support rate. Models that verbalize a probability in text produce fewer than $100$ distinct values, whereas the scalar head of \modelname{} (classifier) produces a smooth distribution with $18$--$50\times$ finer resolution.}
\label{fig:prob_hist}
\end{figure}

A useful forecast must rank hypotheses correctly and assign probabilities of the right magnitude, and the two properties need not coincide. \Cref{fig:headline} reports Brier score as the overall measure and then separates calibration from discrimination. \Cref{fig:prob_hist} shows why a well-ranked forecast can still score poorly: out of the box, models state probabilities biased away from the $0.5$ base rate and drawn from only a handful of distinct values.

\paragraph{Post-training outperforms zero-shot models.} GPT-5.6-sol and Claude Opus 5 rank supported above not-supported hypotheses (AUROC $0.737$ and $0.752$), yet their Brier scores are not significantly better than a base-rate forecaster. Both Qwen3 modes show the same mismatch. All three \modelname{} models obtain Brier below $0.25$ in every domain. The classifier has the best Brier score ($0.177$) and AUROC ($0.820$), whereas the RL policies are the best calibrated, with sMECE $0.032$ against $0.066$ for the classifier. \Cref{tab:multidomain-master-full} reports per-domain results with intervals.

\paragraph{Prior-aware prompting and post-hoc recalibration.} We test two methods to improve calibration without post-training: (1) explicitly instructing models to use a prior of 0.5, and (2) fitting a monotone transform post-hoc \citep{platt1999probabilistic}. We find prompting modestly improves calibration but hurts discrimination (\cref{fig:headline}). Meanwhile, Platt rescaling approaches \modelname{} on calibration, but (by construction) does not affect discrimination (\cref{app:platt}). However, such post-hoc recalibration requires access to ground truth labels, so it is not a substitute for training.

\paragraph{Calibration is amortizable without reasoning.} While the zero-shot thinking variant of Qwen3 outperforms non-thinking, after post-training, this gap disappears. In fact, \modelname{} (thinking) marginally under-performs non-thinking, though it does provide better interpretability through its reasoning traces (\cref{app:trace}). This aligns with recent work showing that calibration can be effectively \textit{amortized} in a single forward pass \citep{zhao2025dist2ill,hollmann2023tabpfn,hollmann2025tabpfn,almeida2026jev}, particularly for tasks that do not require mathematical or symbolic reasoning \citep{sprague2025cot}.

\paragraph{Cross-domain transfer.}\label{sec:transfer}
To test generalization across domains, we train \modelname{} (classifier) on two domains and hold out the third. \cref{app:transfer} reports the results of this transfer experiment for each held-out domain in terms of \textit{retention}: the fraction of full-model performance preserved by the cross-evaluated classifier. Brier retention ranges by domain from $70.0\%$ on Tahoe to $26.3\%$ on GWAS (\cref{fig:retention}). By further decomposing Brier \citep{murphy1973new}, we observe that calibration transfers across domains, but discrimination does not. In other words, the model retains the right probability scale, but not the knowledge needed to rank hypotheses in an unseen domain.

\section{Belief revision}
\label{sec:woe}

The model's stated probabilities alone do not reveal whether a model effectively used the evidence provided or relied on its prior about the target hypothesis. We answer this question by presenting each episode sequentially at every evidence count $k\in\{0,\dots,16\}$: the value at $k=0$ measures target-only prediction (prior), and the change in belief as $k$ increases measures how the model uses the evidence. Every tested model increases its confidence in the ground-truth label as $k$ increases (\cref{app:dose}). In the Tahoe and DepMap domains, the Ibis0 classifier has both the highest magnitude of belief revision and the highest ground-truth probability at $k=16$.

\begin{figure}[tbp]\centering
\includegraphics[width=\textwidth]{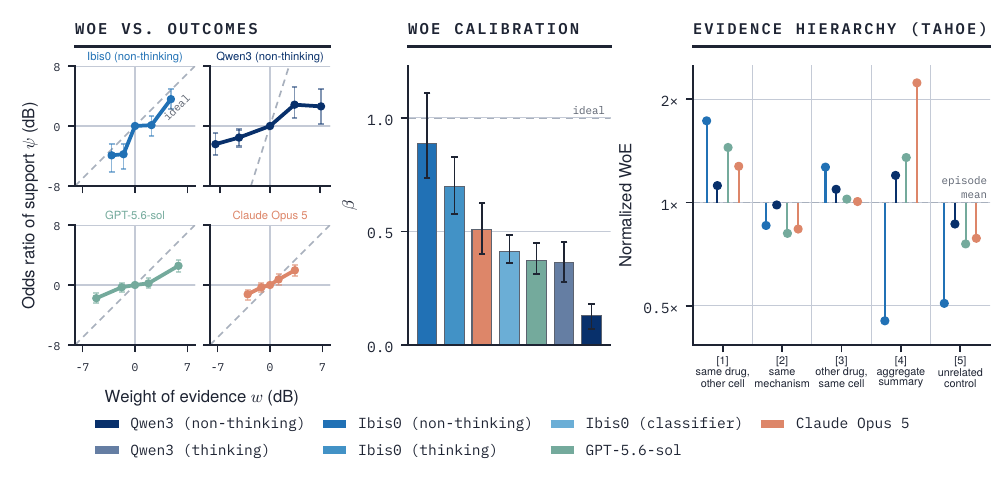}
\caption{\textbf{Do models weigh each piece of evidence at the right magnitude?} \textit{Left:} Odds ratio of support $\psi$ against weight of evidence $w$; calibrated updates lie on the diagonal ($\beta=1$). \textit{Middle:} the slope $\beta$ of each curve; $\beta<1$ indicates overreaction, meaning updates are larger than justified. Post-training brings \modelname{} (non-thinking) closest to calibrated. \textit{Right:} Evidence weights $w$ on Tahoe, normalized by episode mean and grouped by biological relevance. A model that follows the expert hierarchy should show monotonic reduction from left to right. Instead, all models under-weight evidence from non-matching drugs that share a mechanism of action with the target hypothesis [2], and all but \modelname{} over-weight genome-wide aggregate summaries [4], which are only indirectly-related to the target.}
\label{fig:woe_reliability}
\end{figure}

\paragraph{Direction and relevance of belief updates.}\label{sec:woe-direction}
Sorting stubs by their direction relative to $h$ (\cref{app:verdict}), every model's mean update is positive on supporting stubs, negative on refuting stubs, and small on inconclusive ones (\cref{fig:woe_directional}). Ordering stubs by the curated relevance hierarchy (\cref{app:principles}), \modelname{} (non-thinking) aligns most closely in every domain (Spearman $\rho=0.18$--$0.37$ across domains; \cref{app:woe-tier}), despite trailing the classifier on Brier and AUROC (\cref{fig:woe_staircase}). Where models disagree with the expert ordering, they do so together: on GWAS every model over-weights indirect same-SNP evidence, and on Tahoe, Claude Opus 5 prefers aggregate summaries to specific assay readouts (\cref{fig:woe_reliability}, \textit{right}). \Cref{app:woe-direction} reports both analyses in full.

\paragraph{Calibration of the weights of evidence.}\label{sec:woe-reliability}
As observed in \cref{fig:woe_reliability}, every model has $\beta<1$, which indicates overreaction (\cref{sec:metrics}). Values range from $0.13$ for untrained Qwen3 (non-thinking) to $0.89$ for \modelname{} (non-thinking), the only model whose interval includes $1$; the frontier models fall between. At $\beta=0.13$, the updates of Qwen3 (non-thinking) are about eight times larger than the outcomes justify. \modelname{} (classifier) has the best Brier score and AUROC of any model (\cref{sec:primary}) but a lower $\beta$ ($0.41$) than either RL policy: accurate aggregate forecasts do not imply correctly sized updates.

\paragraph{Order and composition of evidence.}\label{sec:revision}
If stated probabilities behave like beliefs, rearranging a fixed evidence set should not change the result, and the joint weight of a matched supporting and refuting pair should equal the sum of the two single-stub weights. We test both properties by holding evidence content fixed while changing its position or composition (\cref{app:legacy}, \cref{fig:belief-revision}). Consistency does not follow from aggregate forecast quality: the classifier has the best Brier score of the \modelname{} models but the largest primacy bias ($+1.01$~dB) and composition residual ($4.16$~dB), whereas the two RL policies reduce the composition residual of Qwen3 (non-thinking) from $7.54$~dB to below $1$~dB. Enabling a reasoning trace reduces primacy bias by about $0.5$~dB both before and after post-training.

\FloatBarrier

\section{Bayesian experimental design}
\label{sec:bed}

We now turn to a key application of calibrated beliefs: a Bayesian experimental design setting in which the model selects \textit{which experiment to run next}.

\paragraph{Experiment setup.} We configure \ourbenchmark{} as a sequential-selection task (\cref{app:bed-details}) on $512$ held-out episodes in each of the three domains. Starting from a target hypothesis $h$ and a pool of $k=16$ \textit{unrevealed} candidate stubs $\mathcal{C} = \{(x_1, \_), \ldots, (x_k, \_)\}$, a \textit{policy} $\pi$ chooses one experiment $x_t$ at a time to reveal the corresponding outcome $y_t$. We measure performance in two ways:

\textit{(1) Belief distribution sharpening:} An efficient policy should increase the probability assigned to the ground truth label, $\theta\, p(\theta = 1 \mid I) + (1 - \theta)\, p(\theta = 0 \mid I)$, in as few timesteps as possible.

\textit{(2) Information gain (IG):} The reduction in \textit{information entropy} \citep{shannon_mathematical_1948,mackay1992information} from experiment $x$ is $\mathbb{H}\big(p(\theta\mid I)\big) - \mathbb{H}\big(p(\theta\mid I,x,y)\big)$. To isolate the effect of experiment selection, we compute IG with the same belief model, \modelname{} (classifier), for all policies.

\paragraph{Policies.} The policies differ in how they use a language model.
\begin{itemize}
    \item \textit{Random}: $\pi_{\text{random}}$ ignores the LM and chooses a uniformly random experiment from $\mathcal{C}$.
    \item \textit{Naive prompting}: $\pi_{\text{naive}}$ prompts a model to directly select a candidate based on $h$ (\cref{app:naive-prompt}).
    \item \textit{EIG}: $\pi_{\text{EIG}}$ uses the model's verbalized belief $p(\theta \mid I)$ to select the experiment that maximizes \textit{expected information gain}, $\text{EIG}(x) = \mathbb{H}\big(p(\theta\mid I)\big) - \mathbb{E}_{y\sim p(y\mid x)}\big[\mathbb{H}\big(p(\theta\mid I,x,y)\big)\big]$ \refstepcounter{equation}(\theequation)\label{eq:eig}, where the expectation arises from uncertainty over possible outcomes $y$. We compare two approaches to sampling outcomes: (i) \textit{EIG-exemplar} draws empirical samples from the training set, and (ii) \textit{EIG-simulate} (described below) trains a simulator $p(y \mid x, \theta)$ to sample outcomes for each candidate.
\end{itemize}

\paragraph{Experiment simulator and a factored EIG estimator.} The standard Monte Carlo estimator for \cref{eq:eig} requires a \textit{simulator} $p(y \mid x, \theta)$ that can ``imagine'' possible outcomes for a candidate $x$. A simulator is easy to train from $(x, y)$ pairs by standard SFT, but conditioning it on $\theta$ requires counterfactual reasoning about how a different experiment bears on the outcome of $x$. We instead marginalize over the label $\theta'$ of the \emph{local} hypothesis $h'$ that the candidate $x$ directly tests, which factors EIG into a supportive and a non-supportive branch:

\begin{equation}
\label{eq:eig-factored}
\resizebox{0.92\linewidth}{!}{$\displaystyle
\mathrm{EIG}(x)\;\approx\;\underbrace{\mathbb{H}\big(p(\theta\mid I)\big)}_{\text{current uncertainty}} - \underbrace{\textcolor{RoyalBlue}{p(\theta'\mid I)}\;\mathop{\mathbb{E}}\limits_{\mathrlap{\rule{0pt}{1.9ex}y^{\scriptscriptstyle +}\sim\, \textcolor{BurntOrange}{p(y\mid x,\theta'{=}1)}}}\!\big[\mathbb{H}\big(\textcolor{RoyalBlue}{p(\theta\mid I,x,y^{\scriptscriptstyle +})}\big)\big]}_{\text{supportive branch}} - \underbrace{\big(1-\textcolor{RoyalBlue}{p(\theta'\mid I)}\big)\;\mathop{\mathbb{E}}\limits_{\mathrlap{\rule{0pt}{1.9ex}y^{\scriptscriptstyle -}\sim\, \textcolor{BurntOrange}{p(y\mid x,\theta'{=}0)}}}\!\big[\mathbb{H}\big(\textcolor{RoyalBlue}{p(\theta\mid I,x,y^{\scriptscriptstyle -})}\big)\big]}_{\text{non-supportive branch}}.
$}
\end{equation}

Our \textbf{factored Monte Carlo EIG estimator (\cref{eq:eig-factored})} uses two trained models: (i) a \textcolor{RoyalBlue}{belief model} (Ibis0) $M\!\approx\!p(\theta\mid I)$ that returns the branch weight $p(\theta'\mid I)$ and the revised belief $p(\theta\mid I,x,y)$, and (ii) an \textcolor{BurntOrange}{outcome simulator} $S\!\approx\!p(y\mid x,\theta')$ that generates a candidate's outcome $y$ from its design $x$ and the value of $\theta'$. We train $S$ by supervised fine-tuning a separate Qwen3-30B-A3B on the \ourbenchmark{} training split. The factorization is efficient. In practice, one outcome sample per branch gives a stable EIG estimate, and the two branches can be evaluated in parallel. Each $\mathrm{EIG}(x)$ therefore costs three \textcolor{RoyalBlue}{belief calls} (the branch weight and the two posteriors) and two \textcolor{BurntOrange}{simulator calls} (one outcome per branch); the prior entropy $\mathbb{H}(p(\theta\mid I))$ is shared across candidates and cancels in the argmax. A standard nested Monte Carlo estimate instead draws many outcomes from $p(y\mid x)$ per candidate and recomputes a posterior for each.

\begin{figure}[htbp]\centering
\includegraphics[width=\textwidth]{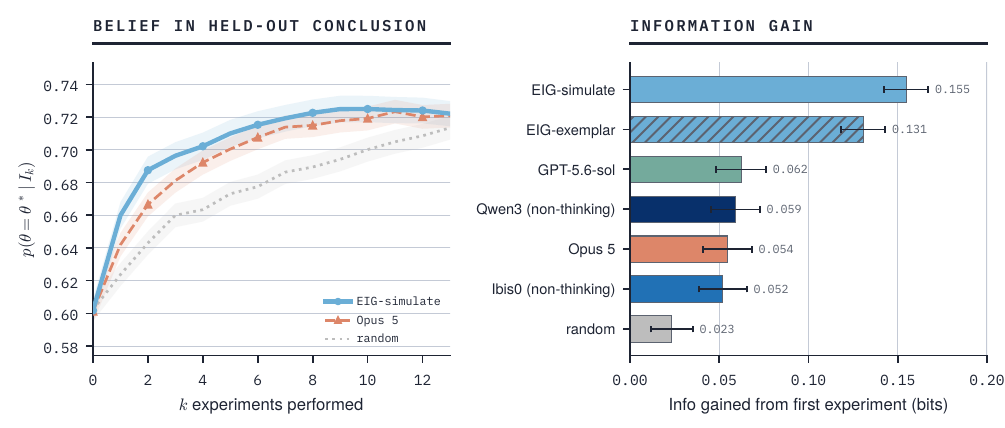}
\caption{\textbf{Experiment selection under the EIG objective.} \textit{Left:} Belief in the held-out conclusion, $p(\theta=\theta^\ast\mid I_k)$, versus experiments performed, pooled over domains ($\pm$s.e.m.). EIG-simulate, which uses \modelname{}, converges more quickly than both Opus 5 (under naive prompting) and a random policy. \textit{Right:} Information gained from the first experiment, $\mathbb{H}[p(\theta)]-\mathbb{H}[p(\theta\mid I_1)]$ (bits, $95\%$ CI), pooled over domains. EIG-simulate and EIG-exemplar extract ${\sim}2.5\times$ the information of the best prompted model.}\label{fig:bed}
\end{figure}

\paragraph{\modelname{} + EIG extracts more information per experiment.}\label{sec:bed-results} \Cref{fig:bed} (right) shows the entropy reduction for the first experiment selected by each policy, averaged over domains. EIG-simulate and EIG-exemplar reduce $0.155$ and $0.131$ bits respectively, versus at most $0.062$ bits for any naive prompting policy (Opus 5, GPT-5.6-sol, Qwen3, \modelname{}) and $0.023$ for random. The EIG policies extract roughly $2.5\times$ as much information as the best frontier policy, with non-overlapping $95\%$ CIs. The advantage is strongest on GWAS and DepMap; on Tahoe, the top information-gain and frontier policies are statistically tied. Prompted directly, \modelname{} (non-thinking) does no better than the other naive policies. In this setting, calibration improved experiment selection only when EIG was estimated and optimized explicitly at test time.

\paragraph{Our approach converges more quickly to the truth.} On belief in the held-out conclusion (\cref{fig:bed}, left), EIG-simulate reaches a given belief in roughly $30$--$40\%$ fewer experiments than random and slightly outperforms prompted Opus 5.

\paragraph{A trained simulator matches empirical exemplar outcomes.} The trained simulator (EIG-simulate) matches or slightly exceeds EIG-exemplar, which draws \textit{actual outcomes from the training set}. A model can therefore learn to simulate outcomes of unseen experiments well enough to select informative ones.

\section{Related work}
\label{sec:related_work}

\paragraph{Scientific reasoning benchmarks.}
Benchmarks for scientific AI span expert question answering~\citep{rein2023gpqa}, research coding and data-driven discovery~\citep{tian2024scicode, majumder2024discoverybench, chen2024scienceagentbench}, simulated environments for scientific reasoning and discovery~\citep{wang2022scienceworld, jansen2024discoveryworld}, and biology research workflows~\citep{laurent2026labbench2, koch2026bixbench3}. Closest to our task is claim verification against retrieved abstracts~\citep{wadden2020scifact, jansen2025matteroffact}, which scores a categorical verdict on existing evidence; \ourbenchmark{} scores a calibrated probability that a hypothesis will be supported by an experiment whose result is not yet observed.

\paragraph{Calibration and proper-scoring training.}
Work on LLM calibration has centered on confidence in answer correctness~\citep{kadavath2022language, geng2024survey}, and alignment tends to leave models overconfident~\citep{openai2023gpt4, zhu2023calibration}. Recent methods post-train against a proper scoring rule to calibrate that confidence~\citep{damani2026beyond, puri2026reaching, baniharouni2026rewarding, leng2025taming, xu2024sayself}; we extend these methods to scientific forecasting.

\paragraph{Bayesian experimental design with LLMs.}
Bayesian experimental design selects experiments by expected information gain~\citep{lindley1956measure, rainforth2024modern}, whose estimation is costly~\citep{rainforth2018nesting, foster2021deep}. A growing line drives selection from a language model's own beliefs, for information gathering~\citep{hu2024uncertainty, kobalczyk2025active, choudhury2026bedllm, grand2026shoot} and for planning perturbation screens~\citep{huang2024sequential, hao2025perturboagent}. Such selection presupposes coherent belief updating, yet language models show primacy and anchoring effects~\citep{wang2023primacy, liu2023lost, lou2024anchoring} and depart from Bayesian updating on sequential evidence~\citep{gupta2025coinflips, kim2025evidence}. We measure updates as log Bayes factors in decibans~\citep{good1950probability, kass1995bayes} and test whether they over- or underreact to the evidence.

\section{Discussion, limitations, and future work}
\label{sec:discussion}
\label{sec:conclusion}

Scientists routinely say that something is \textit{known}. Epistemology and meta-epidemiology researchers \citep{kuhn1997structure, ioannidis2005most, schulz1995empirical} have fruitfully studied the various methodological, social and institutional forces behind that phrase. Our work is motivated by a complementary focus, which we term \emph{agent epistemology}: what compels an AI system to shift its belief in a scientific claim? Framing evidence deliberation as calibrated probabilistic forecasting makes this question measurable. The same approach outlined in this paper could be used to more directly assess methodological notes \citep{sterne2019rob}, absence of detail \citep{chan2004empirical}, or other meta-scientific information such as conflicts of interest or institutional affiliations \citep{lundh2017industry}, questions that today largely rest on human intuition when weighing scientific evidence. In contrast, a well-calibrated model trained across a sufficiently broad scientific literature may help scientists surface systematic credibility issues with studies.

\paragraph{Limitations and future work.}
\label{sec:limitations}
While constructed from real biological data sources, \ourbenchmark{} is synthetic in several respects: evidence stubs are structured summaries of the underlying studies, hypotheses are binary, and every domain is deliberately balanced at $p(\theta) = 0.5$ (P2), which may account for the cross-domain calibration transfer (\Cref{sec:transfer}). We post-train one model family (Qwen3) at one scale, and the factored EIG estimator, while cheaper than nested Monte Carlo~\citep{rainforth2018nesting}, still costs more than naive prompting. Finally, we treat a scientific belief as a single probability, but belief in ordinary usage is richer~\citep{hawthorne2016belief}; we adopt the probabilistic framing because it makes deliberation measurable via Bayesian analysis. We assume that experiments are uniform and have fixed cost, whereas real research programs face budgets and timelines; extending the selection objective to cost-aware acquisition is a natural next step, as is moving from structured evidence stubs to free-text evidence drawn from the literature.

\clearpage


\section*{AI use statement}
This research was carried out by computer scientists and biologists, with the assistance of AI agents and tools. In this work, we used generative AI tools to clean and reformat datasets, to provide feedback on research methodology and experimental design, to implement methods, and to support qualitative analysis of model reasoning traces. We have not used generative AI tools to generate data: AI tools helped write the scripts that construct \ourbenchmark{}, but the data itself is sourced from published scientific studies (\cref{app:benchmark}). Formulating mathematical claims, providing ingredients for or writing proofs, and translation are not applicable to this work. Additionally, we used generative AI tools to draft and edit parts of the paper for readability, to suggest its structure, to summarize and identify related literature, to create and edit software code, to generate figures with matplotlib, and to draft schematic figures with Claude Design. The tools used were Claude Code, Claude Design, OpenAI Codex, and the models Claude Opus 4.8, Claude Opus 5, Claude Opus 5.5, Claude Sonnet 5, Claude Fable 5.1, GPT-5.6-sol, and GPT-6 Astra.

We have reviewed all AI-assisted work. All AI-drafted text was reviewed and revised by the authors; reported numbers were checked against the evaluation outputs; every citation was checked against its source; and AI-assisted code was reviewed and tested by the authors. Because Claude Opus 5 and GPT-5.6-sol are also evaluated as baselines, we note that AI assistance did not produce or alter any reported result. We take responsibility for the final content of this work, including text, claims, and artifacts produced with the aid of generative AI.

\section*{Ethics statement}

This work involves no human subjects and no private or personally identifiable data. EDB-Bio is built entirely from public, curated scientific resources used under their respective licenses; the GWAS data are aggregate summary statistics, not individual genotypes.

Advancing the scientific capabilities of language models is dual-use: the same calibrated hypothesis evaluation that can accelerate beneficial research could in principle inform harmful research. However, EDB-Bio does not include examples of clearly harmful scientific hypotheses, and instead focuses on pre-existing hypotheses of relevance to cancer therapeutic design.

Finally, the Ibis0 models described in this work produce probabilistic judgments that remain imperfect and have not been exhaustively evaluated across other areas of life science; they are shared in the spirit of open-science research, not as a substitute for experimental validation and expert review.

\section*{Reproducibility statement}
\ifanon
\else
We release the artifacts needed to reproduce every result in this paper:
{\raggedright
\begin{itemize}
  \item \textbf{Dataset.} The \ourbenchmark{} benchmark, with commands to regenerate each domain from its source data: \artifacturl{\urldataset}{dataset}.
  \item \textbf{Test sets.} The primary and sensitivity test sets (\cref{app:testset}): \artifacturl{\urltestset}{test sets}.
  \item \textbf{Models.} Weights for all \modelname{} variants and the experimental-design outcome simulator: \artifacturl{\urlweights}{model weights}.
  \item \textbf{Code.} Training configurations, the model-evaluation harness, and the calibration analyses, in a pinned dependency environment: \artifacturl{\urlcode}{code repository}.
  \item \textbf{Training logs and evaluation outputs.} Logs from all training runs and per-episode predictions of every evaluated model: \artifacturl{\urllogs}{training logs}.
\end{itemize}
\par}
Frontier models were queried through their providers' APIs with the model identifiers and settings in \cref{app:frontier}. Code and model weights are released under the Apache-2.0 licence, and the datasets, logs and evaluation outputs under CC BY 4.0; upstream data sources keep their own terms, listed on the dataset card.
\fi

\paperbibliography

\newpage
\appendix
\crefalias{section}{appendix}
\crefalias{subsection}{subappendix}
\crefalias{subsubsection}{subsubappendix}

\addtocontents{toc}{\protect\setcounter{tocdepth}{2}}
\renewcommand{\contentsname}{Appendix contents}
\tableofcontents
\clearpage

\section{\ourbenchmark{} construction details}
\label{app:benchmark}

This appendix expands \cref{sec:benchmark}: an overview of the three domains (\cref{app:domains}), the design principles applied to all of them (\cref{app:principles}), the construction and audit of each domain (\crefrange{app:construction}{app:gwas}), the dataset splits and test-set provenance (\cref{app:testset}), and a known limitation of the dataset construction (\cref{app:overlap}). Each domain is built by a single seeded script that emits \ourbenchmark{} together with a manifest recording every parameter; the scripts are released with the code.

\subsection{Domains}
\label{app:domains}

In each domain, the target measurement is excluded from the evidence pool, and the amount, relevance, and strength of the remaining evidence are controlled during construction. Each stub presents the measurement with its scientific context. In Tahoe, this is the perturbing drug's mechanism, approval status, and chemical structure, and the cell line's lineage and driver mutations; in DepMap and GWAS, it is analogous assay annotations and study provenance, such as gene function, variant consequence, cohort ancestry, and sample size. \Cref{tab:domains-principles} in the main text compares the domains, \crefrange{app:construction}{app:gwas} give construction details, and \cref{app:prompts} gives a complete prompt from each domain.

\paragraph{Tahoe: drug perturbation of cancer cell lines.}
The Tahoe-100M atlas~\citep{tahoe100m} measures transcriptional responses to small-molecule perturbations across cancer cell lines, supporting the study of drug mechanism and of how response varies across cellular contexts. A hypothesis states that ``drug $D$ up/down-regulates gene $G$ in cell line $C$''; its held-out DESeq2 measurement supports the claim when $\mathrm{padj}<0.05$, $|\log_2\mathrm{FC}|>0.5$, and the direction matches. Evidence consists of differential-expression measurements of $G$ under other drug and cell-line conditions, ordered by relevance from the same drug in another cell line, through related mechanisms and other drugs, to a genome-wide aggregate summary and unrelated controls.

\paragraph{DepMap: cancer gene dependency.}
The Broad Institute Cancer Dependency Map (DepMap)~\citep{depmap} uses pooled CRISPR screens to identify genes that cancer cells require for growth, with applications to therapeutic-target discovery and to mapping context-specific cancer vulnerabilities. A hypothesis states that ``gene $G$ is a dependency in cell line $C$''; the held-out screen supports the hypothesis when the Chronos gene-effect score is at most $-0.5$ and the calibrated dependency probability is at least $0.5$~\citep{chronos}. Evidence is multi-modal and ordered by how directly it bears on $G$'s dependency in $C$: the dependency of $G$ in related cell lines, then co-essentiality (genes whose across-cell-line dependency profiles correlate with $G$'s, implying shared function), pharmacological response (the sensitivity of $C$ to drugs that target $G$ or its pathway~\citep{prism}), and the molecular state of $G$ in $C$ (its expression, mutation, and copy-number status). The dependencies of unrelated genes serve as irrelevant controls.

\paragraph{GWAS: genetic associations.}
The NHGRI-EBI GWAS Catalog~\citep{gwascatalog} curates published associations between human genetic variants and traits, providing a resource for mapping inherited risk and trait biology. A hypothesis states that ``SNP $S$ (risk allele $A$) influences trait $T$ in a stated direction''; the held-out association supports the hypothesis when $p<5\times10^{-8}$ and the effect direction matches. Evidence consists of other catalog associations, ordered by relevance from the same gene or variant and a related trait, through pathway-neighbor associations, to unrelated loci and traits.

\subsection{Design principles}
\label{app:principles}
The source datasets are imbalanced and contain assay-specific shortcuts. We apply five principles to remove shortcuts based on label prevalence or assay artifacts. \Cref{tab:principles} records how each principle is instantiated in each domain; the domain sections below give the corresponding selection rules and audits.

\paragraph{P1: No artifact or trivial labels.}
Target measurements dominated by technical artifacts are excluded, as are entities whose label is nearly constant across contexts. The restriction applies to targets; excluded entities may still appear as evidence.

\paragraph{P2: Diversity across claim axes.}
Targets span the biological entities and contexts on which we evaluate generalization, with caps or stratification to prevent any one group from dominating. Each domain is balanced $50{:}50$ between supported and not-supported targets.

\paragraph{P3: A graded relevance hierarchy.}
Evidence pools range from measurements closely related to the target to deliberately unrelated controls. The ordering within each domain's hierarchy was defined by PhD-level bioinformaticians with peer-reviewed publications in the relevant biological fields, rather than by an automated heuristic. In Tahoe, for example, the hierarchy runs from the same drug in another cell line, through drugs with the same mechanism and other drugs in the same cell line, to unrelated controls.

\paragraph{P4: Diversity in evidence sufficiency.}
Episodes are divided approximately equally among evidence-rich, -moderate, and -poor strata. A calibrated model should become confident when evidence is conclusive and remain uncertain when it is not.

\paragraph{P5: No target leakage.}
The held-out target measurement is excluded from its own evidence pool, and every domain is audited for exact matches. Domain-specific near-duplicate risks, such as variants in high linkage disequilibrium, are discussed in the domain sections below. Overlap between episodes across splits is a separate matter, treated in \cref{app:overlap}.

\begin{table}[htbp]\centering\footnotesize
\renewcommand{\arraystretch}{1.2}
\setlength{\tabcolsep}{4pt}
\begin{tabular}{p{0.115\textwidth} >{\raggedright\arraybackslash}p{0.245\textwidth} >{\raggedright\arraybackslash}p{0.245\textwidth} >{\raggedright\arraybackslash}p{0.245\textwidth}}
\toprule
 & \textbf{Tahoe} & \textbf{DepMap} & \textbf{GWAS} \\
\midrule
\textbf{P1} Targets &
  Mitochondrial, ribosomal, pseudogene and housekeeping-lncRNA genes excluded, as are near-zero
  base-mean measurements &
  Pan-essential and non-expressed genes excluded; targets drawn from the $2{,}480$
  \emph{selective}-dependency genes (essential in $2$--$50\%$ of lines) &
  MHC/HLA variants excluded (long-range linkage disequilibrium makes single-variant labels
  unreliable), as are heavily replicated ``textbook'' pairs \\
\textbf{P2} Diversity &
  $49$ cell lines, $179$ drugs across mechanisms, $13{,}083$ genes; balanced within every cell line &
  $30$ lineages, thousands of genes, at most $16$ episodes per gene &
  Thousands of genes and traits across EFO parent categories, with per-gene and per-trait caps; one
  episode per unique (SNP, trait) pair \\
\textbf{P3} Hierarchy &
  Within-assay: same drug, other line $\succ$ same mechanism $\succ$ other drug, same line $\succ$
  aggregate summary $\succ$ unrelated control &
  Cross-modal: same gene, same lineage $\succ$ same gene, other lineage $\succ$ co-essential partner
  $\succ$ drug targeting $G$ $\succ$ molecular state of $G$ in $C$ $\succ$ unrelated control &
  Locus and trait proximity: other SNP in gene, same trait $\succ$ same gene, related trait $\succ$
  target SNP, related trait $\succ$ target SNP, unrelated trait $\succ$ pathway neighbor $\succ$
  unrelated control \\
\textbf{P4} Sufficiency &
  Significance-rich ($\ge4$ significant stubs), -moderate ($1$--$3$), or -poor (none) &
  Rich, moderate or poor in dependency-supporting evidence: dependent related lines,
  over-expression, amplification, damaging mutation, drug sensitivity &
  Rich, moderate or poor by capping significant related stubs. Non-significant related evidence is
  scarce in a catalog of significant hits, so poor pools fill largely with controls \\
\textbf{P5} Leakage &
  No pool stub shares the target's exact (drug, cell line) pair &
  Same-gene stubs come only from other lines; the molecular state of $G$ in $C$ is a different assay
  from the held-out fitness readout, so it is evidence rather than leakage &
  The exact (SNP, trait) pair is excluded. A different SNP in high linkage disequilibrium carries
  nearly the same signal; filtering it needs an LD reference panel and is deferred \\
\bottomrule
\end{tabular}
\caption{How each \ourbenchmark{} design principle is instantiated in the three domains. Full criteria and audits are in \crefrange{app:construction}{app:gwas}.}
\label{tab:principles}
\end{table}

\subsection{Tahoe: drug perturbation of cancer cell lines}
\label{app:construction}

We use the Tahoe-100M pseudobulk differential-expression tables, which report one DESeq2 result for each drug\,$\times$\,cell-line\,$\times$\,gene combination and are stored one cell line per file. Rows with failed DESeq2 fits are discarded, and a result is \emph{significant} iff $\mathrm{padj}<0.05 \wedge |\log_2\mathrm{FC}|>0.5$. Significant results are rare (roughly $9\%$ of measurements). Principles P1--P5 (\cref{app:principles}) are applied as follows.

\textbf{Target principles}
\begin{description}
  \item[P1 (No artifact labels).] Mitochondrial, ribosomal, and pseudogene genes, and measurements with near-zero base mean, are excluded as targets because their differential-expression calls reflect technical properties of the assay rather than drug action. Both remain admissible as evidence. The realized artifact-gene target rate is $0\%$.
  \item[P2 (Diversity across claim axes).] Targets span $49$ cell lines (no line contributes more than $2.0\%$ of episodes), $179$ drugs across mechanisms of action, and $13{,}083$ genes. All available cell lines and a shared set of drugs are retained so that every target has same-drug, different-cell-line evidence. Because significant results are rare, targets are drawn with fixed per-cell-line quotas from pre-classified candidate sets. A significant measurement yields a supported target (claimed direction equal to the observed one) or a refuted target (opposite direction); a non-significant measurement yields an inconclusive target. The supported rate is $0.50$ within every cell line and $49.9\%$ overall.
\end{description}

\textbf{Evidence principles}
\begin{description}
  \item[P3 (Relevance diversity).] Unlike those of DepMap and GWAS, the Tahoe hierarchy is \emph{within-assay}: every stub is a differential-expression measurement of the target gene under another $(\text{drug},\text{cell line})$ condition. We rank a stub's relevance to the target, most to least, as: (1) the same drug in another cell line; (2) a drug with the same mechanism of action; (3) a different drug in the same cell line; (4) a genome-wide aggregate summary for the target drug and cell line; and (5) an unrelated control. Collapsing (1)--(4) into \emph{related} and (5) into \emph{control}, we fill each pool to an $\approx 60{:}40$ ratio, drawing from the top of the hierarchy first. The realized split is $62.5\%$ related and $37.5\%$ control (same-drug $24\%$, same-mechanism $18\%$, different-drug $18\%$ of stubs), and same-drug evidence is present in $100\%$ of episodes. Each stub is tagged with its relation to the target and carries the drug and cell-line context listed in \cref{app:domains}.
  \item[P4 (Sufficiency diversity).] Episodes vary in how much conclusive (significant) related evidence their pool contains: rich ($\ge4$ significant related stubs), moderate ($1$--$3$), or poor (none), in equal thirds. Conclusive related stubs are added from highest to lowest relevance until the stratum's quota is met; inconclusive related stubs are then added round-robin across relevance levels, followed by one genome-wide aggregate summary and unrelated controls. Rich episodes carry $5.7$ conclusive stubs on average against $0.8$ for poor episodes, and $88\%$ of all stubs are inconclusive.
  \item[P5 (No leakage).] A target's own held-out measurement never appears in its pool, and no pool stub shares the target's exact $(\text{drug},\text{cell line})$ pair; both are enforced by construction.
\end{description}

\paragraph{Class balance and the negative split.} Supported and not-supported targets are balanced $50{:}50$. A Tahoe \emph{refuted} outcome (a significant change opposite to the claimed direction) is comparatively common, so negatives are sampled at approximately $67\%$ inconclusive and $33\%$ refuted.

\subsection{DepMap: cancer gene dependency}
\label{app:depmap}

Dependencies comprise $9.7\%$ of valid DepMap gene\,$\times$\,cell-line pairs. Principles P1--P5 (\cref{app:principles}) are applied as follows.

\textbf{Target principles}
\begin{description}
  \item[P1 (No artifact labels).] Pan-essential (common-essential) genes, which are dependencies in nearly every line, and non-expressed genes, which cannot be dependencies, are excluded as targets. Both may still appear as (uninformative) evidence.
  \item[P2 (Diversity across claim axes).] Targets span the $30$ lineages and thousands of genes over which we make claims, with no stratum dominating (at most $16$ episodes per gene). Unlike Tahoe, DepMap has no target-drug or mechanism axis: the perturbation is always a knockout, so drugs enter only as evidence. Because dependencies are rare (close to Tahoe's $\sim\!9\%$ significant rate), targets are sampled to obtain a $50{:}50$ supported-to-not-supported balance. Targets are drawn from the $2{,}480$ \emph{selective}-dependency genes, which are essential in some cell lines but not others ($2$--$50\%$ of lines), so that the support label varies across cell-line contexts.
\end{description}

\textbf{Evidence principles}
\begin{description}
  \item[P3 (Relevance diversity).] We rank a stub's relevance to the target, most to least, as: (1) the same gene knocked out in another cell line of the same lineage; (2) the same gene in a different lineage; (3) a co-essential pathway partner of $G$ knocked out in $C$; (4) a drug targeting $G$ or its pathway profiled in $C$; (5) the molecular state of $G$ in $C$ (expression, copy number, mutation); and (6) an unrelated gene's fitness in $C$. Collapsing (1)--(5) into \emph{related} and (6) into \emph{irrelevant}, we fill each pool to an $\approx 60{:}40$ ratio, drawing from the top of the hierarchy first and guaranteeing multi-modal coverage. This hierarchy is \emph{cross-modal}, spanning genetic, pharmacological and molecular evidence. As in Tahoe, each stub is tagged with its relation to the target (e.g.\ \emph{target gene knocked out in a different cell line of the same lineage}), which keeps the prompt structure common across domains. Each stub also carries its context: the gene (name, family, identifiers), the cell line (lineage, disease, subtype, demographics, driver mutations), and for pharmacological stubs the drug (targets, mechanism, SMILES).
  \item[P4 (Sufficiency diversity).] Episodes vary in how much evidence \emph{supporting} a dependency their pool contains (related cell lines that are themselves dependencies, over-expression, amplification, damaging mutation, drug sensitivity): pools are built to be sufficiency-rich, -moderate, or -poor.
  \item[P5 (No leakage).] A target's own held-out CRISPR measurement never appears in its pool; same-gene stubs are drawn only from \emph{other} cell lines. The target gene's \emph{molecular} state in $C$ (expression, copy number, mutation) is measured by a different assay from the held-out fitness readout and is retained as evidence.
\end{description}

\paragraph{Class balance and the negative split.} Supported and not-supported targets are balanced $50{:}50$, as in Tahoe. DepMap's \emph{refuted} outcome, a knockout that \emph{speeds} growth, is biologically rare (only $2.7\%$ of measurements have $\mathrm{Chronos}>0.25$), so negatives are sampled at approximately $85\%$ inconclusive and $15\%$ refuted.

\subsection{GWAS: genetic associations}
\label{app:gwas}

The GWAS Catalog is a curated collection of published genome-wide-significant associations and contains few natural negatives. Principles P1--P5 (\cref{app:principles}) are applied as follows.

\textbf{Target principles}
\begin{description}
  \item[P1 (No artifact/trivial labels).] Variants in the MHC/HLA region (chr6, $\sim\!25$--$34$~Mb) are excluded as targets because their extreme long-range linkage disequilibrium makes single-variant labels unreliable. Heavily replicated ``textbook'' pairs (those studied in dozens of prior GWAS) are also excluded to reduce dependence on frequently reported associations. Both classes remain admissible as (uninformative or trivially informative) evidence.
  \item[P2 (Diversity across claim axes).] Targets span thousands of genes and traits across the EFO parent categories over which we make claims, with per-gene \emph{and} per-trait caps so that no stratum dominates; one episode corresponds to one unique (SNP, trait) pair. Unlike Tahoe, GWAS has no perturbation: it is observational, so the ``experiment'' is the held-out association study itself. Because positives (genome-wide-significant, direction-consistent associations) are abundant and negatives scarce, targets are sampled across outcome classes to obtain a $50{:}50$ supported-to-not-supported balance.
\end{description}

\textbf{Evidence principles}
\begin{description}
  \item[P3 (Relevance diversity).] We rank a stub's relevance to the target, most to least, as: (1) a different SNP in the target gene associated with the same trait; (2) a different SNP in the target gene associated with a trait in the same EFO parent (related trait); (3) the target SNP associated with a different trait in the same EFO parent; (4) the target SNP associated with a trait in a different EFO parent; (5) a SNP in a pathway-neighbor gene associated with the target trait; and (6) an unrelated SNP and trait (different gene, different EFO parent). This hierarchy is graded by both \emph{genomic} proximity (same SNP $\to$ same gene $\to$ pathway neighbor $\to$ unrelated locus) and \emph{phenotypic} proximity (same trait $\to$ related trait $\to$ unrelated trait). Collapsing (1)--(5) into \emph{related} and (6) into \emph{control}, we fill each pool toward a related:control target of approximately $60{:}40$, drawing from the top of the hierarchy first. Each stub is tagged with its relation to the target.
  \item[P4 (Sufficiency diversity).] Episodes vary in how much genome-wide-significant related evidence their pool contains; pools are built in roughly equal thirds to be sufficiency-rich, -moderate, or -poor by capping the number of significant related stubs. Because the catalog contains only significant hits, non-significant \emph{related} evidence is scarce, so the ``poor'' pools fill largely with unrelated controls and the realized related:control ratio falls below its target.
  \item[P5 (No leakage).] A target's own held-out (SNP, trait) association never appears in its pool; same-gene and same-SNP stubs are drawn only from \emph{other} SNPs or \emph{other} traits, and the exact-pair exclusion is audited to zero. This check does not detect variants in high linkage disequilibrium with the target, which may convey nearly the same signal. Filtering such variants requires an LD reference panel and is not performed.
\end{description}

\paragraph{Class balance and the negative split.} Supported and not-supported targets are balanced $50{:}50$, as in Tahoe and DepMap. A GWAS \emph{direction-inconsistent} outcome is a genome-wide-significant association whose effect points in the opposite direction. It is rare relative to a \emph{sub-threshold} association, so negatives are sampled at $\approx\!85\%$ sub-threshold and $15\%$ direction-inconsistent. This split concerns the \emph{target} outcome. At the \emph{evidence} level, GWAS has no refuting-stub category: once the target association is held out, other catalog associations can corroborate the hypothesis or be uninformative, but do not directly refute it. The directional analyses therefore omit GWAS (\cref{app:principles}).

\subsection{Dataset splits and test-set provenance}
\label{app:testset}

\paragraph{Dataset splits.}
Each domain contains $32{,}000$ training episodes, $256$ separate validation episodes reserved for model selection, and $2{,}000$ test episodes; both test configurations below are drawn from the test split. Within each domain, labels are balanced approximately $50{:}50$, and evidence order is randomized independently of relevance. During training, the context length is sampled as $k\sim\mathrm{Uniform}\{2,\dots,16\}$. The sensitivity set has the same per-domain size as the validation split but is disjoint from it.

\paragraph{Differences between domains.}
All three domains use evidence pools of at most $16$ stubs. They differ in the composition of the not-supported class: approximately $33\%$ of not-supported targets are refuted in Tahoe and $15\%$ in DepMap and GWAS (direction-inconsistent associations), with the remainder inconclusive or sub-threshold. GWAS also has no refuting \emph{evidence} stubs (\cref{app:gwas}).

\paragraph{Test-set configurations.}
The primary and sensitivity test sets are statically materialized and published as one versioned artifact. Each row holds one episode with its evidence pool truncated to the specified context length and corresponds to one forward pass. All models receive identical rendered inputs, including context length and evidence order. A domain column supports both per-domain and macro-averaged analyses from the same tables.

In the \emph{primary} set, each episode appears once at a context length within the training range. It is used for the master table, reliability and ROC panels, and leave-one-out transfer matrix. In the \emph{sensitivity} set, each episode appears at every context length from $0$ to $16$. The resulting nested sweep is used for dose--response and weight-of-evidence analyses.

\begin{table}[htbp]\centering\small
\renewcommand{\arraystretch}{1.15}
\begin{tabular}{l r r}
\toprule
 & \textbf{Primary} & \textbf{Sensitivity} \\
\midrule
Episodes per domain & $2{,}000$ & $256$ \\
Episodes, all three domains & $6{,}000$ & $768$ \\
Context lengths per episode & $1$, from $k \in \{2,\dots,16\}$ & $17$, all of $k \in \{0,\dots,16\}$ \\
Rows per domain & $2{,}000$ & $4{,}352$ \\
Rows, all three domains & $6{,}000$ & $13{,}056$ \\
Models evaluated & $10$ & $7$ \\
Forward passes & $60{,}000$ & $91{,}392$ \\
\bottomrule
\end{tabular}
\caption{Primary and sensitivity test-set configurations. Both are published in the \texttt{testsets/} folder of \repohandle{\hfdataset} as six configs (two sets $\times$ three domains).}
\label{tab:testset-configs}
\end{table}

\paragraph{Primary set.}
The primary set contains $2{,}000$ episodes per domain and $6{,}000$ in total, comprising the complete held-out TEST split of each domain. Each episode is assigned one context length $k \in \{2,\dots,16\}$, matching the \texttt{context\_size\_range} $=[2,16]$ under which every trained model saw its episodes.

Context lengths are allocated in fixed counts. Each of the $15$ values receives $\lfloor 2000/15 \rfloor=133$ episodes, and $k\in\{12,\dots,16\}$ receives one additional episode, for $134$. Uniform random assignment would have a multinomial standard deviation near $11$ episodes per group; fixed allocation removes that variation.

Labels are close to balanced by construction: the supported rate is $50.0\%$ for both GWAS and DepMap and $49.9\%$ for Tahoe.

\paragraph{Sensitivity set.}
The sensitivity set contains $256$ episodes per domain, each drawn from the corresponding primary set and scored at all $17$ context lengths $k \in \{0,1,\dots,16\}$: $4{,}352$ rows per domain, $13{,}056$ per model. The context at $k$ is the first $k$ stubs of the episode's fixed, pre-shuffled evidence pool, so the sweep is nested. The stub revealed at step $j$ is the same for every model and at every $k \ge j$. This permits per-stub attribution in the weight-of-evidence analysis, where each increment is a paired difference between consecutive context lengths within an episode.

The $k=0$ and $k=1$ conditions are below the training range. At $k=0$, the model receives the hypothesis without evidence, providing the reference forecast used by the cumulative weight-of-evidence metric. The $k=1$ condition measures the initial response to evidence.

\paragraph{Deterministic construction.}
Within each domain, TEST episodes are ordered by the SHA-256 hash of the episode identifier. Context lengths are assigned in balanced blocks over this order, and the first $256$ episodes form the sensitivity set. Selection is therefore independent of source-row order. Rebuilding with seed $0$ reproduces all six parquet files byte for byte.

\paragraph{Provenance and release.}
Both sets are published in the \texttt{testsets/} folder of \repohandle{\hfdataset}, at the same tag (\texttt{v1.0.0}) as the \ourbenchmark{} splits they are built from. Release details for the test sets, builder, and \ourbenchmark{} are given in the reproducibility statement.

\subsection{Known limitations of dataset construction}
\label{app:overlap}
\renewcommand{\topfraction}{0.92}
\renewcommand{\textfraction}{0.01}
\renewcommand{\floatpagefraction}{0.9}

\ourbenchmark{} assigns episodes to the training, validation, and test splits by i.i.d.\ sampling (\cref{app:testset}), so test targets can overlap with training evidence. Principle P5 excludes a target's own measurement from its own evidence pool but does not constrain other episodes: the held-out target of one episode may serve as an evidence stub in another. When the two episodes fall on opposite sides of the split, a model trained on \ourbenchmark{} has seen the test target's readout during training. We measure the extent of this overlap and its effect on the reported results, which is small, and leave its removal to a future revision of the dataset.

\paragraph{Extent of the overlap.}
For each test episode we check whether its held-out measurement, identified by the domain's target key, appears as an evidence stub within the first $16$ positions of any training or validation pool, which is the portion of a pool that training renders. \Cref{tab:overlap} gives the result. Overlap affects $0.6\%$ of test episodes in Tahoe, $10.1\%$ in DepMap, and $53.2\%$ in GWAS. A stricter check, whether a test hypothesis itself recurs as a training target, finds no cases in DepMap or GWAS and $11$ in Tahoe, all of them one measurement claimed in opposite directions.

The spread across domains follows from the size of each domain's measurement space (the set of possible target keys) relative to the number of stubs that the training pools draw from it. Tahoe targets come from a product of $179$ drugs, $49$ cell lines, and $13{,}083$ genes, about $115$ million combinations, and the training pools expose $0.39\%$ of them. DepMap's space has two axes and about $2.4$ million pairs, and the restriction to selective-dependency genes under P1 concentrates about $16$ episodes on each of $2{,}047$ genes, so the pools expose $15\%$. GWAS is not a product space: a (SNP, trait) pair exists only if a study reported it, and the relevance tiers enumerate sibling records in the same gene or at the same SNP rather than sampling them. The training pools cover $92.5\%$ of the associations that the dataset touches. Under uniform sampling the overlap rate would equal this coverage, and Tahoe and DepMap fall within a factor of $1.5$ of it. The construction choices that limit identity shortcuts, a narrow target set under P1 and relevance-first pool filling under P3, are the same choices that raise the overlap rate.

\begin{table}[htbp]\centering\footnotesize
\renewcommand{\arraystretch}{1.15}
\setlength{\tabcolsep}{4pt}
\begin{tabular}{l l r r r r}
\toprule
\textbf{Domain} & \textbf{Measurement key} & \textbf{Measurement space} & \textbf{Exposed as stubs} & \textbf{Coverage} & \textbf{Test overlap} \\
\midrule
Tahoe  & (drug, cell line, gene) & $114{,}750{,}993$ & $450{,}307$ & $0.39\%$ & $11$ ($0.6\%$) \\
DepMap & (gene, cell line)       & $2{,}411{,}366$   & $365{,}141$ & $15.1\%$ & $202$ ($10.1\%$) \\
GWAS   & (SNP, trait)            & $198{,}001^{*}$   & $183{,}166$ & $92.5\%$ & $1{,}065$ ($53.2\%$) \\
\bottomrule
\end{tabular}
\caption{Overlap between test targets and training evidence. \emph{Measurement space} is the number of possible target keys; \emph{exposed as stubs} is the number of distinct measurements that appear as an evidence stub in the first $16$ positions of any training or validation pool; \emph{coverage} is their ratio; \emph{test overlap} is the number of the $2{,}000$ test episodes per domain whose held-out measurement is among the exposed stubs. $^{*}$GWAS has no product space, so the entry is the number of distinct (SNP, trait) associations that appear anywhere in the dataset, and coverage is relative to that set.}
\label{tab:overlap}
\end{table}

\begin{figure}[tbp]\centering
\includegraphics[width=\textwidth]{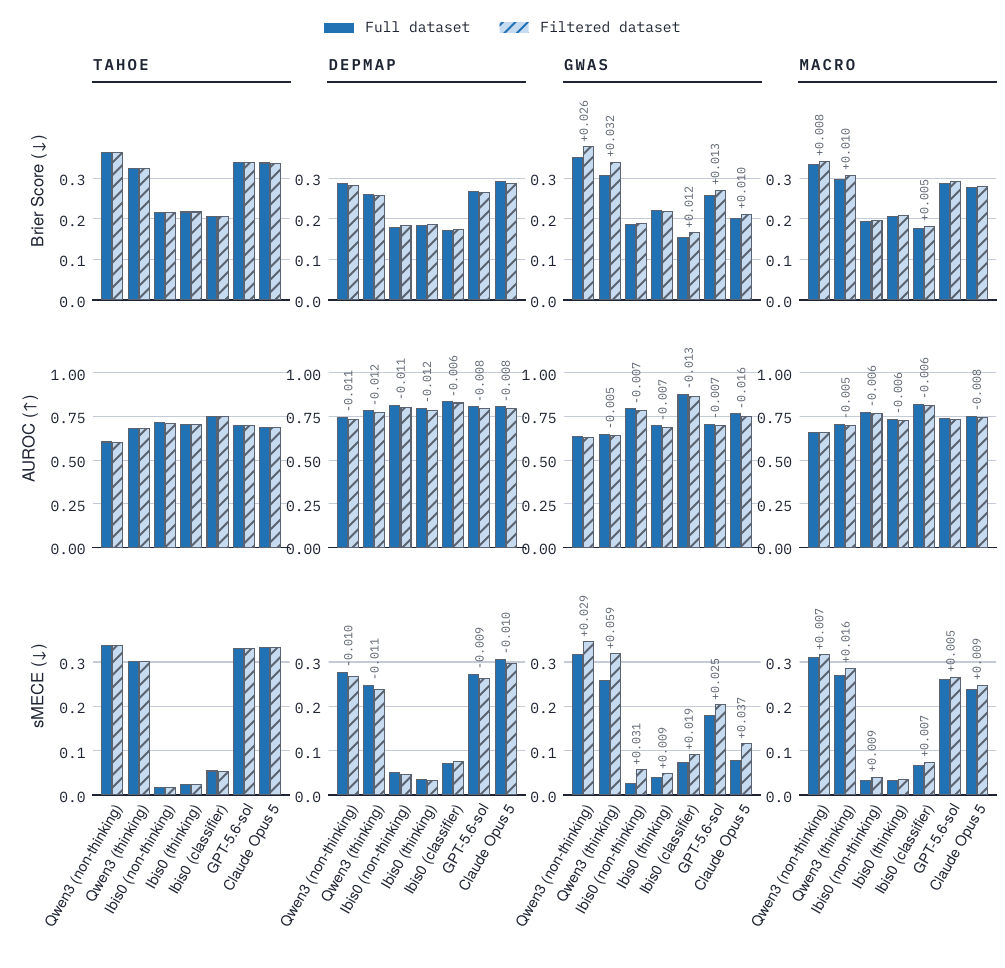}
\caption{\textbf{Metrics with and without the overlapping test episodes.} Solid bars use the full $2{,}000$-episode primary test set in each domain; hatched bars use the filtered dataset, which removes the episodes whose held-out measurement appears in a training or validation pool (\cref{tab:overlap}). Rows are metrics and columns are domains, with the equal-weight macro-average on the right. Labels above a bar pair give the filtered value minus the full value where the difference reaches $0.005$.}
\label{fig:overlap}
\end{figure}

\paragraph{Effect on the reported metrics.}
We recompute Brier score, AUROC, and sMECE on the primary test set with the overlapping episodes removed, for all seven models (\cref{fig:overlap}). GWAS moves the most, as its overlap rate predicts. The absolute changes are small: across the three \modelname{} variants, macro-averaged Brier changes by at most $0.005$ and macro-averaged AUROC by $0.006$. The shifts are similar in sign and size across models, and no ranking changes. The zero-shot models, which were never trained on \ourbenchmark, also score better on the overlapping episodes than on the remainder, and on GWAS their change on removal exceeds that of the \modelname{} RL variants. The overlapping episodes are therefore easier in their own right, and most of the movement is a selection effect rather than memorization. The one pattern consistent with memorization is on DepMap, where Brier score improves on the overlapping subset for the trained models and worsens for the untrained ones; it amounts to about $0.004$ in the full DepMap Brier score.

\paragraph{Addressing the overlap.}
A split grouped by target entity closes the overlap for Tahoe and DepMap, whose pools are gene-scoped, but not for GWAS, whose pools cross gene, SNP, and trait. Alternatively, pools can be built after the split with test measurements excluded, which keeps the in-distribution framing at the cost of a rebuild. We intend to adopt one of these in a future revision.

\FloatBarrier
\renewcommand{\topfraction}{0.7}
\renewcommand{\textfraction}{0.2}
\renewcommand{\floatpagefraction}{0.5}

\section{Model variants and training details}\label{sec:training-details}

\paragraph{Shared setup.} All post-training runs start from \texttt{Qwen3-30B-A3B}. The runs comprise one classifier and two RL policies trained on all three domains and three classifiers trained on leave-one-domain-out mixtures (\cref{tab:runs}). They use the same training split, assign equal weight to each included domain, and render each episode with $K\sim\mathrm{Uniform}\{2,\dots,16\}$ evidence stubs. $K$ is determined from the episode identifier, so a given episode has the same length across runs.

Optimization is AdamW at learning rate $9\times10^{-6}$, weight decay $0.01$ and gradient-norm clipping at $1.0$, with $225$ linear warmup steps and no subsequent decay. These settings are shared by all training mixtures, including the leave-one-out runs.

\paragraph{Checkpoint selection.} Each domain reserves a $256$-episode validation split that is separate from the test episodes and from the equally sized sensitivity subset of the test set. Each run is evaluated on this held-out split every $100$ steps ($200$ for \modelname{} (thinking), whose evaluation dominated its wall-clock time) and stops when its selection metric has gone $1{,}000$ steps without improving (minimum improvement $10^{-4}$ for the classifiers, $10^{-3}$ for RL). No run reached the $9{,}000$-step safety ceiling. Five runs stopped on this patience criterion; the GWAS-held-out classifier terminated on a checkpoint-write error $799$ steps into its patience window, after its selected checkpoint had been written.

Classifier runs are selected by validation cross-entropy and RL runs by mean validation reward over three evaluation environments ($256$ episodes each for \modelname{} (non-thinking), $128$ for \modelname{} (thinking), $4$ rollouts per episode). \Cref{tab:runs} reports the selected checkpoints used in all evaluations.

\begin{table}[htbp]\centering\small
\begin{tabular}{llccc}
\toprule
\textbf{Run} & \textbf{Training domains} & \textbf{Selection metric} & \textbf{Best value} & \textbf{Step} \\
\midrule
\modelname{} (classifier) & Tahoe, DepMap, GWAS & in-domain BCE $\downarrow$ & $0.4785$ & $2400$ \\
\modelname{} (non-thinking) & Tahoe, DepMap, GWAS & eval reward $\uparrow$ & $0.8304$ & $4200$ \\
\modelname{} (thinking) & Tahoe, DepMap, GWAS & eval reward $\uparrow$ & $0.8168$ & $2400$ \\
\midrule
\modelname{} (classifier), Tahoe held out & DepMap, GWAS & in-domain BCE $\downarrow$ & $0.4390$ & $1700$ \\
\modelname{} (classifier), DepMap held out & Tahoe, GWAS & in-domain BCE $\downarrow$ & $0.4794$ & $1300$ \\
\modelname{} (classifier), GWAS held out & Tahoe, DepMap & in-domain BCE $\downarrow$ & $0.5096$ & $1800$ \\
\midrule
Outcome simulator (SFT) & Tahoe, DepMap, GWAS & fixed budget & -- & $1200$ \\
\bottomrule
\end{tabular}
\caption{Training runs and selected checkpoints. All runs start from \texttt{Qwen3-30B-A3B}. The \modelname{} runs use the same optimizer settings and differ in objective and training mixture; the outcome simulator used in \cref{sec:bed} is trained separately (\cref{sec:training-details}). In-domain BCE is averaged over the trained domains only, so it is not comparable across rows with different mixtures.}
\label{tab:runs}
\end{table}

\paragraph{Classifier.} We replace the language-modeling head with a single linear unit ($\mathbb{R}^{d}\!\to\!\mathbb{R}$ for hidden size $d$) on the last-token hidden state, whose sigmoid is the predicted probability of support, and fine-tune the whole transformer with this head (no frozen backbone) under binary cross-entropy against the support label.

All classification runs use a sequence length of $16{,}384$ tokens, which accommodates the longest render in any domain. One episode is processed per data-parallel rank per micro-step, with $16$ gradient-accumulation steps on $8$ ranks, giving $128$ episodes per optimizer step. The classifier produces a continuous scalar forecast (\cref{fig:prob_hist}) rather than a verbalized probability.

\paragraph{Leave-one-out classifiers.} The transfer results in \cref{sec:transfer} come from three additional classification runs that use the same configuration but train on two domains. All three are evaluated on every domain. As for the all-domain classifier, checkpoints are selected by validation cross-entropy, here macro-averaged over the two \emph{training} domains and excluding the omitted one.

\paragraph{RL runs.} The two RL runs use \texttt{prime-rl}~\citep{primerl} under the reward $r=1-(p-y)^2$, where $y\in\{0,1\}$ is the observed binary support label for the target hypothesis (\cref{sec:benchmark}) and $p$ is the probability the model states given the revealed evidence. A completion with no parseable probability receives $r=-1$. Advantages are group-relative: each rollout's reward minus the mean reward of the rollouts sampled for its prompt, with \emph{no} standard-deviation normalization. This is the unbiased, group-mean-baselined advantage of Dr.~GRPO~\citep{drgrpo} rather than the standardized advantage of GRPO. Groups whose advantages are all zero are dropped before the trainer sees them. In place of PPO-style ratio clipping, the policy is updated with \texttt{prime-rl}'s default trust-region objective, the binary total-variation variant of DPPO~\citep{dppo}. It masks a token out of the update when its sampling probability moves by more than $0.2$ in the trust-region-violating direction (conditioned on the advantage sign) and leaves the importance ratio unclipped on the kept tokens. A squared-log-ratio KL penalty (coefficient $10^{-3}$, following Kimi-K2.5~\citep{kimik25}) regularizes the update. The policies are trained, validated and evaluated at sampling temperature $0.7$.

\paragraph{Direct and thinking.} The two RL runs share objective, reward, data and optimizer, and differ in Qwen3's thinking mode. \modelname{} (non-thinking) has thinking disabled and a $512$-token completion budget: the model emits its answer immediately as a \texttt{<probability>$p$</probability>} tag, with no chain of thought. It trains at sequence length $9{,}216$, with $32$ rollouts per optimizer step and $8$ rollouts per example. \modelname{} (thinking) has thinking enabled and a $24{,}576$-token budget, which forces sequence length $32{,}768$; it trains with $16$ rollouts per step and $4$ per example. The reward reads only the stated probability in both cases.

\paragraph{Use in subsequent analyses.} The classifier is included in the weight-of-evidence and dose--response analyses. In the experimental-design evaluation (\cref{sec:bed}), the belief model is the all-domain classifier (step $2400$), and \modelname{} (non-thinking) is the \modelname{} naive-prompting policy.

\paragraph{Outcome simulator.} The simulator $S$ of \cref{sec:bed} is a supervised fine-tune of \texttt{Qwen3-30B-A3B} with thinking disabled (last row of \cref{tab:runs}). Its training data are $180{,}000$ real (experiment, outcome) pairs from the \ourbenchmark{} training split, drawn from GWAS associations, DepMap CRISPR screens and Tahoe evidence stubs and balanced by valence. Each prompt gives an experiment with its result hidden, the hypothesis $\theta'$ that the experiment tests, and a requested valence; the target is the real outcome, labelled supporting or null by the sign of the weight of evidence the classifier assigns it for $\theta'$. No outcome is synthetic. Training uses AdamW at learning rate $10^{-5}$, batch size $64$, sequence length $12{,}288$ and loss on the completion only, for a fixed budget of $1{,}200$ steps (validation loss $0.287$ at step $1{,}100$, the last evaluation). On $100$ held-out test-split experiments per domain, decoded greedily, the supporting outcome carries more weight of evidence for $\theta'$ than the null outcome for the same experiment in $83\%$ of cases.

\FloatBarrier

\section{Evaluation protocol}
\label{app:evalprotocol}

This appendix specifies the estimators, resampling procedures, and decoding rules used for the metrics defined in \cref{sec:metrics}.

\paragraph{Macro-averaging.}
Each aggregate statistic is computed separately within Tahoe, DepMap, and GWAS and then averaged with equal weight. This procedure prevents domain-specific metric distributions from changing the effective domain weights.

If a statistic is undefined for a resampled domain, as can occur for AUROC when a resample contains one class, that domain is omitted from the resampled macro-average. This behavior is implemented for the macro bootstrap and transfer-matrix statistic; the master-table implementation raises an error. No such case occurs in the $2{,}000$-episode point estimates.

\paragraph{Within-domain stratified bootstrap.}
Each bootstrap draw resamples episodes with replacement within each domain, holding the per-domain count fixed, and recomputes the per-domain statistics and their macro-average. For the aggregate results on the primary set (the master table, headline figure, grouped-metric panels, transfer matrix and ROC AUROC table), intervals are the $2.5$th and $97.5$th percentiles over $2{,}000$ draws. Other figures use fewer draws: $1{,}000$ for the dose--response and weight-of-evidence curves, $500$ for the per-bin reliability and ROC bands, and $250$ for the per-position and signed-verdict intervals. Point estimates are computed on the complete evaluation sample.

Resampling within domains keeps domain weights fixed across draws. Pooling episodes before resampling would instead estimate uncertainty for a micro-average with variable domain proportions.

The sMECE estimator is $O(n^2)$ in the number of episodes and costs about $35$ times as much per draw as the other aggregate metrics, so its intervals use $200$ draws. Its point estimate uses the complete sample.

\paragraph{Episode-level resampling.}
Each episode contributes one forecast to the aggregate metrics and up to $16$ updates to the per-stub analyses. Updates from the same episode share a target hypothesis, an evidence pool and an ordering, so they are strongly correlated. The bootstrap for per-stub quantities is therefore \emph{episode-clustered}: it samples complete episodes with all of their stubs, preserving within-episode dependence.

Relevance-hierarchy quantities are macro-averaged over domains using episode clustering. Signed weight-of-evidence results are reported separately by domain because verdict categories differ across domains; their intervals use within-domain episode-clustered bootstraps.

\paragraph{Calibration of weights of evidence.}\label{app:woe-estimator}
This paragraph gives the estimators of the odds ratio of support $\psi$ and the slope $\beta$ (\cref{sec:metrics}), which Analysis~2 of \cref{sec:woe} reports. The unit of analysis is a single belief update. On the sensitivity set each model states a probability $p_k$ for every episode at each evidence count $k\in\{0,\dots,16\}$, and each consecutive pair of forecasts gives one update $(\ell_i,w_i,\theta_i)$: the incoming belief $\ell_i=\operatorname{logit}(p_{k-1})$, the weight of evidence
\[
w_i=\frac{10}{\ln 10}\,\big[\operatorname{logit}(p_k)-\operatorname{logit}(p_{k-1})\big]
\]
in decibans (\cref{sec:benchmark}), and the label $\theta_i\in\{0,1\}$ of the episode, which all of its updates share. Each model contributes $256\times16$ updates per domain. Probabilities are clipped to $[10^{-3},\,1-10^{-3}]$ before the logit. Updates with $w_i=0$, where the stated probability does not change, are excluded. They range from $1\%$ of updates for \modelname{} (classifier) to $68\%$ for \modelname{} (non-thinking); models that state probabilities in text often repeat the same value (\cref{sec:primary}).

\textit{Tiers and belief strata.} Separately for each model and domain, the remaining updates are cut into five quantile bins of $w$, the \emph{tiers}, and five quantile bins of $\ell$, the \emph{belief strata}; tied quantile edges are merged. The \emph{reference tier} is the tier whose mean $w$, averaged over the three domains, is closest to zero. It is chosen once per model and shared by all three domains.

\textit{Odds ratio of support.} For tier $t$ and belief stratum $s$, let $a_s$ and $b_s$ count the supported ($\theta=1$) and unsupported ($\theta=0$) updates in tier $t$, let $c_s$ and $d_s$ count the same for the reference tier, and let $n_s=a_s+b_s+c_s+d_s$. The odds ratio of support of tier $t$ is the Mantel--Haenszel estimate pooled over belief strata~\citep{mantel1959statistical}, in decibans:
\[
\psi_t=\frac{10}{\ln 10}\,\ln\frac{\sum_s a_s d_s/n_s}{\sum_s b_s c_s/n_s}.
\]
This odds ratio compares the odds that the hypothesis is supported after an update in tier $t$ with the same odds after a reference update from the same incoming belief. Pooling within belief strata makes $\psi$ independent of where the belief started. When either sum is zero, as happens when a tier separates the labels perfectly, $0.5$ is added to every cell before pooling. Each tier is plotted at $x_t=\bar w_t-\bar w_{\mathrm{ref}}$, its mean weight of evidence relative to the reference tier, so the reference tier sits at the origin with $\psi=0$. The points are then averaged over the three domains by tier rank.

\textit{Slope.} If a model's probabilities are calibrated, $\psi_t=x_t$ for every tier, up to sampling error. We summarize the points by their least-squares slope through the origin,
\[
\beta=\frac{\sum_t x_t\,\psi_t}{\sum_t x_t^2}=\frac{\sum_t x_t^2\,(\psi_t/x_t)}{\sum_t x_t^2},
\]
where the sums run over the non-reference tiers. The slope is a weighted average of the per-tier ratios $\psi_t/x_t$ with weights $x_t^2$, so the tiers with the largest updates count most. A value of $\beta=1$ indicates calibrated updates, $\beta<1$ indicates overreaction (updates larger than the outcomes justify), and $\beta>1$ indicates underreaction (updates smaller than the outcomes justify). The quantity $\beta-1$ is the log-odds analog of the coefficient in a regression of forecast errors on forecast revisions, which is negative when forecasters overreact to news and positive when they underreact~\citep{coibion2015information,bordalo2020overreaction}.

\textit{Uncertainty.} Each episode contributes up to $16$ updates that share its label, so the updates are not independent. Intervals come from the episode-clustered bootstrap described above: each draw resamples episodes within each domain, holds the bin edges and the reference tier fixed, and recomputes the points and the slope.

\textit{Sampling noise.} Each $p_k$ of a generative model is a single sample (see \emph{Decoding} below), so part of the variation in $w$ is sampling noise that is unrelated to the outcome. Binning on a noisy $w$ biases $\beta$ toward zero and therefore overstates overreaction for these models. The classifier is deterministic and is unaffected.

\paragraph{Stub verdicts.}\label{app:verdict}
Each evidence stub is assigned a verdict relative to the episode's target hypothesis $h^*$, judged against $h^*$'s claim rather than the stub's own effect. In \textbf{Tahoe} the verdict is directional: a stub is \emph{supported} when its differential-expression measurement is significant ($\mathrm{padj}<0.05$, $|\log_2\mathrm{FC}|>0.5$) in the direction $h^*$ claims, \emph{refuted} when significant in the opposite direction, and \emph{inconclusive} otherwise. In \textbf{DepMap}, where $h^*$ (``$G$ is a dependency in $C$'') is directionless, a stub is \emph{supported} when it indicates a dependency ($\mathrm{Chronos}\le-0.5$, $P(\mathrm{dep})\ge0.5$), \emph{refuted} when it indicates a suppressor ($\mathrm{Chronos}>0.25$), and \emph{inconclusive} otherwise. \textbf{GWAS} has no refuting category: a stub is \emph{significant} when $p<5\times10^{-8}$ and \emph{sub-threshold} otherwise. Because effect-direction concordance across different variants and traits is not biologically guaranteed, a direction-flipped association is not treated as refuting.

\paragraph{Paired comparisons.}
Comparisons reported as differences use a paired bootstrap over models evaluated on identical episodes. Each iteration applies one episode resample to every model in the comparison. Leave-one-out deltas and retention intervals use this procedure. Intervals for individual rows in the master table are unpaired.

\paragraph{Calibration: sMECE in place of ECE.}
Expected calibration error (ECE)~\citep{guo2017calibration} averages the absolute difference between each bin's mean forecast and outcome rate, weighted by bin occupancy. Our implementation computes ECE over the same ten equal-width bins used for the Murphy terms, but the reported calibration statistic is sMECE because ECE depends on the number and placement of bin edges.

For sMECE, we estimate the empirical frequency $r(p)=\mathbb{E}[y\mid p]$ by Nadaraya--Watson regression of the outcomes on the forecasts,
\[
  \hat r(p_i) \;=\; \frac{\sum_j K(p_i,p_j)\,y_j}{\sum_j K(p_i,p_j)},
  \qquad
  K(p_i,p_j) \;=\; \phi\!\left(\tfrac{p_i-p_j}{h}\right)
                 + \phi\!\left(\tfrac{p_i+p_j}{h}\right)
                 + \phi\!\left(\tfrac{2-p_i-p_j}{h}\right),
\]
with $\phi$ the standard Gaussian density, and summarize the fit as the mean of $|\hat r(p_i)-p_i|$ over the evaluated episodes. Averaging at the observed forecasts weights the error by forecast density.

The second and third kernel terms reflect the Gaussian across $0$ and $1$ to preserve boundary mass. The bandwidth follows Silverman's rule on the forecasts, $h=1.06\,\hat\sigma n^{-1/5}$, with a minimum value of $0.02$ to prevent a nearly degenerate forecast distribution from producing a near-zero bandwidth.

This is an estimator in the smooth-ECE family introduced by \citet{blasiok2024smooth}, not their estimator: the bandwidth rule and the boundary handling are ours. We therefore refer to the statistic as sMECE.

\paragraph{Murphy decomposition.}
The Murphy decomposition separates the Brier score into reliability, resolution, and uncertainty terms~\citep{murphy1973new,murphy1977reliability,degroot1983comparison}:
\begin{equation}
\label{eq:murphy}
  \mathrm{Brier\ Score} \;=\; \frac1N\sum_i (p_i-\theta_i)^2 \;=\;
  \underbrace{\sum_k w_k(\bar p_k-\bar\theta_k)^2}_{\text{Reliability}} \;-\;
  \underbrace{\sum_k w_k(\bar\theta_k-\bar\theta)^2}_{\text{Resolution}} \;+\;
  \underbrace{\bar\theta(1-\bar\theta)}_{\text{Uncertainty}},
\end{equation}
where $k$ indexes ten equal-width bins, $w_k$ is the bin's share of episodes, $\bar p_k$ and $\bar\theta_k$ are its mean forecast and observed outcome rate, and $\bar\theta$ is the base rate. Bins are right-closed, with the first bin including $0$.

Reliability measures within-bin calibration error and is lower when mean forecasts match observed frequencies. Resolution measures variation in observed outcome rates across forecast bins and is higher when forecasts separate outcomes. Uncertainty depends only on the label distribution and is identical across models evaluated on a fixed test set; for the balanced test set it is $0.25$.

The identity is exact for binned forecasts. The tables report unbinned mean squared error, so the right-hand side of \cref{eq:murphy} differs from the reported Brier score by within-bin dispersion. We use the decomposition to interpret differences among models.

\paragraph{Reading intervals near zero.}
Reliability and sMECE are non-negative and positively biased: resampling injects noise that a squared or absolute deviation can only turn into a larger value. For a row whose true calibration error is close to zero, the bootstrap distribution can lie above the point estimate, leaving the point estimate outside its own interval. Such rows are flagged in the tables. These intervals support comparisons among rows but should not be interpreted as coverage intervals for a near-zero point estimate.

\paragraph{Decoding.}
The generative rows are sampled with a fixed seed at temperature $0.7$, at which the RL policies are also trained and validated (\cref{sec:training-details}). On the primary set each episode is sampled four times and the reported probability is the mean of parsed draws. On the sensitivity set, where every episode is rendered at all $17$ context lengths, each episode is drawn once. Both estimators target the same quantity; the four-draw mean has lower sampling variance. The classifier is deterministic: its probability is the sigmoid of a scalar head applied to the last-token representation. Frontier models use provider-default sampling settings as described in \cref{app:frontier}. All evaluated completions produced a readable probability, and no episodes are excluded.

\paragraph{Complete results table.}
\Cref{tab:multidomain-master-full} reports the macro-average for every model with its per-domain breakdown, the number of episodes scored, and a bootstrap interval on every value.

\clearpage
\providecommand{\mtstack}[2]{\begin{tabular}[t]{@{}r@{}}#1\\[-2pt]{\scriptsize\color{black!45}#2}\end{tabular}}
\begin{table}[t]
\centering
\footnotesize
\setlength{\tabcolsep}{3pt}
\renewcommand{\arraystretch}{0.9}
\begin{tabular}{lrrrrrr}
\toprule
Model & $n$ & Brier Score ($\downarrow$) & sMECE ($\downarrow$) & AUROC ($\uparrow$) & Reliability ($\downarrow$) & Resolution ($\uparrow$) \\
\midrule
Qwen3 (non-thinking) & 6,000 & \mtstack{0.3336}{[0.3237, 0.3436]} & \mtstack{0.3101}{[0.2990, 0.3229]} & \mtstack{0.6608}{[0.6471, 0.6752]} & \mtstack{0.1096}{[0.1024, 0.1189]} & \mtstack{0.0250}{[0.0226, 0.0295]} \\
\quad Tahoe & 2,000 & \mtstack{0.3634}{[0.3459, 0.3827]} & \mtstack{0.3367}{[0.3201, 0.3589]} & \mtstack{0.6039}{[0.5805, 0.6279]} & \mtstack{0.1277}{[0.1140, 0.1452]} & \mtstack{0.0147}{[0.0111, 0.0204]} \\
\quad DepMap & 2,000 & \mtstack{0.2860}{[0.2715, 0.2995]} & \mtstack{0.2773}{[0.2575, 0.2978]} & \mtstack{0.7454}{[0.7239, 0.7671]} & \mtstack{0.0854}{[0.0754, 0.0977]} & \mtstack{0.0485}{[0.0414, 0.0577]} \\
\quad GWAS & 2,000 & \mtstack{0.3513}{[0.3333, 0.3694]} & \mtstack{0.3162}{[0.2966, 0.3391]} & \mtstack{0.6333}{[0.6080, 0.6571]} & \mtstack{0.1156}{[0.1022, 0.1314]} & \mtstack{0.0120}{[0.0089, 0.0176]} \\
\addlinespace
Qwen3 (thinking) & 6,000 & \mtstack{0.2973}{[0.2896, 0.3047]} & \mtstack{0.2696}{[0.2607, 0.2818]} & \mtstack{0.7040}{[0.6914, 0.7170]} & \mtstack{0.0829}{[0.0772, 0.0907]} & \mtstack{0.0339}{[0.0307, 0.0387]} \\
\quad Tahoe & 2,000 & \mtstack{0.3235}{[0.3103, 0.3373]} & \mtstack{0.3014}{[0.2770, 0.3244]} & \mtstack{0.6822}{[0.6586, 0.7049]} & \mtstack{0.0988}{[0.0877, 0.1131]} & \mtstack{0.0224}{[0.0172, 0.0288]} \\
\quad DepMap & 2,000 & \mtstack{0.2611}{[0.2492, 0.2725]} & \mtstack{0.2481}{[0.2301, 0.2678]} & \mtstack{0.7824}{[0.7630, 0.8029]} & \mtstack{0.0751}{[0.0659, 0.0859]} & \mtstack{0.0626}{[0.0546, 0.0722]} \\
\quad GWAS & 2,000 & \mtstack{0.3074}{[0.2933, 0.3213]} & \mtstack{0.2593}{[0.2385, 0.2822]} & \mtstack{0.6474}{[0.6228, 0.6721]} & \mtstack{0.0748}{[0.0644, 0.0877]} & \mtstack{0.0165}{[0.0124, 0.0231]} \\
\addlinespace
\modelname{} (non-thinking) & 6,000 & \mtstack{0.1933}{[0.1888, 0.1977]} & \mtstack{0.0316}{[0.0258, 0.0450]} & \mtstack{0.7729}{[0.7616, 0.7846]} & \mtstack{0.0020}{[0.0017, 0.0038]} & \mtstack{0.0579}{[0.0540, 0.0631]} \\
\quad Tahoe & 2,000 & \mtstack{0.2149}{[0.2083, 0.2215]} & \mtstack{0.0174}{[0.0141, 0.0389]} & \mtstack{0.7128}{[0.6902, 0.7348]} & \mtstack{0.0007}{[0.0006, 0.0028]} & \mtstack{0.0347}{[0.0289, 0.0424]} \\
\quad DepMap & 2,000 & \mtstack{0.1787}{[0.1693, 0.1881]} & \mtstack{0.0517}{[0.0340, 0.0677]} & \mtstack{0.8125}{[0.7931, 0.8314]} & \mtstack{0.0042}{[0.0026, 0.0078]} & \mtstack{0.0747}{[0.0663, 0.0845]} \\
\quad GWAS & 2,000 & \mtstack{0.1861}{[0.1781, 0.1944]} & \mtstack{0.0258}{[0.0155, 0.0431]} & \mtstack{0.7934}{[0.7731, 0.8127]} & \mtstack{0.0010}{[0.0006, 0.0035]} & \mtstack{0.0644}{[0.0559, 0.0739]} \\
\addlinespace
\modelname{} (thinking) & 6,000 & \mtstack{0.2069}{[0.2025, 0.2111]} & \mtstack{0.0327}{[0.0296, 0.0431]} & \mtstack{0.7329}{[0.7209, 0.7450]} & \mtstack{0.0023}{[0.0021, 0.0044]} & \mtstack{0.0448}{[0.0413, 0.0500]} \\
\quad Tahoe & 2,000 & \mtstack{0.2169}{[0.2102, 0.2235]} & \mtstack{0.0238}{[0.0172, 0.0456]} & \mtstack{0.7064}{[0.6839, 0.7286]} & \mtstack{0.0016}{[0.0009, 0.0042]} & \mtstack{0.0347}{[0.0287, 0.0424]} \\
\quad DepMap & 2,000 & \mtstack{0.1828}{[0.1747, 0.1907]} & \mtstack{0.0348}{[0.0287, 0.0522]} & \mtstack{0.7961}{[0.7765, 0.8160]} & \mtstack{0.0028}{[0.0020, 0.0055]} & \mtstack{0.0689}{[0.0607, 0.0787]} \\
\quad GWAS & 2,000 & \mtstack{0.2209}{[0.2128, 0.2289]} & \mtstack{0.0394}{[0.0314, 0.0558]} & \mtstack{0.6961}{[0.6730, 0.7198]} & \mtstack{0.0024}{[0.0015, 0.0059]} & \mtstack{0.0307}{[0.0249, 0.0392]} \\
\addlinespace
\modelname{} (classifier) & 6,000 & \mtstack{0.1771}{[0.1707, 0.1831]} & \mtstack{0.0658}{[0.0588, 0.0762]} & \mtstack{0.8195}{[0.8099, 0.8295]} & \mtstack{0.0075}{[0.0065, 0.0104]} & \mtstack{0.0790}{[0.0753, 0.0849]} \\
\quad Tahoe & 2,000 & \mtstack{0.2063}{[0.1971, 0.2157]} & \mtstack{0.0541}{[0.0389, 0.0707]} & \mtstack{0.7485}{[0.7275, 0.7683]} & \mtstack{0.0042}{[0.0028, 0.0082]} & \mtstack{0.0470}{[0.0407, 0.0556]} \\
\quad DepMap & 2,000 & \mtstack{0.1713}{[0.1602, 0.1821]} & \mtstack{0.0710}{[0.0589, 0.0886]} & \mtstack{0.8334}{[0.8163, 0.8500]} & \mtstack{0.0076}{[0.0054, 0.0124]} & \mtstack{0.0855}{[0.0780, 0.0952]} \\
\quad GWAS & 2,000 & \mtstack{0.1537}{[0.1416, 0.1654]} & \mtstack{0.0724}{[0.0584, 0.0882]} & \mtstack{0.8766}{[0.8617, 0.8915]} & \mtstack{0.0107}{[0.0078, 0.0160]} & \mtstack{0.1046}{[0.0964, 0.1153]} \\
\addlinespace
GPT-5.6-sol & 6,000 & \mtstack{0.2879}{[0.2793, 0.2964]} & \mtstack{0.2602}{[0.2490, 0.2712]} & \mtstack{0.7365}{[0.7235, 0.7485]} & \mtstack{0.0830}{[0.0775, 0.0908]} & \mtstack{0.0430}{[0.0395, 0.0484]} \\
\quad Tahoe & 2,000 & \mtstack{0.3385}{[0.3228, 0.3549]} & \mtstack{0.3295}{[0.3095, 0.3479]} & \mtstack{0.6987}{[0.6769, 0.7208]} & \mtstack{0.1204}{[0.1072, 0.1360]} & \mtstack{0.0304}{[0.0247, 0.0377]} \\
\quad DepMap & 2,000 & \mtstack{0.2683}{[0.2547, 0.2829]} & \mtstack{0.2723}{[0.2533, 0.2888]} & \mtstack{0.8046}{[0.7846, 0.8237]} & \mtstack{0.0880}{[0.0773, 0.1003]} & \mtstack{0.0665}{[0.0581, 0.0765]} \\
\quad GWAS & 2,000 & \mtstack{0.2570}{[0.2426, 0.2720]} & \mtstack{0.1788}{[0.1623, 0.2005]} & \mtstack{0.7061}{[0.6828, 0.7287]} & \mtstack{0.0408}{[0.0339, 0.0509]} & \mtstack{0.0321}{[0.0261, 0.0407]} \\
\addlinespace
Claude Opus 5 & 6,000 & \mtstack{0.2772}{[0.2689, 0.2851]} & \mtstack{0.2390}{[0.2291, 0.2505]} & \mtstack{0.7519}{[0.7392, 0.7646]} & \mtstack{0.0786}{[0.0730, 0.0857]} & \mtstack{0.0502}{[0.0461, 0.0556]} \\
\quad Tahoe & 2,000 & \mtstack{0.3376}{[0.3229, 0.3525]} & \mtstack{0.3320}{[0.3114, 0.3498]} & \mtstack{0.6863}{[0.6640, 0.7082]} & \mtstack{0.1142}{[0.1017, 0.1291]} & \mtstack{0.0254}{[0.0205, 0.0320]} \\
\quad DepMap & 2,000 & \mtstack{0.2923}{[0.2776, 0.3068]} & \mtstack{0.3065}{[0.2880, 0.3244]} & \mtstack{0.8048}{[0.7848, 0.8240]} & \mtstack{0.1132}{[0.1010, 0.1269]} & \mtstack{0.0683}{[0.0598, 0.0780]} \\
\quad GWAS & 2,000 & \mtstack{0.2016}{[0.1907, 0.2124]} & \mtstack{0.0785}{[0.0646, 0.0951]} & \mtstack{0.7647}{[0.7435, 0.7862]} & \mtstack{0.0083}{[0.0059, 0.0136]} & \mtstack{0.0569}{[0.0495, 0.0669]} \\
\midrule
constant p = base rate & 6,000 & \mtstack{0.2500}{[0.2500, 0.2500]} & \mtstack{0.0000$^{\dagger}$}{[0.0027, 0.0187]} & \mtstack{0.5000}{[0.5000, 0.5000]} & \mtstack{0.0000$^{\dagger}$}{[0.0000, 0.0004]} & \mtstack{0.0000}{[0.0000, 0.0000]} \\
\quad Tahoe & 2,000 & \mtstack{0.2500}{[0.2500, 0.2500]} & \mtstack{0.0000}{[0.0000, 0.0250]} & \mtstack{0.5000}{[0.5000, 0.5000]} & \mtstack{0.0000$^{\dagger}$}{[0.0000, 0.0006]} & \mtstack{0.0000}{[0.0000, 0.0000]} \\
\quad DepMap & 2,000 & \mtstack{0.2500}{[0.2500, 0.2500]} & \mtstack{0.0000$^{\dagger}$}{[0.0005, 0.0276]} & \mtstack{0.5000}{[0.5000, 0.5000]} & \mtstack{0.0000$^{\dagger}$}{[0.0000, 0.0006]} & \mtstack{0.0000}{[0.0000, 0.0000]} \\
\quad GWAS & 2,000 & \mtstack{0.2500}{[0.2500, 0.2500]} & \mtstack{0.0000$^{\dagger}$}{[0.0005, 0.0265]} & \mtstack{0.5000}{[0.5000, 0.5000]} & \mtstack{0.0000$^{\dagger}$}{[0.0000, 0.0007]} & \mtstack{0.0000}{[0.0000, 0.0000]} \\
\bottomrule
\end{tabular}
\\[2pt]
\parbox{\linewidth}{\footnotesize\raggedright Macro-average over Tahoe, DepMap, GWAS; 95\% CI from a within-domain bootstrap resampling episodes (2,000 draws; sMECE 200, it is ~35x costlier per draw). The unindented row of each block is that macro-average; the indented rows below it are the single domains, bootstrapped over that domain's episodes alone. The constant row ignores all evidence; the test set is balanced ~50:50 by construction, so it scores $\approx$0.25 Brier. Rows above that line do not beat guessing. $\dagger$ point estimate outside its bootstrap interval, as expected for the positively-biased binned metrics (Reliability, sMECE) on rows near the zero floor. Compare rows by these intervals; do not read them as covering the point.}
\caption{Calibration and discrimination on the held-out multi-domain test set, in full.}
\label{tab:multidomain-master-full}
\end{table}

\FloatBarrier

\section{Frontier baseline settings}
\label{app:frontier}

The frontier baselines, GPT-5.6-sol (OpenAI, API model identifier \texttt{gpt-5.6-sol}) and Claude Opus 5 (Anthropic, \texttt{claude-opus-5}), are evaluated on the same episodes as the other models, with reasoning effort and output budget set explicitly for each provider. Reasoning effort is \texttt{high} for both, mapped to \texttt{output\_config.effort} on Anthropic and \texttt{reasoning\_effort} on OpenAI. The output budget is $24{,}000$ tokens for both, mapped to \texttt{max\_tokens} and \texttt{max\_completion\_tokens}, respectively; the local thinking-mode models use the same budget. On Anthropic, this limit includes both reasoning and response tokens, and truncation before the final response produces an unreadable probability (\cref{app:evalprotocol}).

The local generative models are decoded at temperature $0.7$ with a fixed seed, matching their training and validation temperature (\cref{sec:training-details}). The classifier does not sample: a scalar head reads its probability off the last-token representation. Claude Opus 5 rejects temperature, top-$p$ and top-$k$, so the frontier rows use provider-default sampling. Both frontier models receive the system prompt of the local thinking-mode models.

Both models were queried on 29 July 2026 for the test sets and on 4 August 2026 for the conflict and position probes. Declines and API errors are recorded separately from unreadable completions, so policy refusals are not counted as parse failures. Remaining per-provider API details are in the released evaluation code.

\FloatBarrier

\section{Evidence accumulation and per-update belief revision}
\label{app:dose}

This appendix reports the evidence-accumulation and per-stub analyses summarized in \Cref{sec:woe}. All use the sensitivity set, which presents each of $256$ episodes per domain at every evidence count $k\in\{0,\dots,16\}$.

\subsection{Cumulative evidence accumulation}
\label{app:dose-curves}

\begin{figure}[tbp]\centering
\includegraphics[width=\textwidth]{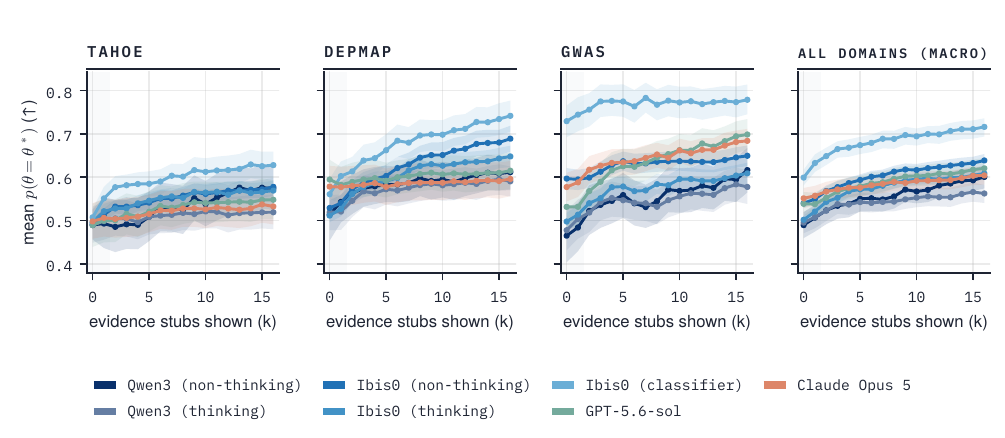}
\caption{\textbf{Confidence in the true label increases as evidence accumulates.} Mean probability each model assigns to the true label vs. the number of evidence stubs shown ($256$ episodes per domain). The slope reflects integration of new evidence (up to $k=16$ stubs; shading marks $k<2$, below the range seen in training). The intercept reflects prior knowledge: $p>0.5$ at $k=0$ means that the model predicts the target hypothesis \textit{with no evidence}. Above-chance \textit{a priori} predictions on GWAS, most notably by \modelname{} (classifier), reflect non-random association of alleles in high linkage disequilibrium in \ourbenchmark{} (see \cref{app:gwas}).}
\label{fig:dose}
\end{figure}

\Cref{fig:dose} reports the mean probability assigned to the true label (supported or not supported) as evidence accumulates,
\[
    p^{\text{true}}_k = \theta p_k + (1 - \theta) (1 - p_k).
\]
Every model assigns more probability to the true label at $k=16$ than at $k=0$ (\cref{fig:dose}); the paired gain excludes zero in $19$ of the $21$ model--domain curves. The exceptions are the two frontier models on DepMap, whose mean changes are small relative to their uncertainty. Starting points vary widely: \modelname{} (classifier) begins at $0.729$ on GWAS before seeing any evidence and gains only $0.050$, whereas on DepMap it begins near $0.5$ and gains $0.181$.

\begin{figure}[tbp]\centering
\includegraphics[width=\textwidth]{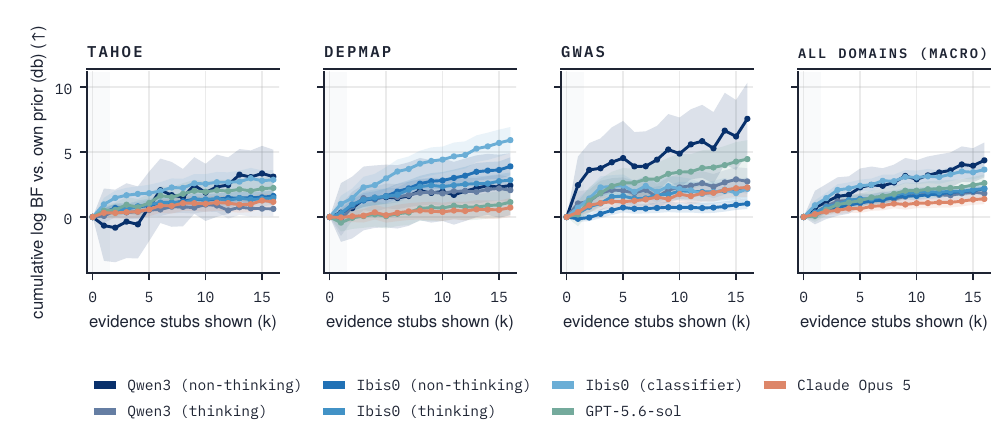}
\caption{\textbf{Cumulative evidence-induced log-odds change by domain.} Cumulative log Bayes factor toward the observed outcome, $s\,[\operatorname{logit}(p_k)-\operatorname{logit}(p_0)]$ in decibans, against the number of stubs shown. Each curve is relative to that model's own $k=0$ forecast. Panels show the three domains and their macro-average on a shared $y$ axis; bands are $95\%$ intervals from within-domain episode resampling.}
\label{fig:logbf}
\end{figure}

This value depends on the prior $p_0$, the model's belief in the hypothesis at the start of the episode, and several models in \cref{fig:dose} start at $p^{\text{true}}_0 > 0.5$. We therefore also measure how far the evidence moves each model from its $k{=}0$ belief. The \textit{cumulative weight of evidence} is the running sum of the per-stub updates $w_j$ (\cref{sec:benchmark}) and, for an ideal Bayesian, the log Bayes factor of the revealed evidence. We orient it toward the ground truth:
\[
    w_k^{\text{cum}} = s\,[\operatorname{logit}(p_k) - \operatorname{logit}(p_0)] \cdot \frac{10}{\log 10}, \quad
    s = \begin{cases}
    +1, & \theta = 1\\
    -1, & \theta = 0
    \end{cases}
\]
so that $w_k^{\text{cum}}$, in decibans, is positive when evidence has pushed belief toward the true outcome (\cref{fig:logbf}). The metric is \textit{prior-agnostic}: it is zero at $k{=}0$ by construction and measures how far a model updates, independent of its starting belief. It is \textit{cumulative}: differencing against step $k-1$ gives the marginal counterpart, $\log BF_k - \log BF_{k-1}$. It is also \textit{direction-independent}: the per-episode sign $s$ swaps the numerator and denominator of $BF_k$ on episodes that are not supported, so a positive value means that the stubs moved belief in the correct direction whether or not the target hypothesis is supported.

By $k{=}16$, models differ roughly threefold in how far the evidence moves them: Qwen3 (non-thinking) moves farthest toward the truth ($+4.37$~dB $[+3.05, +5.74]$), then \modelname{} (classifier) ($+3.65$ $[+3.14, +4.17]$), while Claude Opus 5 moves least ($+1.39$ $[+0.99, +1.76]$). The most responsive model, Qwen3 (non-thinking), also has the worst Brier score. Movement this large can reflect strong response to useful evidence or indiscriminate overreaction; the per-stub analyses in \cref{app:woe-direction,app:woe-tier} and the calibration of weights of evidence in \cref{sec:woe-reliability} separate the two.

\FloatBarrier
\subsection{Direction of belief revision}
\label{app:woe-direction}

\paragraph{Direction of belief revision.} We test whether models revise belief in the expected direction on evidence that plausibly supports or refutes the target hypothesis $h^*$. In each domain, stubs are sorted into heuristic \emph{supporting} and \emph{refuting} categories by their direction relative to $h^*$. For example, when $h^*$ says drug $D$ up-regulates gene $G$, a related drug that significantly up-regulates $G$ counts as supporting and one that significantly \emph{down}-regulates it as refuting; a stub with no statistically significant effect is \emph{inconclusive}. Full per-domain definitions are given in \Cref{app:verdict}.

\begin{figure}[tbp]\centering
\includegraphics[width=\textwidth]{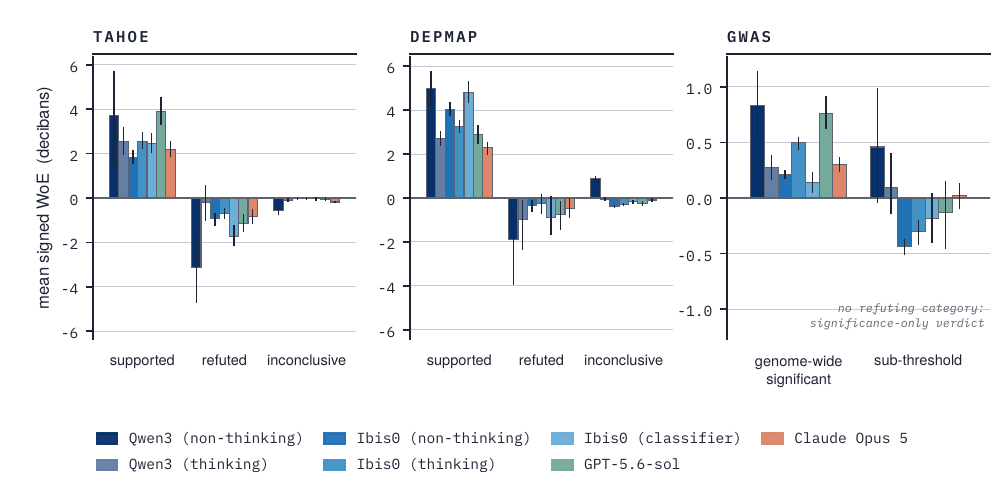}
\caption{\textbf{Do models distinguish supporting vs. refuting evidence?} Mean signed $w_j$ in decibans; positive raises belief in the target hypothesis and negative lowers it. Every model's mean update has the expected sign for supporting and refuting evidence, and most models place little weight on inconclusive evidence. (GWAS admits no refuting evidence (\cref{app:principles}), so its panel contrasts genome-wide-significant with sub-threshold associations.)}
\label{fig:woe_directional}
\end{figure}

Belief revisions have the expected signs (\cref{fig:woe_directional}): every model's mean update is positive on supporting stubs, negative on refuting stubs, and small on inconclusive ones. Magnitudes differ substantially across models. Because all models see the same stubs in the same order, the difference reflects how strongly each model weighs the same category of evidence. Qwen3 (non-thinking) produces the largest updates in both directions, and \cref{sec:woe-reliability} shows that they overreact; its reinforcement-learned counterpart \modelname{} (non-thinking) responds more selectively (\cref{app:woe-tier}).

\FloatBarrier
\subsection{Evidence by relevance tier}
\label{app:woe-tier}

\begin{figure}[tbp]\centering
\includegraphics[width=\textwidth]{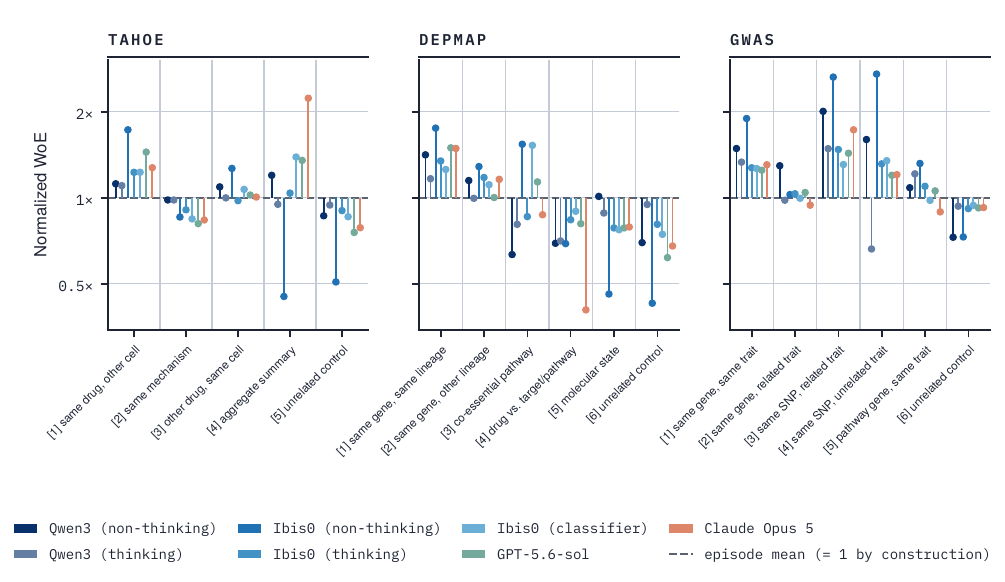}
\caption{\textbf{Do models assign higher weight to more relevant evidence in an expert biological knowledge hierarchy?} Each stem shows how strongly a tier of evidence moves a model, relative to that model's average stub in the same episode ($1\times$ is average). Tiers run left to right in the expert-ranked order, so a model that follows the hierarchy shows heights falling from left to right. On DepMap, the models largely follow the expert ordering. On GWAS, models assign more weight to indirect evidence (same SNP) than to more direct evidence (same gene, same trait). On Tahoe, Claude gives disproportionately high weight to aggregate summary evidence over more-specific assay readouts.}
\label{fig:woe_staircase}
\end{figure}

\paragraph{Evidence by relevance tier.} A model that deliberates well should reserve its largest updates for the evidence that bears most directly on the target. \Cref{fig:woe_staircase} plots each model's mean $|w_j|$ by relevance tier and summarizes its alignment with the curated hierarchy (\cref{app:principles}) as a per-domain Spearman correlation $\rho$ between $|w_j|$ and tier rank (positive when more-relevant stubs draw larger updates). Among the trained models, \modelname{} (non-thinking) aligns most closely in every domain ($\rho=0.22$, $0.37$, $0.18$ on Tahoe, DepMap, and GWAS), despite trailing the classifier on final Brier and AUROC (the classifier's correlations are only $0.09$, $0.19$, $0.09$). Aggregate forecast quality and the structure of individual updates therefore capture different behaviors. Selectivity is also distinct from responsiveness, the overall magnitude of belief change: Qwen3 (non-thinking) produces large updates on average yet aligns only weakly with the hierarchy ($\rho=0.07$ on Tahoe).

The cross-model ordering in \cref{fig:woe_staircase} also tests the hierarchy itself. Agreement is strongest on DepMap. On GWAS, models over-weight indirect same-SNP evidence relative to the curated top tier, and on Tahoe, Claude Opus 5 places disproportionate weight on aggregate summary evidence over more-specific assay readouts. These discrepancies may reflect model failures, limitations of the expert ordering, or both.

\FloatBarrier

\section{Full cross-domain transfer results}
\label{app:transfer}

Raw performance on the omitted domain is hard to interpret because Qwen3 and the all-domain classifier set a different floor and ceiling for each metric. We therefore normalize the leave-one-domain-out result between these two references. For domain $d$ and metric $m$, we report
\begin{equation}
\label{eq:retention}
  \text{retention} \;=\; \frac{m_{\text{leave-out-}d} - m_{\text{untrained}}}{m_\text{all domains} - {m_{\text{untrained}}}}
  \qquad \text{on domain } d,
\end{equation}
where $100\%$ denotes the full improvement from the untrained model (Qwen3) to the all-domain classifier and $0\%$ denotes no improvement over Qwen3. For higher-is-better metrics, both differences are reversed. Paired bootstrap intervals use identical episodes for all three models.

\begin{figure}[tbp]\centering
\includegraphics[width=\textwidth]{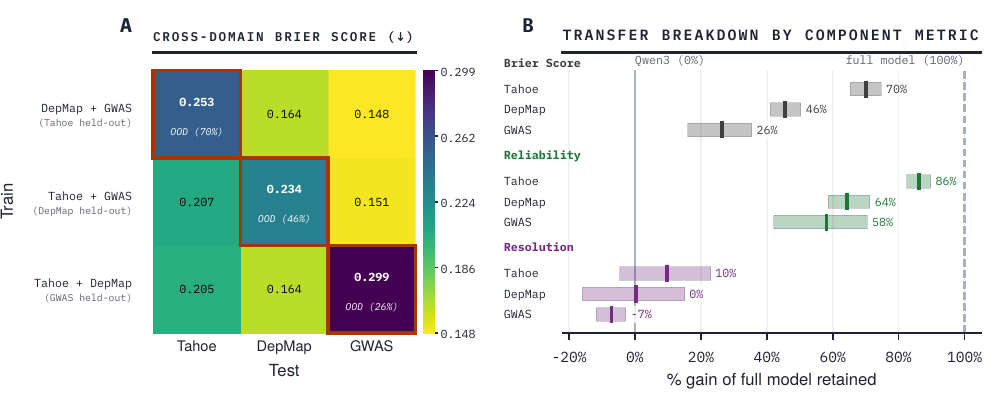}
\caption{\textbf{Transfer to domains excluded from training.} \textbf{A:} We train \modelname{} (classifier) on two domains (rows) and evaluate on a third held-out domain (columns). Outlined diagonal cells are o.o.d.\ evaluations on the excluded domain and report the raw Brier score and the retention percentage (\cref{eq:retention}). \textbf{B:} Retention on the excluded domains, decomposed into Brier score and its Murphy reliability and resolution terms. Boxes show paired-bootstrap $95\%$ intervals and internal lines show point estimates; vertical lines mark Qwen3 ($0\%$) and the all-domain classifier ($100\%$). Reliability retains $58$--$86\%$ of the full-model improvement, whereas effectively no resolution improvement is retained.
}
\label{fig:retention}
\end{figure}

\Cref{tab:transfer-full} reports the complete numerical results underlying \Cref{fig:retention} and \cref{sec:transfer}. Block A adds the all-domain reference and macro-averages omitted from the main figure. Block B reports the held-out-domain decomposition in absolute units, as a paired loss relative to the all-domain classifier, and on the normalized retention scale.

\begin{table}[htbp]
\centering
\scriptsize
\setlength{\tabcolsep}{3pt}
\renewcommand{\arraystretch}{1.08}
\caption{\textbf{Full cross-domain transfer results.} \textbf{A:} Brier score for every classifier training mix and test domain, with $95\%$ bootstrap intervals. Bold cells evaluate a domain excluded from training; other per-domain cells are descriptive in-domain evaluations. Macro is the equal-domain average and is bootstrapped directly as a stratified macro statistic. \textbf{B:} Exact results on the excluded domain for Brier score and its Murphy reliability and resolution terms. The Qwen3 and full-model references denote the released checkpoint and the all-domain classifier, respectively. Loss vs.\ full is signed so that positive values indicate worse performance: leave-one-out minus full for Brier and reliability, and full minus leave-one-out for resolution. Gain retained places the leave-one-out classifier between Qwen3 ($0\%$) and the full model ($100\%$). Loss and retention intervals use paired bootstraps over identical episodes.}
\label{tab:transfer-full}
\providecommand{\tmstack}[2]{\begin{tabular}[t]{@{}c@{}}#1\\[-2pt]{\scriptsize\color{black!45}#2}\end{tabular}}%
\begin{tabular}{@{}lcccc@{}}
\multicolumn{5}{@{}l}{\textbf{A. Absolute Brier performance}} \\[2pt]
\toprule
Training mix & Tahoe & DepMap & GWAS & Macro \\
\midrule
DepMap + GWAS (Tahoe held out) & \tmstack{\textbf{0.2535}}{[0.2426, 0.2641]} & \tmstack{0.1641}{[0.1539, 0.1739]} & \tmstack{0.1483}{[0.1367, 0.1599]} & \tmstack{0.1886}{[0.1825, 0.1945]} \\
Tahoe + GWAS (DepMap held out) & \tmstack{0.2072}{[0.1994, 0.2151]} & \tmstack{\textbf{0.2337}}{[0.2253, 0.2417]} & \tmstack{0.1509}{[0.1400, 0.1613]} & \tmstack{0.1972}{[0.1919, 0.2025]} \\
Tahoe + DepMap (GWAS held out) & \tmstack{0.2047}{[0.1955, 0.2139]} & \tmstack{0.1639}{[0.1534, 0.1743]} & \tmstack{\textbf{0.2993}}{[0.2870, 0.3114]} & \tmstack{0.2226}{[0.2165, 0.2287]} \\
\midrule
All domains & \tmstack{0.2063}{[0.1971, 0.2157]} & \tmstack{0.1713}{[0.1602, 0.1821]} & \tmstack{0.1537}{[0.1416, 0.1654]} & \tmstack{0.1771}{[0.1707, 0.1831]} \\
\bottomrule
\end{tabular}
\vspace{0.8em}
\begin{tabular}{@{}llccccc@{}}
\multicolumn{7}{@{}l}{\textbf{B. Held-out-domain transfer decomposition}} \\[2pt]
\toprule
Held-out & Metric & Qwen3 & Leave-one-out & Full model & Loss vs. full & Gain retained \\
\midrule
Tahoe & Brier Score & \tmstack{0.3634}{[0.3459, 0.3827]} & \tmstack{0.2535}{[0.2426, 0.2641]} & \tmstack{0.2063}{[0.1971, 0.2157]} & \tmstack{0.0471}{[0.0378, 0.0571]} & \tmstack{70.0\%}{[65.3, 74.7]\%} \\
 & Reliability & \tmstack{0.1277}{[0.1140, 0.1452]} & \tmstack{0.0213}{[0.0161, 0.0290]} & \tmstack{0.0042}{[0.0028, 0.0082]} & \tmstack{0.0170}{[0.0115, 0.0229]} & \tmstack{86.2\%}{[82.5, 89.7]\%} \\
 & Resolution & \tmstack{0.0147}{[0.0111, 0.0204]} & \tmstack{0.0178}{[0.0137, 0.0245]} & \tmstack{0.0470}{[0.0407, 0.0556]} & \tmstack{0.0291}{[0.0229, 0.0356]} & \tmstack{9.8\%}{[-4.6, 22.9]\%} \\
\addlinespace[2pt]
DepMap & Brier Score & \tmstack{0.2860}{[0.2715, 0.2995]} & \tmstack{0.2337}{[0.2253, 0.2417]} & \tmstack{0.1713}{[0.1602, 0.1821]} & \tmstack{0.0624}{[0.0528, 0.0722]} & \tmstack{45.6\%}{[41.0, 50.3]\%} \\
 & Reliability & \tmstack{0.0854}{[0.0754, 0.0977]} & \tmstack{0.0354}{[0.0290, 0.0436]} & \tmstack{0.0076}{[0.0054, 0.0124]} & \tmstack{0.0278}{[0.0202, 0.0348]} & \tmstack{64.3\%}{[58.6, 71.3]\%} \\
 & Resolution & \tmstack{0.0485}{[0.0414, 0.0577]} & \tmstack{0.0486}{[0.0415, 0.0573]} & \tmstack{0.0855}{[0.0780, 0.0952]} & \tmstack{0.0369}{[0.0298, 0.0449]} & \tmstack{0.4\%}{[-16.0, 15.0]\%} \\
\addlinespace[2pt]
GWAS & Brier Score & \tmstack{0.3513}{[0.3333, 0.3694]} & \tmstack{0.2993}{[0.2870, 0.3114]} & \tmstack{0.1537}{[0.1416, 0.1654]} & \tmstack{0.1456}{[0.1296, 0.1625]} & \tmstack{26.3\%}{[16.1, 35.4]\%} \\
 & Reliability & \tmstack{0.1156}{[0.1022, 0.1314]} & \tmstack{0.0546}{[0.0459, 0.0666]} & \tmstack{0.0107}{[0.0078, 0.0160]} & \tmstack{0.0439}{[0.0329, 0.0557]} & \tmstack{58.1\%}{[42.1, 70.5]\%} \\
 & Resolution & \tmstack{0.0120}{[0.0089, 0.0176]} & \tmstack{0.0054}{[0.0035, 0.0097]} & \tmstack{0.1046}{[0.0964, 0.1153]} & \tmstack{0.0992}{[0.0898, 0.1086]} & \tmstack{-7.1\%}{[-11.8, -2.8]\%} \\
\bottomrule
\end{tabular}

\end{table}

\FloatBarrier

\section{Experimental design details}
\label{app:bed-details}

This appendix collects material supporting \cref{sec:bed}.

\subsection{Naive-prompting selection prompt}
\label{app:naive-prompt}

The naive-prompting policy $\pi_{\text{naive}}$ (\cref{sec:bed}) asks a model to name the single most informative candidate directly, given the target hypothesis $h^\ast$ and the evidence revealed so far. Because this is a selection task rather than the probability elicitation of the belief model, it uses a distinct system prompt that forbids a probability and fixes the answer format to \verb|<choice>i</choice>|: reusing the belief prompt caused format-strict models to emit \verb|<probability>| and silently fall back to random selection. The system prompt is

\begin{promptbox}[promptsystem]{System prompt}
\begin{PromptText}
You are an expert scientist selecting which experiment to run next. You are
shown a target hypothesis, the evidence gathered so far, and a numbered list
of candidate experiments whose results are hidden. Do NOT output a probability.
Your only task is to choose the single candidate whose hidden result would most
change your belief about the target hypothesis. Respond with just its index in
the exact form <choice>i</choice>.
\end{PromptText}
\end{promptbox}

The user message reuses the belief model's rendering of the target hypothesis and the revealed evidence---the \texttt{CONTEXT EVIDENCE} and \texttt{TARGET HYPOTHESIS} blocks documented in \cref{app:prompts}---then appends the numbered candidate experiments, each rendered with its result hidden, followed by the selection instruction:

\begin{promptbox}[promptuser]{User prompt template}
\begin{PromptText}
<CONTEXT EVIDENCE and TARGET HYPOTHESIS, rendered as in the prompts appendix>

CANDIDATE NEXT EXPERIMENTS (results hidden):
[0] <candidate experiment setup, result hidden>
[1] <candidate experiment setup, result hidden>
...

Which single candidate experiment, once its result is revealed, would most
change your belief about the target hypothesis? Answer with just its index as
<choice>i</choice>.
\end{PromptText}
\end{promptbox}

The same prompt is used for every naive policy (Qwen3, \modelname{}, GPT-5.6-sol, Opus 5); only the answering model changes. The belief tracker that scores each policy's trajectory is held fixed across policies, so the comparison isolates selection quality.

\FloatBarrier

\section{Post-hoc recalibration}
\label{app:platt}

We test how much of each model's Brier deficit lies in its probability scale rather than its ranking, and whether the trained models keep their advantage when every model is recalibrated.

\paragraph{Procedure.}
For each model and each domain we fit Platt scaling \citep{platt1999probabilistic} to the model's own reported probabilities: a two-parameter affine map on the logit, $\hat{p} = \sigma\!\left(a \cdot \operatorname{logit}(p) + b\right)$, estimated by logistic regression of the observed outcome on $\operatorname{logit}(p)$ with negligible regularization ($C = 10^{6}$). The map is \emph{cross-fitted} over two random folds: $a$ and $b$ are estimated on one half of a domain's episodes and applied to the other, so no episode is scored under a map to which it contributed. Refitting within each bootstrap draw propagates fit variability into the intervals in \cref{tab:multidomain-recal}. We recalibrate all seven models, trained and untrained alike.

\providecommand{\mtstack}[2]{\begin{tabular}[t]{@{}r@{}}#1\\[-2pt]{\scriptsize\color{black!45}#2}\end{tabular}}
\begin{table}[t]
\centering
\footnotesize
\setlength{\tabcolsep}{3pt}
\renewcommand{\arraystretch}{0.9}
\begin{tabular}{lrrrrrr}
\toprule
Model & $n$ & Brier Score ($\downarrow$) & sMECE ($\downarrow$) & AUROC ($\uparrow$) & Reliability ($\downarrow$) & Resolution ($\uparrow$) \\
\midrule
\multicolumn{7}{l}{\textit{As reported}} \\
Qwen3 (non-thinking) & 6,000 & \mtstack{0.3336}{[0.3237, 0.3436]} & \mtstack{0.3101}{[0.2990, 0.3229]} & \mtstack{0.6608}{[0.6471, 0.6752]} & \mtstack{0.1096}{[0.1024, 0.1189]} & \mtstack{0.0250}{[0.0226, 0.0295]} \\
Qwen3 (thinking) & 6,000 & \mtstack{0.2973}{[0.2896, 0.3047]} & \mtstack{0.2696}{[0.2607, 0.2818]} & \mtstack{0.7040}{[0.6914, 0.7170]} & \mtstack{0.0829}{[0.0772, 0.0907]} & \mtstack{0.0339}{[0.0307, 0.0387]} \\
\modelname{} (non-thinking) & 6,000 & \mtstack{0.1933}{[0.1888, 0.1977]} & \mtstack{0.0316}{[0.0258, 0.0450]} & \mtstack{0.7729}{[0.7616, 0.7846]} & \mtstack{0.0020}{[0.0017, 0.0038]} & \mtstack{0.0579}{[0.0540, 0.0631]} \\
\modelname{} (thinking) & 6,000 & \mtstack{0.2069}{[0.2025, 0.2111]} & \mtstack{0.0327}{[0.0296, 0.0431]} & \mtstack{0.7329}{[0.7209, 0.7450]} & \mtstack{0.0023}{[0.0021, 0.0044]} & \mtstack{0.0448}{[0.0413, 0.0500]} \\
\modelname{} (classifier) & 6,000 & \mtstack{0.1771}{[0.1707, 0.1831]} & \mtstack{0.0658}{[0.0588, 0.0762]} & \mtstack{0.8195}{[0.8099, 0.8295]} & \mtstack{0.0075}{[0.0065, 0.0104]} & \mtstack{0.0790}{[0.0753, 0.0849]} \\
GPT-5.6-sol & 6,000 & \mtstack{0.2879}{[0.2793, 0.2964]} & \mtstack{0.2602}{[0.2490, 0.2712]} & \mtstack{0.7365}{[0.7235, 0.7485]} & \mtstack{0.0830}{[0.0775, 0.0908]} & \mtstack{0.0430}{[0.0395, 0.0484]} \\
Claude Opus 5 & 6,000 & \mtstack{0.2772}{[0.2689, 0.2851]} & \mtstack{0.2390}{[0.2291, 0.2505]} & \mtstack{0.7519}{[0.7392, 0.7646]} & \mtstack{0.0786}{[0.0730, 0.0857]} & \mtstack{0.0502}{[0.0461, 0.0556]} \\
\addlinespace
\multicolumn{7}{l}{\textit{After cross-fit Platt scaling}} \\
Qwen3 (non-thinking) & 6,000 & \mtstack{0.2315}{[0.2274, 0.2353]} & \mtstack{0.0603}{[0.0417, 0.0657]} & \mtstack{0.6580}{[0.6363, 0.6698]} & \mtstack{0.0043}{[0.0027, 0.0075]} & \mtstack{0.0214}{[0.0180, 0.0266]} \\
Qwen3 (thinking) & 6,000 & \mtstack{0.2148}{[0.2109, 0.2192]} & \mtstack{0.0281}{[0.0239, 0.0380]} & \mtstack{0.7033}{[0.6880, 0.7152]} & \mtstack{0.0018}{[0.0013, 0.0034]} & \mtstack{0.0363}{[0.0324, 0.0407]} \\
\modelname{} (non-thinking) & 6,000 & \mtstack{0.1920}{[0.1874, 0.1969]} & \mtstack{0.0172}{[0.0162, 0.0283]} & \mtstack{0.7723}{[0.7591, 0.7828]} & \mtstack{0.0006$^{\dagger}$}{[0.0007, 0.0020]} & \mtstack{0.0574}{[0.0533, 0.0628]} \\
\modelname{} (thinking) & 6,000 & \mtstack{0.2062}{[0.2017, 0.2105]} & \mtstack{0.0257}{[0.0247, 0.0363]} & \mtstack{0.7327}{[0.7185, 0.7437]} & \mtstack{0.0016}{[0.0012, 0.0031]} & \mtstack{0.0442}{[0.0406, 0.0496]} \\
\modelname{} (classifier) & 6,000 & \mtstack{0.1706}{[0.1657, 0.1754]} & \mtstack{0.0212}{[0.0184, 0.0292]} & \mtstack{0.8193}{[0.8087, 0.8285]} & \mtstack{0.0014}{[0.0011, 0.0029]} & \mtstack{0.0799}{[0.0754, 0.0852]} \\
GPT-5.6-sol & 6,000 & \mtstack{0.2055}{[0.2011, 0.2103]} & \mtstack{0.0239}{[0.0205, 0.0331]} & \mtstack{0.7360}{[0.7219, 0.7472]} & \mtstack{0.0010}{[0.0008, 0.0026]} & \mtstack{0.0450}{[0.0404, 0.0499]} \\
Claude Opus 5 & 6,000 & \mtstack{0.1997}{[0.1950, 0.2046]} & \mtstack{0.0314}{[0.0263, 0.0406]} & \mtstack{0.7512}{[0.7377, 0.7633]} & \mtstack{0.0017}{[0.0014, 0.0034]} & \mtstack{0.0511}{[0.0468, 0.0569]} \\
\bottomrule
\end{tabular}
\\[2pt]
\parbox{\linewidth}{\footnotesize\raggedright Macro-average over Tahoe, DepMap, GWAS; 95\% CI from a within-domain bootstrap resampling episodes (2,000 draws; sMECE 200, it is ~35x costlier per draw). $\dagger$ point estimate outside its bootstrap interval --- expected for the positively-biased binned metrics (Reliability, sMECE) on rows near the zero floor. Compare rows by these intervals; do not read them as covering the point.}
\caption{Cross-fit Platt recalibration of every evaluated system: a two-parameter affine map on the logit, fitted out-of-fold within each domain. Because it requires labelled target-domain data, it bounds how much of each gap a post-hoc fix can close and is not a deployable method. The trained systems are also recalibrated so that the two blocks are comparable.}
\label{tab:multidomain-recal}
\end{table}

\paragraph{Calibration and ranking after scaling.}
sMECE decreases for every model, by a factor of $5.1$ to $10.9$ for the four untrained models and $1.3$ to $3.1$ for the three trained ones. Murphy reliability follows the same split, falling by a factor of $26$ to $80$ against $1.4$ to $5.4$. The trained models gain less because they start better calibrated.

Post-hoc scaling reduces calibration error without materially changing rank discrimination. The largest change in AUROC is $0.003$, for Qwen3 (non-thinking); every other change is below $0.001$, and the model ordering by AUROC is unchanged. A single affine map on the logit is monotone and preserves ranking. The small changes arise because the two cross-fitting folds use different maps, which can reorder observations across the fold boundary.

\paragraph{Forecast metrics after scaling.}
Recalibration removes most of the untrained models' Brier deficit: Brier falls by $0.077$ to $0.102$ for the four untrained models, against $0.001$ to $0.007$ for the three trained ones. \modelname{} (classifier) still has the lowest recalibrated Brier at $0.171$, ahead of Claude Opus 5 at $0.200$, with non-overlapping intervals.

Matching the treatment reverses the sMECE comparison. Recalibrated GPT-5.6-sol reaches $0.024$, against $0.017$ for recalibrated \modelname{} (non-thinking) and $0.021$ for \modelname{} (classifier), with overlapping intervals. Comparing recalibrated frontier models with unscaled trained ones understates the trained models on this metric, because the classifier's as-reported sMECE of $0.066$ is the one value in its row with appreciable headroom.

\providecommand{\mtstack}[2]{\begin{tabular}[t]{@{}r@{}}#1\\[-2pt]{\scriptsize\color{black!45}#2}\end{tabular}}
\begin{table}[t]
\centering
\footnotesize
\setlength{\tabcolsep}{3pt}
\renewcommand{\arraystretch}{0.9}
\begin{tabular}{lrrrrrr}
\toprule
Model & $n$ & Brier Score ($\downarrow$) & sMECE ($\downarrow$) & AUROC ($\uparrow$) & Reliability ($\downarrow$) & Resolution ($\uparrow$) \\
\midrule
Qwen3 (non-thinking) & 6,000 & \mtstack{0.2315}{[0.2274, 0.2353]} & \mtstack{0.0603}{[0.0417, 0.0657]} & \mtstack{0.6580}{[0.6363, 0.6698]} & \mtstack{0.0043}{[0.0027, 0.0075]} & \mtstack{0.0214}{[0.0180, 0.0266]} \\
\quad Tahoe & 2,000 & \mtstack{0.2451}{[0.2419, 0.2487]} & \mtstack{0.0656}{[0.0383, 0.0743]} & \mtstack{0.6034}{[0.5508, 0.6223]} & \mtstack{0.0029}{[0.0006, 0.0078]} & \mtstack{0.0048}{[0.0007, 0.0138]} \\
\quad DepMap & 2,000 & \mtstack{0.2092}{[0.1987, 0.2186]} & \mtstack{0.0703}{[0.0469, 0.0968]} & \mtstack{0.7431}{[0.7218, 0.7654]} & \mtstack{0.0082}{[0.0041, 0.0154]} & \mtstack{0.0486}{[0.0410, 0.0588]} \\
\quad GWAS & 2,000 & \mtstack{0.2401}{[0.2348, 0.2443]} & \mtstack{0.0450}{[0.0165, 0.0634]} & \mtstack{0.6274}{[0.5936, 0.6502]} & \mtstack{0.0018}{[0.0003, 0.0051]} & \mtstack{0.0109}{[0.0051, 0.0177]} \\
\addlinespace
Qwen3 (thinking) & 6,000 & \mtstack{0.2148}{[0.2109, 0.2192]} & \mtstack{0.0281}{[0.0239, 0.0380]} & \mtstack{0.7033}{[0.6880, 0.7152]} & \mtstack{0.0018}{[0.0013, 0.0034]} & \mtstack{0.0363}{[0.0324, 0.0407]} \\
\quad Tahoe & 2,000 & \mtstack{0.2245}{[0.2182, 0.2312]} & \mtstack{0.0302}{[0.0204, 0.0455]} & \mtstack{0.6814}{[0.6540, 0.7027]} & \mtstack{0.0023}{[0.0010, 0.0050]} & \mtstack{0.0278}{[0.0204, 0.0338]} \\
\quad DepMap & 2,000 & \mtstack{0.1863}{[0.1773, 0.1951]} & \mtstack{0.0275}{[0.0212, 0.0428]} & \mtstack{0.7829}{[0.7614, 0.8020]} & \mtstack{0.0011}{[0.0007, 0.0046]} & \mtstack{0.0645}{[0.0568, 0.0745]} \\
\quad GWAS & 2,000 & \mtstack{0.2337}{[0.2277, 0.2391]} & \mtstack{0.0265}{[0.0172, 0.0444]} & \mtstack{0.6456}{[0.6190, 0.6702]} & \mtstack{0.0019}{[0.0005, 0.0037]} & \mtstack{0.0166}{[0.0113, 0.0233]} \\
\addlinespace
\modelname{} (non-thinking) & 6,000 & \mtstack{0.1920}{[0.1874, 0.1969]} & \mtstack{0.0172}{[0.0162, 0.0283]} & \mtstack{0.7723}{[0.7591, 0.7828]} & \mtstack{0.0006$^{\dagger}$}{[0.0007, 0.0020]} & \mtstack{0.0574}{[0.0533, 0.0628]} \\
\quad Tahoe & 2,000 & \mtstack{0.2149}{[0.2082, 0.2220]} & \mtstack{0.0197}{[0.0146, 0.0373]} & \mtstack{0.7131}{[0.6875, 0.7333]} & \mtstack{0.0004}{[0.0003, 0.0026]} & \mtstack{0.0349}{[0.0282, 0.0424]} \\
\quad DepMap & 2,000 & \mtstack{0.1754}{[0.1665, 0.1844]} & \mtstack{0.0119}{[0.0114, 0.0240]} & \mtstack{0.8116}{[0.7909, 0.8297]} & \mtstack{0.0005}{[0.0003, 0.0025]} & \mtstack{0.0737}{[0.0653, 0.0841]} \\
\quad GWAS & 2,000 & \mtstack{0.1858}{[0.1770, 0.1952]} & \mtstack{0.0200}{[0.0142, 0.0348]} & \mtstack{0.7921}{[0.7703, 0.8113]} & \mtstack{0.0008}{[0.0004, 0.0030]} & \mtstack{0.0635}{[0.0553, 0.0732]} \\
\addlinespace
\modelname{} (thinking) & 6,000 & \mtstack{0.2062}{[0.2017, 0.2105]} & \mtstack{0.0257}{[0.0247, 0.0363]} & \mtstack{0.7327}{[0.7185, 0.7437]} & \mtstack{0.0016}{[0.0012, 0.0031]} & \mtstack{0.0442}{[0.0406, 0.0496]} \\
\quad Tahoe & 2,000 & \mtstack{0.2164}{[0.2095, 0.2234]} & \mtstack{0.0176}{[0.0135, 0.0325]} & \mtstack{0.7065}{[0.6816, 0.7278]} & \mtstack{0.0007}{[0.0003, 0.0027]} & \mtstack{0.0325}{[0.0264, 0.0404]} \\
\quad DepMap & 2,000 & \mtstack{0.1819}{[0.1726, 0.1908]} & \mtstack{0.0274}{[0.0226, 0.0424]} & \mtstack{0.7951}{[0.7748, 0.8152]} & \mtstack{0.0019}{[0.0007, 0.0043]} & \mtstack{0.0692}{[0.0601, 0.0793]} \\
\quad GWAS & 2,000 & \mtstack{0.2202}{[0.2127, 0.2272]} & \mtstack{0.0321}{[0.0227, 0.0499]} & \mtstack{0.6966}{[0.6701, 0.7179]} & \mtstack{0.0023}{[0.0010, 0.0053]} & \mtstack{0.0310}{[0.0248, 0.0397]} \\
\addlinespace
\modelname{} (classifier) & 6,000 & \mtstack{0.1706}{[0.1657, 0.1754]} & \mtstack{0.0212}{[0.0184, 0.0292]} & \mtstack{0.8193}{[0.8087, 0.8285]} & \mtstack{0.0014}{[0.0011, 0.0029]} & \mtstack{0.0799}{[0.0754, 0.0852]} \\
\quad Tahoe & 2,000 & \mtstack{0.2032}{[0.1960, 0.2108]} & \mtstack{0.0270}{[0.0178, 0.0436]} & \mtstack{0.7486}{[0.7262, 0.7675]} & \mtstack{0.0019}{[0.0007, 0.0041]} & \mtstack{0.0477}{[0.0401, 0.0553]} \\
\quad DepMap & 2,000 & \mtstack{0.1646}{[0.1559, 0.1730]} & \mtstack{0.0181}{[0.0133, 0.0323]} & \mtstack{0.8331}{[0.8152, 0.8489]} & \mtstack{0.0007}{[0.0005, 0.0032]} & \mtstack{0.0851}{[0.0773, 0.0947]} \\
\quad GWAS & 2,000 & \mtstack{0.1440}{[0.1345, 0.1531]} & \mtstack{0.0187}{[0.0134, 0.0291]} & \mtstack{0.8762}{[0.8606, 0.8909]} & \mtstack{0.0016}{[0.0008, 0.0042]} & \mtstack{0.1069}{[0.0984, 0.1164]} \\
\addlinespace
GPT-5.6-sol & 6,000 & \mtstack{0.2055}{[0.2011, 0.2103]} & \mtstack{0.0239}{[0.0205, 0.0331]} & \mtstack{0.7360}{[0.7219, 0.7472]} & \mtstack{0.0010}{[0.0008, 0.0026]} & \mtstack{0.0450}{[0.0404, 0.0499]} \\
\quad Tahoe & 2,000 & \mtstack{0.2192}{[0.2125, 0.2260]} & \mtstack{0.0203}{[0.0143, 0.0351]} & \mtstack{0.6983}{[0.6737, 0.7198]} & \mtstack{0.0009}{[0.0003, 0.0033]} & \mtstack{0.0310}{[0.0241, 0.0387]} \\
\quad DepMap & 2,000 & \mtstack{0.1801}{[0.1711, 0.1895]} & \mtstack{0.0259}{[0.0175, 0.0393]} & \mtstack{0.8042}{[0.7830, 0.8222]} & \mtstack{0.0014}{[0.0007, 0.0036]} & \mtstack{0.0703}{[0.0613, 0.0803]} \\
\quad GWAS & 2,000 & \mtstack{0.2173}{[0.2100, 0.2249]} & \mtstack{0.0254}{[0.0166, 0.0426]} & \mtstack{0.7054}{[0.6801, 0.7273]} & \mtstack{0.0008}{[0.0003, 0.0031]} & \mtstack{0.0337}{[0.0258, 0.0415]} \\
\addlinespace
Claude Opus 5 & 6,000 & \mtstack{0.1997}{[0.1950, 0.2046]} & \mtstack{0.0314}{[0.0263, 0.0406]} & \mtstack{0.7512}{[0.7377, 0.7633]} & \mtstack{0.0017}{[0.0014, 0.0034]} & \mtstack{0.0511}{[0.0468, 0.0569]} \\
\quad Tahoe & 2,000 & \mtstack{0.2232}{[0.2169, 0.2296]} & \mtstack{0.0157}{[0.0120, 0.0303]} & \mtstack{0.6854}{[0.6603, 0.7069]} & \mtstack{0.0002$^{\dagger}$}{[0.0002, 0.0019]} & \mtstack{0.0259}{[0.0203, 0.0337]} \\
\quad DepMap & 2,000 & \mtstack{0.1788}{[0.1689, 0.1878]} & \mtstack{0.0253}{[0.0156, 0.0362]} & \mtstack{0.8035}{[0.7836, 0.8230]} & \mtstack{0.0009}{[0.0004, 0.0026]} & \mtstack{0.0714}{[0.0627, 0.0813]} \\
\quad GWAS & 2,000 & \mtstack{0.1971}{[0.1881, 0.2060]} & \mtstack{0.0532}{[0.0391, 0.0691]} & \mtstack{0.7647}{[0.7420, 0.7854]} & \mtstack{0.0041}{[0.0025, 0.0077]} & \mtstack{0.0561}{[0.0476, 0.0664]} \\
\bottomrule
\end{tabular}
\\[2pt]
\parbox{\linewidth}{\footnotesize\raggedright Macro-average over Tahoe, DepMap, GWAS; 95\% CI from a within-domain bootstrap resampling episodes (2,000 draws; sMECE 200, it is ~35x costlier per draw). The unindented row of each block is that macro-average; the indented rows below it are the single domains, bootstrapped over that domain's episodes alone. $\dagger$ point estimate outside its bootstrap interval --- expected for the positively-biased binned metrics (Reliability, sMECE) on rows near the zero floor. Compare rows by these intervals; do not read them as covering the point.}
\caption{Cross-fit Platt recalibration by domain, with the map fitted separately within each of Tahoe, DepMap and GWAS. The as-reported breakdown is in the complete master table.}
\label{tab:multidomain-recal-per-domain}
\end{table}

\paragraph{Recalibration by domain.}
\Cref{tab:multidomain-recal-per-domain} fits $a$ and $b$ separately within Tahoe, DepMap and GWAS. The gains are not uniform across domains. Tahoe remains the hardest domain for every model after scaling, with recalibrated Brier between $0.203$ and $0.245$, against $0.165$ to $0.209$ on DepMap. \modelname{} (classifier) has the widest spread, from $0.144$ on GWAS to $0.203$ on Tahoe, a range that the macro-average conceals.

\paragraph{Requirement for representative labels.}
Because $a$ and $b$ are estimated from other labeled episodes in the same domain, the procedure applies only when representative historical outcomes exist in the deployment domain. It is unavailable for a new assay with no calibration sample and may be invalid under distribution shift.

\paragraph{Resolution after scaling.}
Resolution is computed over ten equal-width forecast bins, so a monotone rescaling changes it only by shifting episodes across a bin edge. The seven resolutions change by at most $0.004$ in either direction. After scaling, Qwen3 (non-thinking) is at $0.021$ and Qwen3 (thinking) at $0.036$, compared with $0.080$ for the classifier. Coarse quantization of the verbalized forecasts may contribute to this difference, but the current comparison does not isolate that mechanism.

\FloatBarrier

\section{Prompts: one worked example per domain}
\label{app:prompts}

A shared function renders every episode as a prompt. The prompt contains a \texttt{CONTEXT EVIDENCE} header, one numbered block per evidence stub beginning with its relevance tag, a \texttt{TARGET HYPOTHESIS} header, the target block, and a fixed answer-format instruction. Domain-specific functions render individual stubs and targets. \Cref{sec:task} defines the episode structure; this appendix provides one rendered example per domain.

Each system prompt describes the assay and evidence-generation procedure, defines the reported quantities, states the support-label rule, and supplies an answer-format instruction. The \emph{direct} instruction requests only a probability; the \emph{thinking} instruction requests a reasoning trace followed by the same probability tag. Each evaluated model receives the same instruction in every domain (\cref{sec:training-details,app:frontier}); the examples use the direct instruction.

\begin{promptbox}[promptuser]{Direct instruction}
\begin{PromptText}
Respond with ONLY a single line in this exact format — no reasoning, no explanation, no other text before or after:
<probability>0.XX</probability>
\end{PromptText}
\end{promptbox}

\begin{promptbox}[promptuser]{Thinking instruction}
\begin{PromptText}
Think through the evidence carefully, then commit to a single probability in the closed interval [0, 1].
  After your reasoning, output your probability in this exact format:
<probability>0.XX</probability>
\end{PromptText}
\end{promptbox}

The examples are drawn from the \texttt{primary} test-set configuration. Long lines wrap at the margin, with each continuation marked by $\hookrightarrow$, and the sole \texttt{!=} in the Tahoe system prompt replaces a \texttt{$\neq$} glyph the document font cannot set. Evidence pools are truncated after the first two stubs, and the number of omitted stubs is marked in place. The displayed text is otherwise unchanged.

\subsection{Tahoe: drug perturbation of cancer cell lines}

Episode \texttt{eb33e117-0ab} asks whether encorafenib upregulates \texttt{LINC00342} in \texttt{SNU-1}. The episode contains six stubs, and the held-out DESeq2 result supports the hypothesis.

The two displayed stubs both measure the target drug and gene and are statistically significant; one reports a decrease in a lung line and the other an increase of similar magnitude in a skin line. Of the four omitted stubs, one is the target cell line perturbed by a different drug (a null result), and three are unrelated drug\,$\times$\,cell-line baselines. No stub reports encorafenib in \texttt{SNU-1}, so the pool constrains the answer without containing it.

\begin{promptbox}[promptsystem]{System prompt}
\begin{PromptText}
You are a scientific reasoning assistant analyzing drug-perturbation evidence from Tahoe-100M, a large-scale pseudobulk RNA-seq dataset measuring drug × cell-line × gene differential expression.

How the evidence was generated:
  - Cells from a given cell line are treated with a drug (or DMSO vehicle control).
  - Each cell's transcripts are captured by single-cell RNA-seq; counts are UMI-deduplicated per gene.
  - Single cells from the same (drug, cell_line, plate, replicate) bucket are *pseudobulked* by summing per-gene UMI counts.
  - For each gene, DESeq2 fits a negative-binomial GLM comparing treated vs control pseudobulk samples.

Column definitions (DESeq2 conventions):
  - log2FoldChange: shrunken log2(treated / control) of normalized counts. Positive = upregulated; negative = downregulated.
  - lfcSE:         standard error of the log2FoldChange estimate (uncertainty).
  - pvalue:        Wald test p-value for log2FoldChange != 0.
  - padj:          Benjamini-Hochberg adjusted p-value (FDR-controlled across all genes in that drug × cell_line comparison).
  - baseMean:      mean of normalized counts across both treated AND control samples combined (an expression-level indicator, NOT a control baseline).
  - n_cells_trt / n_cells_ctrl: number of single cells pooled into the treated / control pseudobulk.

You will be shown:
  1. CONTEXT — prior experimental results: drug × cell-line × gene differential-expression measurements, plus aggregate perturbation summaries.
  2. A TARGET HYPOTHESIS — a claim of the form "drug D upregulates/downregulates gene G in cell line C".

Weigh the evidence and report your degree of belief that a held-out experiment would *support* the target hypothesis. Supported means padj < 0.05 AND |log2FoldChange| > 0.5 in the claimed direction.

Respond with ONLY a single line in this exact format — no reasoning, no explanation, no other text before or after:
<probability>0.XX</probability>
\end{PromptText}
\end{promptbox}

\begin{promptbox}[promptuser]{User prompt \textnormal{(episode \texttt{eb33e117-0ab}, 6 stubs, label = supported)}}
\begin{PromptText}
CONTEXT EVIDENCE
================

[1] same drug as target, different cell line, target gene
    Drug: Encorafenib
      MOA (broad)         inhibitor/antagonist
      MOA (fine)          RAF inhibitor
      Targets             BRAF
      Human-approved      yes
      In clinical trials  yes
      Notes               Approved for use in BRAF-mutant melanoma.
      Canonical SMILES    CC(C)N1C=C(C(=N1)C2=C(C(=CC(=C2)Cl)NS(=O)(=O)C)F)C3=NC(=NC=C3)NCC(C)NC(=O)OC
      PubChem CID         50922675
    Cell line: NCI-H2347
      DepMap ID                   ACH-000875
      Cellosaurus ID              CVCL_1550
      Organ                       Lung
      Driver gene                 B2M
      Driver variant type         Stopgain
      Driver zygosity             Hom
      Driver protein/cDNA effect  p.Q28*
      Driver mechanism (LoF/GoF)  LoF
      Driver gene type            Suppressor
    Gene: LINC00342
      Ensembl ID  ENSG00000232931
    Experiment:
      Plate                                              3
      n_cells treated                                    2747
      n_cells control                                    3106
      baseMean (avg normalized counts across both arms)  90.74
    Result:
      log2FoldChange      -0.473 ± 0.153  (downregulates)
      Wald statistic      -3.101
      p-value (raw)       0.00193
      p-value (adjusted)  0.01023

[2] same drug as target, different cell line, target gene
    Drug: Encorafenib
      MOA (broad)         inhibitor/antagonist
      MOA (fine)          RAF inhibitor
      Targets             BRAF
      Human-approved      yes
      In clinical trials  yes
      Notes               Approved for use in BRAF-mutant melanoma.
      Canonical SMILES    CC(C)N1C=C(C(=N1)C2=C(C(=CC(=C2)Cl)NS(=O)(=O)C)F)C3=NC(=NC=C3)NCC(C)NC(=O)OC
      PubChem CID         50922675
    Cell line: LOX-IMVI
      DepMap ID                   ACH-000750
      Cellosaurus ID              CVCL_1381
      Organ                       Skin
      Driver gene                 ARID1A
      Driver variant type         Stopgain
      Driver zygosity             Het
      Driver protein/cDNA effect  p.Q1212*
      Driver mechanism (LoF/GoF)  LoF
      Driver gene type            Suppressor
    Gene: LINC00342
      Ensembl ID  ENSG00000232931
    Experiment:
      Plate                                              3
      n_cells treated                                    2163
      n_cells control                                    2021
      baseMean (avg normalized counts across both arms)  154.3
    Result:
      log2FoldChange      +0.535 ± 0.136  (upregulates)
      Wald statistic      3.925
      p-value (raw)       8.66e-05
      p-value (adjusted)  0.00113

[... 4 further evidence stubs omitted ...]

TARGET HYPOTHESIS
=================
Drug Encorafenib upregulates gene LINC00342 in cell line SNU-1.

    Drug: Encorafenib
      MOA (broad)         inhibitor/antagonist
      MOA (fine)          RAF inhibitor
      Targets             BRAF
      Human-approved      yes
      In clinical trials  yes
      Notes               Approved for use in BRAF-mutant melanoma.
      Canonical SMILES    CC(C)N1C=C(C(=N1)C2=C(C(=CC(=C2)Cl)NS(=O)(=O)C)F)C3=NC(=NC=C3)NCC(C)NC(=O)OC
      PubChem CID         50922675
    Cell line: SNU-1
      DepMap ID                   ACH-000932
      Cellosaurus ID              CVCL_0099
      Organ                       Esophagus/Stomach
      Driver gene                 ARID1A
      Driver variant type         Frameshift
      Driver zygosity             Het
      Driver protein/cDNA effect  p.D1633fs
      Driver mechanism (LoF/GoF)  LoF
      Driver gene type            Suppressor
    Gene: LINC00342
      Ensembl ID  ENSG00000232931

Respond using the format specified in the system message:
  <probability>0.XX</probability>
\end{PromptText}
\end{promptbox}

\subsection{DepMap: cancer gene dependency}

Episode \texttt{65b4265e-9f1} asks whether \texttt{DIS3L} is a dependency in the mesothelioma line \texttt{MERO25}. The held-out screen does not support the hypothesis.

The first displayed stub reports the target gene's mutation status in the target line rather than a knockout result, an example of the multi-modal evidence pool. The four omitted stubs are two unrelated knockouts in \texttt{MERO25} and two knockouts of \texttt{DIS3L} in other pleural mesothelioma lines, the latter scoring $-0.25$ and $-0.45$ against a threshold of $-0.5$.

\begin{promptbox}[promptsystem]{System prompt}
\begin{PromptText}
You are a scientific reasoning assistant analyzing cancer gene-dependency evidence from DepMap, a large-scale project of genome-wide CRISPR knockout screens of cancer cell lines with matched molecular and pharmacological profiles.
How the evidence was generated:
  - Each cell line is screened by pooled genome-wide CRISPR: every gene is knocked out (one gene per cell), the culture grows ~3 weeks, and cells whose knocked-out gene was required for proliferation drop out of the pool.
  - The depletion of each gene is summarized by the Chronos gene-effect score; a calibrated P(dependency) is derived from it.
  - Matched profiles, measured independently of the screen, give each line's baseline mRNA expression, copy number, somatic mutations, and (for some) drug sensitivity.
Quantity definitions:
  - Chronos gene-effect: a CRISPR-knockout fitness score, scaled so 0 = no effect on growth and -1 = as lethal as a typical core-essential gene (more negative = more essential; positive = knockout speeds growth). Analogous to a log2 fold-change.
  - P(dependency): the calibrated probability in [0,1] that the gene is a true dependency in that cell line, inferred from the Chronos score. Analogous to statistical significance.
You will be shown:
  1. CONTEXT — prior measurements, each labeled by its relation to the target.
  2. A TARGET HYPOTHESIS — a claim of the form "gene G is a dependency in cell line C".
Weigh the evidence and report your degree of belief that a held-out genome-wide CRISPR screen would support the target hypothesis. Supported means the screen scores G at Chronos gene-effect <= -0.5 AND P(dependency) >= 0.5.
Respond with ONLY a single line in this exact format — no reasoning, no explanation, no other text before or after:
<probability>0.XX</probability>
\end{PromptText}
\end{promptbox}

\begin{promptbox}[promptuser]{User prompt \textnormal{(episode \texttt{65b4265e-9f1}, 6 stubs, label = not supported)}}
\begin{PromptText}
CONTEXT EVIDENCE
================

[1] somatic mutation status of the target gene in the target cell line
Somatic mutation status of DIS3L in MERO25
      Gene: DIS3L — DIS3 like exosome 3'-5' exoribonuclease (family: Exosome complex)
      Cell line: MERO25 (DepMap ACH-001556) — Pleura / Pleural Mesothelioma / Pleural Mesothelioma, Biphasic Type; Male; Primary
      Result: wild-type

[2] an unrelated gene knocked out in the target cell line
CRISPR knockout of KRT25 in MERO25
      Gene: KRT25 — keratin 25 (family: Keratins, type I)
      Cell line: MERO25 (DepMap ACH-001556) — Pleura / Pleural Mesothelioma / Pleural Mesothelioma, Biphasic Type; Male; Primary
      Result:
        Chronos gene-effect  -0.02  (no clear fitness effect)
        P(dependency)        0.03   (almost certainly not a dependency)

[... 4 further evidence stubs omitted ...]

TARGET HYPOTHESIS
=================
Gene DIS3L is a dependency (essential for fitness) in cell line MERO25: a held-out genome-wide CRISPR knockout screen of MERO25 would score gene DIS3L at Chronos gene-effect <= -0.5 AND P(dependency) >= 0.5.
  Gene: DIS3L — DIS3 like exosome 3'-5' exoribonuclease (family: Exosome complex)
  Cell line: MERO25 (DepMap ACH-001556) — Pleura / Pleural Mesothelioma / Pleural Mesothelioma, Biphasic Type; Male; Primary

Respond using the format specified in the system message:
  <probability>0.XX</probability>
\end{PromptText}
\end{promptbox}

\subsection{GWAS: genetic associations}

Episode \texttt{6b7db1d8-f2b} asks whether \texttt{rs2238689} (T allele) decreases a phospholipid ratio measurement. The held-out study supports the hypothesis at genome-wide significance.

GWAS relevance depends on locus and trait similarity: a stub may share the target locus, trait category, both, or neither. The omitted stubs include a missense variant $2{,}731$ bp away in the same gene, associated at $p=4\times10^{-15}$ with a trait in the target's own EFO parent, and one association reported at $p=8\times10^{-8}$, below the genome-wide significance threshold.

\begin{promptbox}[promptsystem]{System prompt}
\begin{PromptText}
You are an expert geneticist and epidemiologist analyzing genetic-association evidence from the NHGRI-EBI GWAS Catalog, a curated collection of published genome-wide association studies (GWAS) of human traits and diseases.
How the evidence was generated:
  - A GWAS genotypes many individuals and tests each common variant (SNP) for association with a trait, reporting an effect size and a p-value per SNP.
  - An association is called genome-wide significant at p < 5e-8, the threshold that corrects for ~1 million independent tests across the genome.
  - Each study reports, for the lead SNP, the risk allele whose effect is measured, its frequency, and the direction/size of effect (odds ratio for disease traits, beta for quantitative traits).
Quantity definitions:
  - p-value: statistical significance of the SNP-trait association; < 5e-8 is genome-wide significant.
  - Risk allele: the allele whose effect is reported; different studies may report opposite alleles for the same SNP, so direction is always stated relative to the named allele.
  - OR (odds ratio, binary/disease traits): OR > 1 means the risk allele increases disease odds, OR < 1 means it decreases them (protective).
  - Beta (quantitative traits): the named allele raises the trait when the direction is 'increase', lowers it when 'decrease'.
  - RAF: risk-allele frequency in the study population.
  - Mapped gene: gene assigned to the SNP by genomic position (not a causal claim); EFO parent: the Experimental Factor Ontology category grouping related traits.
You will be shown:
  1. CONTEXT — prior GWAS associations, each labeled by its relation to the target.
  2. A TARGET HYPOTHESIS — a claim that a SNP influences a trait in a stated direction.
Weigh the evidence and report your degree of belief that a held-out GWAS would support the target hypothesis. Supported means the held-out study reports p < 5e-8 for this SNP-trait pair with a consistent direction of effect.
Respond with ONLY a single line in this exact format — no reasoning, no explanation, no other text before or after:
<probability>0.XX</probability>
\end{PromptText}
\end{promptbox}

\begin{promptbox}[promptuser]{User prompt \textnormal{(episode \texttt{6b7db1d8-f2b}, 6 stubs, label = supported)}}
\begin{PromptText}
CONTEXT EVIDENCE
================

[1] the target SNP, associated with a trait in a different EFO parent
rs2238689 for type 2 diabetes mellitus
      SNP: rs2238689 — chr19:45675403
        Risk allele: C
        RAF: 0.4177
        Mapped gene: GIPR
        Functional class: intron_variant
      Gene context:
        Reactome pathways: G alpha (s) signalling events, Glucagon-type ligand receptors
        GTEx expression: Cells - EBV-transformed lymphocytes 23.8 TPM, Adipose - Visceral (Omentum) 10.8 TPM, Heart - Atrial Appendage 9.3 TPM
      Study: GCST009379
        Ancestry: European
        Sample size: 898,130
        Platform: Affymetrix, Illumina [~ 27000000] (imputed)
      Result: p = 5.0e-09 (genome-wide significant); OR/beta = 1.04
        95% CI: [1.03-1.05]
        Direction: C allele increases type 2 diabetes mellitus

[2] an unrelated SNP and trait (different gene, different EFO parent)
rs12324089 for testosterone measurement
      SNP: rs12324089 — chr15:64025657
        Risk allele: G
        RAF: 0.409
        Mapped gene: DAPK2
        Functional class: intron_variant
      Gene context:
        Reactome pathways: Caspase activation via Dependence Receptors in the absence of ligand
        GTEx expression: Thyroid 42.0 TPM, Muscle - Skeletal 19.2 TPM, Lung 16.1 TPM
      Study: GCST90483487
        Ancestry: European
        Sample size: 235,096
        Platform: NR [23899339] (imputed)
      Result: p = 4.0e-08 (genome-wide significant); OR/beta = 0.0152
        95% CI: [0.0097-0.0207] unit decrease
        Direction: G allele decreases testosterone measurement

[... 4 further evidence stubs omitted ...]

TARGET HYPOTHESIS
=================
SNP rs2238689 (T allele) decreases phospholipids:total lipids ratio, high density lipoprotein cholesterol measurement.
A held-out GWAS would report p < 5e-8 for this SNP-trait pair with a consistent direction of effect.
  SNP: rs2238689 — chr19:45675403
    Risk allele: T
    RAF: 0.600605
    Mapped gene: GIPR
    Functional class: intron_variant
  Gene context:
    Reactome pathways: G alpha (s) signalling events, Glucagon-type ligand receptors
    GTEx expression: Cells - EBV-transformed lymphocytes 23.8 TPM, Adipose - Visceral (Omentum) 10.8 TPM, Heart - Atrial Appendage 9.3 TPM
  Trait: phospholipids:total lipids ratio, high density lipoprotein cholesterol measurement
    EFO parent: Other measurement

Respond using the format specified in the system message:
  <probability>0.XX</probability>
\end{PromptText}
\end{promptbox}

\FloatBarrier

\section{A thinking-mode reasoning trace}
\label{app:trace}

\modelname{} (thinking) emits a reasoning trace before committing to a probability. This appendix reproduces one complete trace and the episode that produced it. The rollout below is a single sample from the trained policy, on an episode from the Tahoe validation split rather than the test set.

Episode \texttt{6a6866fd-49a} asks whether belumosudil, a ROCK1/ROCK2 inhibitor, downregulates \texttt{SGMS2} in the colorectal line \texttt{HCT15}. The prompt renders seven evidence stubs. Two measure the same drug and gene in other cell lines, and both report \emph{up}regulation: $+0.833$ at $\text{padj} = 1.7\times10^{-17}$ in \texttt{HOP62}, and $+0.146$ at $\text{padj} = 0.045$ in \texttt{HT-29}. One measures the target cell line under a different drug and is null, as are the four remaining stubs. The held-out measurement refutes the hypothesis rather than leaving it unresolved: belumosudil significantly \emph{up}regulates \texttt{SGMS2} in \texttt{HCT15} ($+0.554$, $\text{padj} = 0.0056$). The trace reads the same-drug stubs as evidence against the claimed direction and settles at $0.20$.

The trace also questions one of the prompt's relevance tags. Stub~[4] is labeled as sharing the target drug's mechanism, a match that holds on the coarse MOA field, where both drugs read \texttt{inhibitor/antagonist}, but not below it: artesunate's fine MOA is a JAK/STAT inhibitor, and belumosudil's listed targets are ROCK1 and ROCK2. The trace registers the doubt (``same mechanism as Belumosudil? Or not sure'') and sets the stub aside as null.

The trace is highly redundant. The conclusion that the same-drug stubs argue against a downregulation claim is restated at least four times, and the answer is approached, qualified and revisited before it is committed.

Rendering follows the conventions of \cref{app:prompts}, except that the completion is set as prose with its paragraph breaks preserved. One drug name is transliterated (\texttt{18beta-Glycyrrhetinic acid}) because the document font cannot set its Greek letter. The \texttt{<think>} tags enclosing the trace are dropped. The text is otherwise unchanged, and the evidence pool is shown in full.

\subsection{Prompt}

The system prompt is the Tahoe system prompt of \cref{app:prompts} carrying the thinking instruction in place of the direct one, with a further paragraph inserted before it that names considerations to weigh:

\begin{promptbox}[promptsystem]{System prompt \textnormal{(addition over \cref{app:prompts})}}
\begin{PromptText}
Reason about what the context implies — for example: the drug's transcriptional footprint in this and nearby cell lines, whether mechanistically similar drugs show consistent effects on the target gene, whether the gene appears broadly responsive or quiet, the drug's aggregate perturbation strength in the target cell line, and any contradictions between evidence pieces. Then commit to a single probability in the closed interval [0, 1].
\end{PromptText}
\end{promptbox}

\begin{promptbox}[promptuser]{User prompt \textnormal{(episode \texttt{6a6866fd-49a}, 7 stubs, label = not supported)}}
\begin{PromptText}
CONTEXT EVIDENCE
================

[1] unrelated drug × cell line, included as a baseline
    Drug: Anethole trithione
      MOA (broad)         unclear
      MOA (fine)          unclear
      Human-approved      no
      In clinical trials  yes
      Notes               Investigated for dry mouth in clinical trials.
      Canonical SMILES    COC1=CC=C(C=C1)C2=CC(=S)SS2
      PubChem CID         2194
    Cell line: hTERT-HPNE
      Cellosaurus ID       CVCL_C466
      Organ                Pancreas
      Driver gene          TERT
      Driver variant type  Transgene
      Driver gene type     Oncogene
    Gene: SGMS2
      Ensembl ID  ENSG00000164023
    Experiment:
      Plate                                              4
      n_cells treated                                    2002
      n_cells control                                    2917
      baseMean (avg normalized counts across both arms)  26.75
    Result:
      log2FoldChange      +0.115 ± 0.334  (upregulates)
      Wald statistic      0.3436
      p-value (raw)       0.7312
      p-value (adjusted)  0.9005

[2] unrelated drug × cell line, included as a baseline
    Drug: DT-061
      MOA (broad)         activator/agonist
      MOA (fine)          unclear
      Human-approved      no
      In clinical trials  yes
      Notes               Investigated for oncology, no approval yet for human use.
      Canonical SMILES    C1CC(C(C(C1)N2C3=CC=CC=C3OC4=CC=CC=C42)O)NS(=O)(=O)C5=CC=C(C=C5)OC(F)(F)F
      PubChem CID         91885558
    Cell line: CHP-212
      DepMap ID                   ACH-000120
      Cellosaurus ID              CVCL_1125
      Organ                       Peripheral Nervous System
      Driver gene                 NRAS
      Driver variant type         Missense
      Driver zygosity             Het
      Driver protein/cDNA effect  p.Q61K
      Driver mechanism (LoF/GoF)  GoF
      Driver gene type            Oncogene
    Gene: SGMS2
      Ensembl ID  ENSG00000164023
    Experiment:
      Plate                                              2
      n_cells treated                                    524
      n_cells control                                    917
      baseMean (avg normalized counts across both arms)  6.928
    Result:
      log2FoldChange      +0.285 ± 0.552  (upregulates)
      Wald statistic      0.5167
      p-value (raw)       0.6053
      p-value (adjusted)  0.9021

[3] same drug as target, different cell line, target gene
    Drug: Belumosudil (mesylate)
      MOA (broad)         inhibitor/antagonist
      MOA (fine)          unclear
      Targets             ROCK2, ROCK1
      Human-approved      yes
      In clinical trials  yes
      Notes               Approved for chronic graft-versus-host disease.
      Canonical SMILES    CC(C)NC(=O)COC1=CC=CC(=C1)C2=NC3=CC=CC=C3C(=N2)NC4=CC5=C(C=C4)NN=C5.CS(=O)(=O)O
      PubChem CID         146681181
    Cell line: HOP62
      DepMap ID                   ACH-000861
      Cellosaurus ID              CVCL_1285
      Organ                       Lung
      Driver gene                 B2M
      Driver variant type         Missense
      Driver zygosity             Het
      Driver protein/cDNA effect  p.E70Q
      Driver gene type            Suppressor
    Gene: SGMS2
      Ensembl ID  ENSG00000164023
    Experiment:
      Plate                                              2
      n_cells treated                                    2743
      n_cells control                                    8616
      baseMean (avg normalized counts across both arms)  292.5
    Result:
      log2FoldChange      +0.833 ± 0.093  (upregulates)
      Wald statistic      8.991
      p-value (raw)       2.45e-19
      p-value (adjusted)  1.67e-17

[4] drug with same mechanism of action as the target drug
    Drug: Artesunate
      MOA (broad)         inhibitor/antagonist
      MOA (fine)          JAK/STAT inhibitor
      Human-approved      yes
      In clinical trials  yes
      Notes               Used for severe malaria, often partnered with other antimalarial drugs.
      Canonical SMILES    CC1CCC2C(C(OC3C24C1CCC(O3)(OO4)C)OC(=O)CCC(=O)O)C
      PubChem CID         6917864
    Cell line: RPMI-7951
      DepMap ID                   ACH-000348
      Cellosaurus ID              CVCL_1666
      Organ                       Skin
      Driver gene                 BRAF
      Driver variant type         Missense
      Driver zygosity             Het
      Driver protein/cDNA effect  p.V600E
      Driver mechanism (LoF/GoF)  GoF
      Driver gene type            Oncogene
    Gene: SGMS2
      Ensembl ID  ENSG00000164023
    Experiment:
      Plate                                              4
      n_cells treated                                    1577
      n_cells control                                    2477
      baseMean (avg normalized counts across both arms)  22.73
    Result:
      log2FoldChange      +0.200 ± 0.361  (upregulates)
      Wald statistic      0.555
      p-value (raw)       0.5789
      p-value (adjusted)  0.8409

[5] unrelated drug × cell line, included as a baseline
    Drug: 18beta-Glycyrrhetinic acid
      MOA (broad)         unclear
      MOA (fine)          unclear
      Human-approved      no
      In clinical trials  yes
      Notes               Used in cosmetics, derived from licorice root, no approval for human drugs.
      Canonical SMILES C[C@]12CC[C@](C[C@H]1C3=CC(=O)[C@@H]4[C@]5(CC[C@@H](C([C@@H]5CC[C@]4([C@@]3(CC2)C)C)(C)C)O)C)(C)C(=O)O
      PubChem CID         10114
    Cell line: HS-578T
      DepMap ID                   ACH-000148
      Cellosaurus ID              CVCL_0332
      Organ                       Breast
      Driver gene                 CDKN2A
      Driver variant type         Deletion
      Driver zygosity             Hom
      Driver protein/cDNA effect  DEL
      Driver mechanism (LoF/GoF)  LoF
      Driver gene type            Suppressor
    Gene: SGMS2
      Ensembl ID  ENSG00000164023
    Experiment:
      Plate                                              4
      n_cells treated                                    857
      n_cells control                                    2654
      baseMean (avg normalized counts across both arms)  43.66
    Result:
      log2FoldChange      +0.255 ± 0.218  (upregulates)
      Wald statistic      1.17
      p-value (raw)       0.242
      p-value (adjusted)  0.6025

[6] same drug as target, different cell line, target gene
    Drug: Belumosudil (mesylate)
      MOA (broad)         inhibitor/antagonist
      MOA (fine)          unclear
      Targets             ROCK2, ROCK1
      Human-approved      yes
      In clinical trials  yes
      Notes               Approved for chronic graft-versus-host disease.
      Canonical SMILES    CC(C)NC(=O)COC1=CC=CC(=C1)C2=NC3=CC=CC=C3C(=N2)NC4=CC5=C(C=C4)NN=C5.CS(=O)(=O)O
      PubChem CID         146681181
    Cell line: HT-29
      DepMap ID                   ACH-000552
      Cellosaurus ID              CVCL_0320
      Organ                       Bowel
      Driver gene                 APC
      Driver variant type         Stopgain
      Driver zygosity             Het
      Driver protein/cDNA effect  p.E853*
      Driver mechanism (LoF/GoF)  LoF
      Driver gene type            Suppressor
    Gene: SGMS2
      Ensembl ID  ENSG00000164023
    Experiment:
      Plate                                              2
      n_cells treated                                    2179
      n_cells control                                    5453
      baseMean (avg normalized counts across both arms)  1.18e+03
    Result:
      log2FoldChange      +0.146 ± 0.056  (upregulates)
      Wald statistic      2.584
      p-value (raw)       0.00976
      p-value (adjusted)  0.04521

[7] different drug, same cell line as target, target gene
    Drug: Adenosine
      MOA (broad)         unclear
      MOA (fine)          unclear
      Targets             ADORA1, ADORA2A, ADORA2B, ADORA3
      Human-approved      yes
      In clinical trials  yes
      Notes               Used for certain arrhythmias; approved as a drug.
      Canonical SMILES    C1=NC(=C2C(=N1)N(C=N2)C3C(C(C(O3)CO)O)O)N
      PubChem CID         60961
    Cell line: HCT15
      DepMap ID                   ACH-000997
      Cellosaurus ID              CVCL_0292
      Organ                       Bowel
      Driver gene                 ACVR2A
      Driver variant type         Frameshift
      Driver zygosity             Hom
      Driver protein/cDNA effect  p.K437fs
      Driver mechanism (LoF/GoF)  LoF
      Driver gene type            Suppressor
    Gene: SGMS2
      Ensembl ID  ENSG00000164023
    Experiment:
      Plate                                              4
      n_cells treated                                    925
      n_cells control                                    2226
      baseMean (avg normalized counts across both arms)  83.5
    Result:
      log2FoldChange      +0.142 ± 0.205  (upregulates)
      Wald statistic      0.6909
      p-value (raw)       0.4897
      p-value (adjusted)  0.7822

TARGET HYPOTHESIS
=================
Drug Belumosudil (mesylate) downregulates gene SGMS2 in cell line HCT15.

    Drug: Belumosudil (mesylate)
      MOA (broad)         inhibitor/antagonist
      MOA (fine)          unclear
      Targets             ROCK2, ROCK1
      Human-approved      yes
      In clinical trials  yes
      Notes               Approved for chronic graft-versus-host disease.
      Canonical SMILES    CC(C)NC(=O)COC1=CC=CC(=C1)C2=NC3=CC=CC=C3C(=N2)NC4=CC5=C(C=C4)NN=C5.CS(=O)(=O)O
      PubChem CID         146681181
    Cell line: HCT15
      DepMap ID                   ACH-000997
      Cellosaurus ID              CVCL_0292
      Organ                       Bowel
      Driver gene                 ACVR2A
      Driver variant type         Frameshift
      Driver zygosity             Hom
      Driver protein/cDNA effect  p.K437fs
      Driver mechanism (LoF/GoF)  LoF
      Driver gene type            Suppressor
    Gene: SGMS2
      Ensembl ID  ENSG00000164023

Respond using the format specified in the system message:
  <probability>0.XX</probability>
\end{PromptText}
\end{promptbox}

\subsection{Trace}

\begin{promptbox}[promptmodel]{Completion \textnormal{(reasoning trace, verbatim)}}
\begin{prosecompletion}
Okay, let's tackle this step by step. So the task is to determine the probability that the target hypothesis (Belumosudil downregulates SGMS2 in HCT15) is supported based on the context provided. The context has seven previous experiments, and I need to analyze each to see how they relate to the target hypothesis.

First, let's understand the target hypothesis. The drug is Belumosudil, which is an inhibitor/antagonist targeting ROCK2 and ROCK1. The cell line is HCT15, which has a frameshift mutation in ACVR2A, a suppressor gene. The gene in question is SGMS2, and the hypothesis is that Belumosudil downregulates it. To support this, the experiment needs to have a padj \textless{} 0.05 and \textbar{}log2FC\textbar{} \textgreater{} 0.5 in the direction of downregulation.

Now, looking at the context. There are several experiments with Belumosudil (same drug) in different cell lines. Let's check those first.

Context [3] is Belumosudil in HOP62 cell line. The result here is a log2FC of +0.833, which is a strong upregulation (padj is very low, 1.67e-17). So this is a clear upregulation of SGMS2. But the target hypothesis here is a downregulation. That's a contradiction. However, the cell line here is HOP62, which is different from HCT15.

Context [6] is Belumosudil in HT-29, another cell line. The log2FC here is +0.146, which is a slight upregulation, and the padj is 0.04521 (just under 0.05, so it's significant). But again, it's an upregulation, not downregulation. So again, the same drug in different cell lines shows upregulation.

Then there's context [7], which is a different drug (Adenosine) in the same HCT15 cell line. The log2FC here is +0.142, which is a slight upregulation, but the padj is 0.7822, which is not significant. So that doesn't support either direction.

Other contexts [1], [2], [4], [5] are unrelated drugs, but maybe they can give some general idea. For example, [4] is Artesunate (same mechanism as Belumosudil? Or not sure). The result here is +0.20, which is upregulation, but not significant (padj 0.8409). Not helpful here.

Now, the key is to see if there's any evidence in the context that supports downregulation of SGMS2 in HCT15 or similar cell lines. The context for Belumosudil in different cell lines (3 and 6) both show upregulation, but the target hypothesis is about downregulation. So this might suggest that the drug is more likely to upregulate, which would make the target hypothesis less probable.

But wait, the context for HCT15 with Adenosine (context [7]) doesn't show significant upregulation. So maybe the gene is not always responsive, but the drug in other cell lines does show upregulation. However, the target hypothesis is about HCT15. Since the context for Belumosudil in other cell lines shows upregulation, but the target hypothesis is about a different direction (downregulation), that might suggest that the target hypothesis is less likely.

However, I should also check if there's any evidence in the context that directly contradicts the target hypothesis. For example, if other experiments in the same cell line or similar cell lines show downregulation, but in this case, none of the other contexts for HCT15 show that. The same drug in other cell lines shows upregulation, and the unrelated drug in HCT15 (context [7]) isn't significant.

But I need to consider the strength of the context. The context for the same drug (Belumosudil) in different cell lines (HOP62 and HT-29) both show upregulation. That's a pattern, but the target hypothesis here is for HCT15. The question is whether this pattern in other cell lines supports or contradicts the hypothesis here.

Another thing to consider is the gene's responsiveness. The gene SGMS2 is being tested in multiple contexts, and in the same drug's context, it's upregulated in different cell lines. That suggests that the gene might be a consistent target for this drug. However, the target hypothesis here is the opposite. So this would make the probability of the target hypothesis being supported lower.

But what about the other contexts? For example, context [1] and [2] are unrelated drugs, but they also show upregulation. However, those are not directly relevant. The key is the same drug in different cell lines.

Also, the base mean in the context experiments for the same drug (context [3] and [6]) are higher, which might indicate more reliable data. But the target hypothesis here is for HCT15.

So, considering the context, the same drug in different cell lines shows upregulation of SGMS2. The target hypothesis is that it's downregulated in HCT15. Therefore, the context suggests that the drug is more likely to upregulate, which would make the target hypothesis (downregulation) less probable. However, the context doesn't have any direct evidence for the same cell line (HCT15) and the same drug.

But the problem is to weigh the evidence. The fact that in two different cell lines, the same drug upregulates the gene might suggest that the gene is a target of the drug. If the target hypothesis here is in the opposite direction, that might lower the probability. However, there's no direct evidence against it.

But in the context, the same drug in different cell lines shows upregulation, which is a positive sign for the direction. But the target hypothesis here is for downregulation, which is opposite. So that would make the probability of the target hypothesis being supported lower.

But I need to be careful here. Are there any other factors? For example, the driver mutations in the cell lines. The HCT15 cell line has a LoF mutation in ACVR2A. But the other cell lines in context [3] (HOP62) has a B2M mutation (suppressor), and HT-29 (context [6]) has APC (LoF). So the cell lines vary in their driver mutations. But the same drug in different cell lines still shows upregulation.

Therefore, based on the context, the evidence for the same drug in other cell lines showing upregulation of SGMS2 would make the target hypothesis (downregulation) less likely. However, the context doesn't have any experiments that directly contradict it.

But how does that translate into a probability? If the context shows that in two different cell lines, the same drug upregulates the gene, that's a pattern. The target hypothesis is the opposite. So the probability might be low, say 0.2 or 0.3. But I need to think carefully.

Alternatively, maybe the context doesn't have enough evidence to make a strong conclusion. However, since the same drug in other cell lines shows upregulation, it's more likely that the gene is upregulated, not downregulated. Therefore, the probability that the target hypothesis (downregulation) is supported is low.

But I need to consider the possibility that the target hypothesis could still be correct. For example, if there's a cell line-specific effect. But the context doesn't have any evidence for that here.

So, based on the context provided, I would say that the probability is around 0.2 (20\% chance that the target hypothesis is supported). But I need to be precise here.
\end{prosecompletion}
\end{promptbox}

\begin{promptbox}[promptmodel]{Answer}
\begin{PromptText}
<probability>0.20</probability>
\end{PromptText}
\end{promptbox}

\FloatBarrier

\section{Additional analyses of belief revision}
\label{app:legacy}

This appendix reports the two probes of belief revision summarized in \cref{sec:revision}, Analysis~3. Both use the multi-domain test set and the current roster, with $256$ episodes per domain.

The preceding analyses ask whether individual updates have sensible signs and magnitudes; these probes ask whether a model combines the same evidence consistently. If the stated probabilities behave like beliefs, rearranging a fixed evidence set should not materially change the result, and the joint weight of evidence of a matched pair should be predictable from the two single-stub weights. Both probes hold evidence content fixed while changing its position or composition.

\begin{figure}[htbp]\centering
\includegraphics[width=\textwidth]{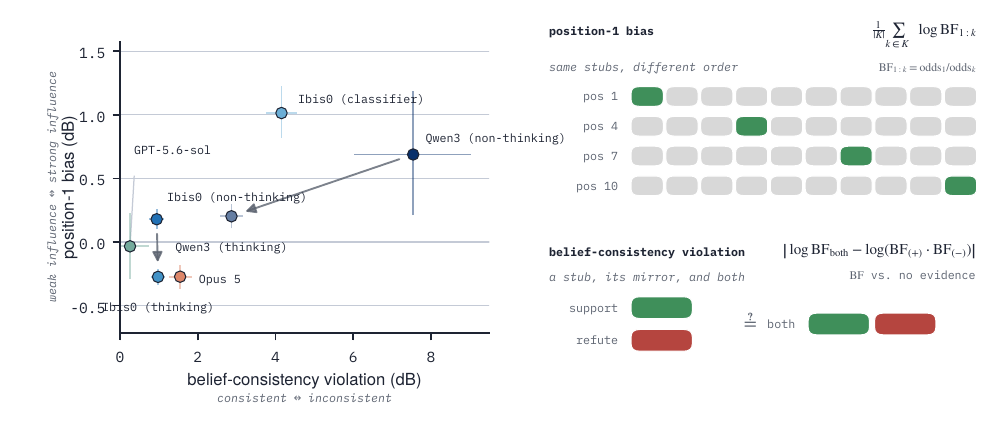}
\caption{\textbf{Ordering effects: Given the same evidence, are models consistent in belief revision?} Each point is one model; the schematic at right shows the two manipulations. \textbf{Position-1 bias} (vertical): a single ``compelling'' evidence stub (supporting and highly-relevant) is placed at varying positions within an otherwise identical context. We report the stated belief when the stub appears first minus the mean stated belief when it appears in the later three positions, in decibans. Positive values indicate that the same stub moves the belief further when it appears first, a form of primacy bias. \textbf{Belief-consistency violation} (horizontal): a matched pair of supporting/refuting stubs is constructed by inverting the direction of effect while keeping all other details the same, and we measure how far the joint update departs from the product of the two single-stub updates. Zero on either axis is the unbiased value, so the lower left is best. Arrows join the two matched pairs that differ only in whether a reasoning trace is verbalized. Across both measures, thinking reduces bias. \modelname{} (classifier) is well calibrated yet shows pronounced bias on both axes, whereas RL post-training improves calibration \textit{and} reduces the observed biases.}
\label{fig:belief-revision}
\end{figure}

The \textbf{position probe} places one compelling stub at four locations in an otherwise identical ten-stub context. The reported position-1 bias is the belief when that stub appears first minus the mean belief when it appears later. The \textbf{composition probe} presents a supporting stub, a constructed equal-and-opposite refuting stub, and the pair; its residual measures departure from adding the two single-stub weights of evidence; it covers Tahoe and DepMap only. \Cref{sec:conflict,sec:position} give the construction and results of each probe, and \cref{fig:belief-revision} summarizes both.

Consistency does not follow from aggregate forecast quality (\cref{fig:belief-revision}): the classifier has the best Brier score of the \modelname{} models but the largest position-1 bias ($+1.01$~dB) and composition residual ($4.16$~dB). Post-training still improves consistency: all three trained variants reduce the Qwen3 (non-thinking) composition residual of $7.54$~dB. The classifier reaches $4.16$~dB, and the two RL policies reach $0.94$~dB (\modelname{} non-thinking) and $0.98$~dB (\modelname{} thinking), in the range of the frontier models (GPT-5.6-sol $0.26$~dB, Claude Opus 5 $1.55$~dB).

Reasoning traces are associated with a separate shift in position dependence. Within both the Qwen3 and \modelname{} pairs, enabling a trace reduces position-1 bias by about $0.5$~dB. Because the same shift appears before and after post-training, it is more naturally associated with the response process than with the proper-scoring objective alone.

\FloatBarrier
\subsection{Composition of supporting and refuting evidence}
\label{sec:conflict}

The conflict probe selects the clearest supporting stub in an episode and constructs a matched refuting stub of equal magnitude. In Tahoe this reverses the signs of the differential-expression fold change and Wald statistic, leaving the adjusted $p$-value unchanged. In DepMap it reflects the Chronos gene effect about zero and reads the accompanying dependency probability off the empirical relation between the two quantities in the test set, so that both numbers on the rendered card move together. GWAS is excluded because the catalog records published associations but not failures to associate, which a refuting stub would require. The same property restricts the directional analysis in \cref{sec:woe}.

The supporting stub, the refuting stub, and the pair are presented separately. The pair is presented in both orders and the two are averaged, so that order effects cancel and do not enter the residual. Departure from additive weight-of-evidence updating is measured as
\[
  \left|\log \mathrm{BF}_{\mathrm{both}}
  -\log\left(\mathrm{BF}_{\mathrm{support}}\mathrm{BF}_{\mathrm{refute}}\right)\right| ,
\]
where each Bayes factor is taken relative to the same episode with no evidence. Zero denotes exact agreement between the paired update and the sum of the two single-stub updates. The reference for each model is therefore its own single-stub responses.

\begin{figure}[htbp]\centering
\includegraphics[width=\textwidth]{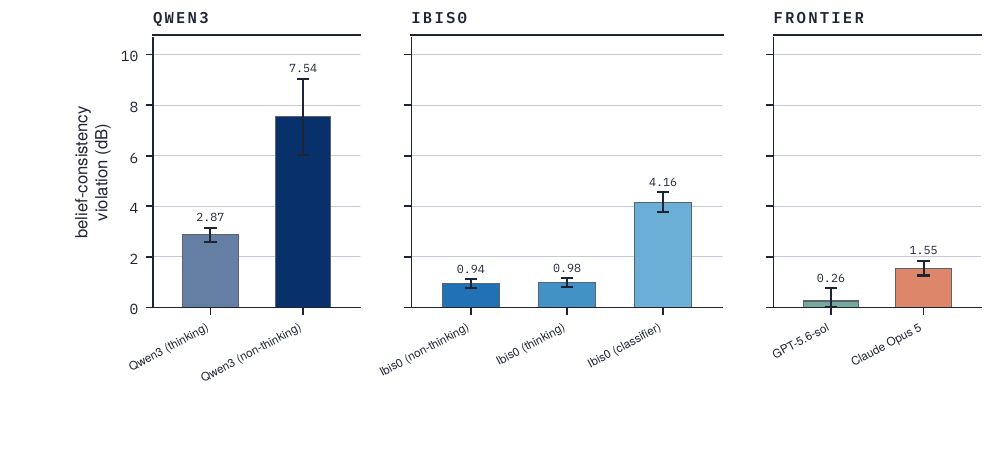}
\caption{\textbf{Composition of conflicting evidence.} Absolute departure from additive weight-of-evidence updating for a supporting stub and its constructed equal-and-opposite counterpart, macro-averaged over Tahoe and DepMap. Panels group models by training regime and bars are ordered within each panel; the vertical scale is shared. Intervals are $95\%$ episode bootstrap intervals with episodes resampled within domain, $256$ episodes per domain. Lower values indicate closer agreement with multiplicative combination of Bayes factors.}
\label{fig:conflict}
\end{figure}

\Cref{fig:conflict} reports the result over $256$ episodes in each of Tahoe and DepMap, grouped by training regime. Within Qwen3, the violation is $7.54$~dB $[6.03, 9.04]$ in non-thinking mode and $2.87$~dB $[2.58, 3.16]$ in thinking mode. Among the \modelname{} models it is $0.94$~dB $[0.75, 1.13]$ for \modelname{} (non-thinking) and $0.98$~dB $[0.80, 1.15]$ for \modelname{} (thinking), which differ from each other by less than their intervals, and $4.16$~dB $[3.76, 4.55]$ for \modelname{} (classifier). The two frontier models give $0.26$~dB $[0.01, 0.75]$ for GPT-5.6-sol and $1.55$~dB $[1.27, 1.85]$ for Claude Opus 5.

The clearest effect in the figure is that both reinforcement-learning policies are far more consistent than the Qwen3 checkpoint they were trained from. The classifier is not: it violates consistency more than either frontier model and both policies, while achieving the lowest Brier score of any model in \cref{sec:primary}. Post-training against a proper scoring rule therefore does not improve this property on its own, and the output channel separates the models that share a training objective. GPT-5.6-sol is the most consistent model measured, so post-training on \ourbenchmark{} is not required to obtain the property.

\subsection{Dependence of evidence updates on sequence position}
\label{sec:position}

The position probe inserts the same evidence stub at positions $1$, $4$, $7$, and $10$ of a ten-stub context and fills the remaining slots with the same neutral padding in the same order. The four conditions contain the same evidence in different arrangements, and an order-invariant model reports the same probability in all four. The context holds ten stubs rather than the sixteen of the earlier pilot because DepMap episodes carry about eleven stubs that can serve as padding; a sixteen-stub context can be constructed for only $14$ of $2{,}000$ episodes.

Reported values are the stated belief in decibans, centred within episode across the four positions to remove variation in how compelling each episode's probe stub is. The statistic of interest is the belief at the first position minus the mean over the later three. Centring cancels in any contrast between positions, so this quantity is a raw within-episode difference in log-odds and does not depend on which positions were sampled. A contrast between the first and last positions alone would not suffice: for \modelname{} (non-thinking) the last position is as elevated as the first, and the two cancel.

\Cref{tab:position-bias} gives the statistic for each model. It is positive and separated from zero for \modelname{} (classifier), both modes of Qwen3, and \modelname{} (non-thinking), so in those four models the same stub moves the belief further when it appears first. It is negative and separated from zero for \modelname{} (thinking) and Claude Opus 5, and contains zero for GPT-5.6-sol.

\begin{table}[t]
\centering
\small
\begin{tabular}{lrr}
\toprule
Model & Position-1 bias (dB) & 95\% CI \\
\midrule
\modelname{} (classifier) & \textbf{+1.01} & \textcolor{black!45}{$[+0.82, +1.22]$} \\
Qwen3 (non-thinking) & \textbf{+0.69} & \textcolor{black!45}{$[+0.21, +1.18]$} \\
Qwen3 (thinking) & \textbf{+0.20} & \textcolor{black!45}{$[+0.11, +0.30]$} \\
\modelname{} (non-thinking) & \textbf{+0.18} & \textcolor{black!45}{$[+0.10, +0.26]$} \\
GPT-5.6-sol & -0.03 & \textcolor{black!45}{$[-0.29, +0.23]$} \\
Claude Opus 5 & \textbf{-0.27} & \textcolor{black!45}{$[-0.37, -0.18]$} \\
\modelname{} (thinking) & \textbf{-0.27} & \textcolor{black!45}{$[-0.34, -0.21]$} \\
\bottomrule
\end{tabular}
\caption{\textbf{Dependence of the stated belief on the first evidence position.} Stated belief at position $1$ minus the mean over positions $4$, $7$ and $10$, in decibans, macro-averaged over the three domains with $256$ episodes per domain. Positive values indicate that the same stub moves the belief further when it appears first. Intervals are $95\%$ episode bootstrap intervals resampled within domain; bold marks an interval excluding zero.}
\label{tab:position-bias}
\end{table}

The two matched pairs that differ only in whether the model verbalizes a reasoning trace move together. Paired within episode over $768$ episodes, adding the trace changes first-position bias by $-0.49$~dB $[-0.96, -0.01]$ for Qwen3 and by $-0.45$~dB $[-0.54, -0.36]$ for the reinforcement-learning policy. The two estimates agree to within $0.04$~dB although only one pair is trained, which suggests that deliberating before answering and training against a proper scoring rule act on this property separately. Training reduces the effect from $+0.69$~dB for Qwen3 (non-thinking) to $+0.18$~dB for \modelname{} (non-thinking), and the trace shifts it downward from that level; in the reinforcement-learning pair, the shift passes zero into a small preference for later positions. The Qwen3 interval only just excludes zero, so the paired result rests mainly on the policy pair.

The classifier, the only model that states no probability in text, shows the largest effect in the roster. It also achieves the lowest Brier score in \cref{sec:primary}.

\begin{figure}[htbp]\centering
\includegraphics[width=\textwidth]{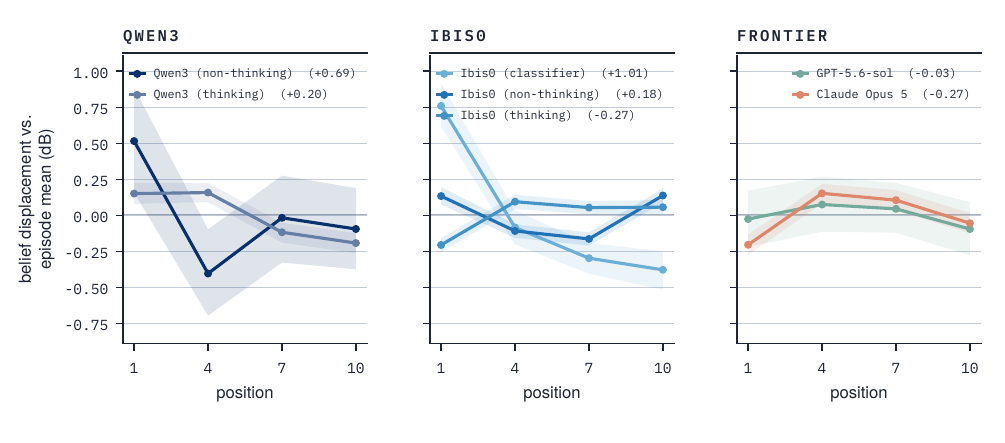}
\caption{\textbf{Dependence of the stated belief on evidence position.} The same evidence stub placed at four positions of a ten-stub context, with the remaining slots held to identical neutral padding in identical order. Panels group models that differ in whether a reasoning trace is verbalized. Values are decibans, centred within episode, macro-averaged over the three domains with episodes resampled within domain; bands are $95\%$ intervals, $256$ episodes per domain. Legend entries give the position-1 bias of \Cref{tab:position-bias}. A flat line at zero indicates a belief that does not depend on the order of the evidence.}
\label{fig:woe_position}
\end{figure}

\section{Structured and numeric-only baselines on Tahoe}
\label{app:tahoe-baselines}

Rhaister~\citep{svensson2026back} predicts a differential-expression outcome for a specified cell, treatment, and gene. We convert its predicted log fold change to a probability by assigning its sign relative to the target direction and applying a sigmoid, and also report a hard-thresholded version. The published episodes do not record drug concentration, so treatment identity is defined by drug. Rhaister can provide a forecast only when its fit data contain the target gene under a compatible treatment.

The \emph{prompt-local} setting fits Rhaister only to the evidence stubs shown in the prompt. The \emph{all-context} setting uses training-split target rows and all structured context evidence, with held-out target rows excluded. A numeric-only gradient-boosted tree is trained on the Tahoe training split using DESeq2 summary statistics and counts, without gene, drug or cell-line identities or text; it produces a forecast for every test episode.

\begin{table}[htbp]
\centering
\small
\begin{tabular}{llccc}
\toprule
Baseline & Fit data & Coverage & Brier ($\downarrow$) & Hard Brier ($\downarrow$) \\
\midrule
Rhaister prompt-local & Prompt evidence only & 85.1\% & 0.2310 & 0.4130 \\
Rhaister all-context & Train targets + context & 88.5\% & 0.2295 & 0.3968 \\
Numeric-only GBM & DESeq2 numeric features & 100.0\% & 0.2206 & n/a \\
\addlinespace
All-domain classifier & Three-domain training & 100.0\% & 0.2063 & n/a \\
\bottomrule
\end{tabular}
\caption{Tahoe-only structured baselines on the $2{,}000$-episode test set. Coverage is the fraction of episodes for which a baseline produces a forecast.}
\label{tab:additional_baselines}
\end{table}

\section{Extended related work}
\label{app:related}

This appendix expands the summary in \cref{sec:related_work}.

\paragraph{Scientific reasoning and evidence benchmarks.}
Benchmarks for scientific AI have moved from expert-written question answering~\citep{rein2023gpqa} toward tasks drawn from research practice: long-context reasoning over scientific documents~\citep{cui2025curie}, research coding~\citep{tian2024scicode}, data-driven discovery from datasets~\citep{majumder2024discoverybench, chen2024scienceagentbench}, simulated environments in which an agent must form hypotheses and run experiments~\citep{wang2022scienceworld, jansen2024discoveryworld}, and biology research workflows spanning literature retrieval, protocol troubleshooting, and study-scale computational analysis~\citep{laurent2024labbench, laurent2026labbench2, mitchener2025bixbench, koch2026bixbench3}. Literature agents answer questions by retrieving and synthesizing published evidence~\citep{skarlinski2024paperqa2}. Closest in form to our task is claim verification, where SciFact and SciFact-Open ask whether a retrieved abstract supports or refutes a claim~\citep{wadden2020scifact, wadden2022scifactopen} and Matter-of-Fact asks whether a literature-supported claim is feasible~\citep{jansen2025matteroffact}. These tasks score a categorical verdict about existing evidence. \ourbenchmark{} instead asks for a probability that a held-out measurement will support the hypothesis, scores it with a proper scoring rule, and controls the relevance and sufficiency of the evidence pool so that belief revision can be measured update by update.

\paragraph{Calibration and confidence estimation in LLMs.}
Work on LLM calibration has centered on confidence in answer \emph{correctness} for factual or multiple-choice question answering, where large models can be reasonably calibrated when probabilities are read from token logits or elicited as $P(\text{true})$~\citep{kadavath2022language, jiang2021know, desai2020calibration, geng2024survey}. A recurring finding is that alignment (RLHF and instruction tuning) degrades this calibration and leaves models systematically overconfident~\citep{openai2023gpt4, zhu2023calibration}, with a corresponding risk that users over-trust confident-sounding output~\citep{rathi2025overrely}. Proposed remedies span verbalized confidence elicitation~\citep{lin2022teaching, tian2023just}, logit- and sequence-likelihood scores~\citep{duan2024sar}, sampling-consistency measures~\citep{wang2023selfconsistency, chen2024bsdetector}, semantic-entropy estimators~\citep{kuhn2023semantic, farquhar2024detecting}, and auxiliary or post-hoc calibrators~\citep{ulmer2024apricot, shen2024thermometer, guo2017calibration}. Test-time reasoning does not reliably improve calibration and can worsen it~\citep{mei2026reasoning}, which motivates our direct-mode elicitation of a single probability. The task closest to ours is probabilistic forecasting scored by strictly proper rules~\citep{gneiting2007strictly, brier1950verification, murphy1973new}, where LLMs now approach human-crowd accuracy on real-world events~\citep{halawi2024forecasting, schoenegger2024silicon}. We elicit calibrated probabilities for \emph{novel scientific hypotheses} under a controlled, manipulable evidence pool, and report discrimination (AUROC) and calibration (ECE~\citep{naeini2015obtaining}) separately rather than accuracy alone.

\paragraph{Proper-scoring post-training.}
A recent line of work post-trains language models against strictly proper scoring rules~\citep{gneiting2007strictly}, using the Brier score or log loss of a verbalized confidence as a reinforcement-learning reward or a supervised objective~\citep{damani2026beyond, puri2026reaching, baniharouni2026rewarding, leng2025taming, xu2024sayself}. These methods target confidence in the correctness of an answer or a reasoning chain on question-answering tasks, so the label is whether the model's own output is right. We apply the same family of objectives to a target external to the model: the outcome of a held-out experiment, conditioned on an evidence pool. The model must weigh evidence instead of assessing its own reliability, and the reward comes from resolved scientific outcomes. We compare a negative-Brier reward in direct and thinking modes with binary cross-entropy on a scalar head (\cref{sec:methods}).

\paragraph{Bayesian experimental design with LLMs.}
Classical Bayesian experimental design chooses the experiment that maximizes expected information gain (EIG)~\citep{lindley1956measure, chaloner1995bayesian, rainforth2024modern}, but estimating EIG is costly, requiring nested Monte-Carlo integration~\citep{rainforth2018nesting}, variational or contrastive bounds~\citep{foster2019variational, poole2019variational}, or amortized design policies~\citep{foster2021deep, blau2022optimizing, mueller2022transformers}. A growing line instead drives information-theoretic selection from a language model's own beliefs: LLMs pick EIG-maximizing queries for interactive information gathering~\citep{hu2024uncertainty, mazzaccara2024learning, kobalczyk2025active, choudhury2026bedllm, grand2026shoot}, elicit user intent by asking informative questions~\citep{li2025eliciting, handa2024bayesian}, serve as uncertainty-bearing surrogates for Bayesian optimization in scientific domains~\citep{ramos2025bayesian, liu2024large}, and plan iterative perturbation screens~\citep{huang2024sequential, hao2025perturboagent}. Prior studies also report that direct prompting for information gathering can reduce performance~\citep{li2024mediq} and that off-the-shelf agents may ask uninformative questions~\citep{grand2026shoot}. The multi-domain evaluation in \cref{sec:bed} applies model-based EIG selection to evidence-deliberation episodes from Tahoe, DepMap, and GWAS.

\paragraph{Belief revision, order effects, and weights of evidence.}
Whether language models revise beliefs coherently has been studied through cognitive-bias probes and Bayesian-updating tasks. Models show primacy effects in multiple-choice and long-context settings~\citep{wang2023primacy, liu2023lost, raimondi2025primacy} and anchor on initial numeric hints~\citep{lou2024anchoring}; surveys catalog these and other human-like biases~\citep{sumita2024cognitive}. On sequential evidence, models approach Bayesian posteriors given enough observations~\citep{gupta2025coinflips} but discount earlier evidence relative to an ideal filter~\citep{zhang2025discounted} and depart from Bayesian invariances even when instructed to follow them~\citep{kim2025evidence}. We measure updates in decibans, the unit of weight of evidence that Good and Turing introduced for sequential evidence~\citep{good1950probability, good1979turing}, and summarize each update as a log Bayes factor~\citep{kass1995bayes}. Whether a model's weights of evidence are the right size, the question of \cref{sec:woe}, has to our knowledge not been asked of language models before; the nearest prior work is support theory's account of how human probability judgments depend on how the evidence is described~\citep{tversky1994support}.

\paragraph{Prediction of perturbation outcomes in biology.}
Predicting the transcriptional response to a perturbation is an active area. GEARS uses gene-interaction graphs to extrapolate to unseen combinations~\citep{roohani2023gears}, and single-cell foundation models such as Geneformer and scGPT are fine-tuned for the task~\citep{theodoris2023geneformer, cui2024scgpt}. Systematic benchmarks find that these models do not yet outperform simple linear baselines~\citep{ahlmann2025linear, wei2026benchmarking}. The virtual-cell program aims at models that predict perturbed cell state across contexts~\citep{bunne2024virtualcell, roohani2025vcc, adduri2025state}, and perturbation atlases such as Tahoe-100M supply the training data~\citep{tahoe100m}. These models predict the measurement itself from molecular inputs. \ourbenchmark{} asks a different question: given prior measurements as text, how confident should a reader be that the next one will support a hypothesis. The two are complementary, and a perturbation model could serve as the outcome simulator in our experimental-design loop (\cref{sec:bed}).

\paragraph{AI for science and autonomous discovery.}
Language-model agents now automate parts of the research cycle, and the systems differ mainly in which parts. One line couples a model to laboratory automation so that it plans and executes experiments~\citep{boiko2023coscientist}. A second generates and ranks hypotheses or research proposals, in biomedicine~\citep{gottweis2026coscientist} and in machine learning, where the loop extends to running the experiments and writing the paper~\citep{lu2024aiscientist, yamada2025aiscientistv2}. A third organizes several agents into a research team that iterates between hypothesis, experiment, and analysis, with wet-lab validation of the resulting designs and candidate treatments~\citep{swanson2025virtuallab, ghareeb2026robin}, and recent systems extend this loop to long-horizon, data-driven discovery~\citep{kosmos}. Each of these systems must decide which hypotheses to pursue and when the evidence suffices, decisions that depend on the model's probabilistic judgment. \ourbenchmark{} measures that judgment in isolation, and the experimental-design results in \cref{sec:bed} show one way to use calibrated belief for better decisions inside such a loop.


@article{brier1950verification,
  title={Verification of forecasts expressed in terms of probability},
  author={Brier, Glenn W.},
  journal={Monthly Weather Review},
  volume={78}, number={1}, pages={1--3}, year={1950},
  doi={10.1175/1520-0493(1950)078%3C0001:VOFEIT%3E2.0.CO;2}
}

@article{murphy1973new,
  title={A new vector partition of the probability score},
  author={Murphy, Allan H.},
  journal={Journal of Applied Meteorology},
  volume={12}, number={4}, pages={595--600}, year={1973}
}

@inproceedings{guo2017calibration,
  title={On calibration of modern neural networks},
  author={Guo, Chuan and Pleiss, Geoff and Sun, Yu and Weinberger, Kilian Q.},
  booktitle={International Conference on Machine Learning (ICML)},
  pages={1321--1330}, year={2017}
}

@article{murphy1977reliability,
  title={Reliability of Subjective Probability Forecasts of Precipitation and Temperature},
  author={Murphy, Allan H. and Winkler, Robert L.},
  journal={Journal of the Royal Statistical Society, Series C (Applied Statistics)},
  volume={26}, number={1}, pages={41--47}, year={1977},
  doi={10.2307/2346866}
}

@article{degroot1983comparison,
  title={The Comparison and Evaluation of Forecasters},
  author={DeGroot, Morris H. and Fienberg, Stephen E.},
  journal={Journal of the Royal Statistical Society, Series D (The Statistician)},
  volume={32}, number={1-2}, pages={12--22}, year={1983},
  doi={10.2307/2987588}
}

@inproceedings{blasiok2024smooth,
  title={Smooth {ECE}: Principled Reliability Diagrams via Kernel Smoothing},
  author={B{\l}asiok, Jaros{\l}aw and Nakkiran, Preetum},
  booktitle={International Conference on Learning Representations (ICLR)}, year={2024}, note={arXiv:2309.12236}
}

@incollection{platt1999probabilistic,
  title={Probabilistic Outputs for Support Vector Machines and Comparisons to Regularized Likelihood Methods},
  author={Platt, John C.},
  booktitle={Advances in Large Margin Classifiers},
  editor={Smola, Alexander J. and Bartlett, Peter and Sch{\"o}lkopf, Bernhard and Schuurmans, Dale},
  pages={61--74}, publisher={MIT Press}, year={1999}
}

@inproceedings{drgrpo,
  title={Understanding {R1}-Zero-Like Training: A Critical Perspective},
  author={Liu, Zichen and Chen, Changyu and Li, Wenjun and Qi, Penghui and Pang, Tianyu and Du, Chao and Lee, Wee Sun and Lin, Min},
  booktitle={Conference on Language Modeling (COLM)}, year={2025}, note={arXiv:2503.20783}
}

@article{dppo,
  title={Rethinking the Trust Region in {LLM} Reinforcement Learning},
  author={Qi, Penghui and Zhou, Xiangxin and Liu, Zichen and Pang, Tianyu and Du, Chao and Lin, Min and Lee, Wee Sun},
  journal={arXiv preprint arXiv:2602.04879}, year={2026}
}

@article{kimik25,
  title={{Kimi K2.5}: Visual Agentic Intelligence},
  author={{Kimi Team}},
  journal={arXiv preprint arXiv:2602.02276}, year={2026}
}

@article{tahoe100m,
  title={{Tahoe-100M}: A Giga-Scale Single-Cell Perturbation Atlas for Context-Dependent Gene Function and Cellular Modeling},
  author={Zhang, Jesse and Ubas, Airol A. and de Borja, Richard and Svensson, Valentine and Thomas, Nicole and Thakar, Neha and Lai, Ian and others},
  journal={bioRxiv}, year={2025},
  doi={10.1101/2025.02.20.639398},
  publisher={Cold Spring Harbor Laboratory}
}

@misc{primerl,
  title={{PRIME-RL}},
  author={{Prime Intellect}},
  year={2025},
  howpublished={\url{https://github.com/PrimeIntellect-ai/prime-rl}},
  note={Software}
}

@article{tversky1994support,
  title={Support theory: A nonextensional representation of subjective probability},
  author={Tversky, Amos and Koehler, Derek J.},
  journal={Psychological Review},
  volume={101}, number={4}, pages={547--567}, year={1994}
}

@article{gneiting2007strictly,
  title={Strictly proper scoring rules, prediction, and estimation},
  author={Gneiting, Tilmann and Raftery, Adrian E.},
  journal={Journal of the American Statistical Association},
  volume={102}, number={477}, pages={359--378}, year={2007}
}

@article{rainforth2024modern,
  title={Modern Bayesian experimental design},
  author={Rainforth, Tom and Foster, Adam and Ivanova, Desi R and Bickford Smith, Freddie},
  journal={Statistical Science},
  volume={39},
  number={1},
  pages={100--114},
  year={2024},
  publisher={Institute of Mathematical Statistics}
}

@misc{depmap,
  title        = {{DepMap} {24Q4} {Public}},
  author       = {{DepMap, Broad}},
  year         = {2024},
  howpublished = {Figshare+ dataset, \url{https://doi.org/10.25452/figshare.plus.27993248.v1}}
}

@article{chronos,
  title   = {{Chronos}: a cell population dynamics model of {CRISPR} experiments that improves inference of gene fitness effects},
  author  = {Dempster, Joshua M. and Boyle, Isabella and Vazquez, Francisca and Root, David E. and Boehm, Jesse S. and Hahn, William C. and Tsherniak, Aviad and McFarland, James M.},
  journal = {Genome Biology},
  volume  = {22},
  pages   = {343},
  year    = {2021},
  doi     = {10.1186/s13059-021-02540-7}
}

@article{prism,
  title   = {Discovering the anticancer potential of non-oncology drugs by systematic viability profiling},
  author  = {Corsello, Steven M. and others},
  journal = {Nature Cancer},
  volume  = {1},
  number  = {2},
  pages   = {235--248},
  year    = {2020},
  doi     = {10.1038/s43018-019-0018-6}
}

@article{gwascatalog,
  title   = {The {NHGRI-EBI} {GWAS} {Catalog}: knowledgebase and deposition resource},
  author  = {Sollis, Elliot and Mosaku, Abayomi and Abid, Ala and Buniello, Annalisa and Cerezo, Maria and Gil, Laurent and Groza, Tudor and G{\"u}ne{\c{s}}, Osman and Hall, Peggy and Hayhurst, James and Ibrahim, Arwa and Ji, Yue and John, Sajo and Lewis, Elizabeth and MacArthur, Jacqueline A. L. and McMahon, Aoife and Osumi-Sutherland, David and Panoutsopoulou, Kalliope and Pendlington, Zo{\"e} and Ramachandran, Santhi and Stefancsik, Ray and Stewart, Jonathan and Whetzel, Patricia and Wilson, Robert and Hindorff, Lucia and Cunningham, Fiona and Lambert, Samuel A. and Inouye, Michael and Parkinson, Helen and Harris, Laura W.},
  journal = {Nucleic Acids Research},
  volume  = {51},
  number  = {D1},
  pages   = {D977--D985},
  year    = {2023}
}

@article{svensson2026back,
  title={Back to basics: Observed statistics are sufficient to predict drug responses},
  author={Svensson, Valentine and Khan, Umair and Heydari, Hamed and Ubas, Airol A and Thomas, Nicole and Merico, Daniele and Goodarzi, Hani and Yu, John and Alidoust, Nima and Gandhi, Shreshth},
  journal={bioRxiv},
  year={2026},
  doi={10.64898/2026.06.09.731197}
}

@article{kadavath2022language,
  title={Language Models (Mostly) Know What They Know},
  author={Kadavath, Saurav and Conerly, Tom and Askell, Amanda and Henighan, Tom and Drain, Dawn and Perez, Ethan and Schiefer, Nicholas and Hatfield-Dodds, Zac and DasSarma, Nova and others},
  journal={arXiv preprint arXiv:2207.05221}, year={2022}
}

@article{jiang2021know,
  title={How Can We Know When Language Models Know? On the Calibration of Language Models for Question Answering},
  author={Jiang, Zhengbao and Araki, Jun and Ding, Haibo and Neubig, Graham},
  journal={Transactions of the Association for Computational Linguistics (TACL)},
  volume={9}, pages={962--977}, year={2021},
  doi={10.1162/tacl_a_00407}
}

@inproceedings{desai2020calibration,
  title={Calibration of Pre-trained Transformers},
  author={Desai, Shrey and Durrett, Greg},
  booktitle={Proceedings of the 2020 Conference on Empirical Methods in Natural Language Processing (EMNLP)},
  pages={295--302}, year={2020},
  doi={10.18653/v1/2020.emnlp-main.21}
}

@inproceedings{geng2024survey,
  title={A Survey of Confidence Estimation and Calibration in Large Language Models},
  author={Geng, Jiahui and Cai, Fengyu and Wang, Yuxia and Koeppl, Heinz and Nakov, Preslav and Gurevych, Iryna},
  booktitle={Proceedings of the 2024 Conference of the North American Chapter of the Association for Computational Linguistics (NAACL)},
  pages={6577--6595}, year={2024}, note={arXiv:2311.08298},
  doi={10.18653/v1/2024.naacl-long.366}
}

@article{openai2023gpt4,
  title={{GPT-4} Technical Report},
  author={OpenAI},
  journal={arXiv preprint arXiv:2303.08774}, year={2023}
}

@inproceedings{zhu2023calibration,
  title={On the Calibration of Large Language Models and Alignment},
  author={Zhu, Chiwei and Xu, Benfeng and Wang, Quan and Zhang, Yongdong and Mao, Zhendong},
  booktitle={Findings of the Association for Computational Linguistics: EMNLP 2023},
  pages={9778--9795}, year={2023}, note={arXiv:2311.13240},
  doi={10.18653/v1/2023.findings-emnlp.654}
}

@inproceedings{rathi2025overrely,
  title={Humans overrely on overconfident language models, across languages},
  author={Rathi, Neil and Jurafsky, Dan and Zhou, Kaitlyn},
  booktitle={Conference on Language Modeling (COLM)}, year={2025}, note={arXiv:2507.06306}
}

@inproceedings{leng2025taming,
  title={Taming Overconfidence in {LLMs}: Reward Calibration in {RLHF}},
  author={Leng, Jixuan and Huang, Chengsong and Zhu, Banghua and Huang, Jiaxin},
  booktitle={International Conference on Learning Representations (ICLR)}, year={2025}, note={arXiv:2410.09724}
}

@inproceedings{xu2024sayself,
  title={{SaySelf}: Teaching {LLMs} to Express Confidence with Self-Reflective Rationales},
  author={Xu, Tianyang and Wu, Shujin and Diao, Shizhe and Liu, Xiaoze and Wang, Xingyao and Chen, Yangyi and Gao, Jing},
  booktitle={Proceedings of the 2024 Conference on Empirical Methods in Natural Language Processing (EMNLP)},
  pages={5985--5998}, year={2024}, note={arXiv:2405.20974},
  doi={10.18653/v1/2024.emnlp-main.343}
}

@article{lin2022teaching,
  title={Teaching Models to Express Their Uncertainty in Words},
  author={Lin, Stephanie and Hilton, Jacob and Evans, Owain},
  journal={Transactions on Machine Learning Research (TMLR)}, year={2022}, note={arXiv:2205.14334}
}

@inproceedings{tian2023just,
  title={Just Ask for Calibration: Strategies for Eliciting Calibrated Confidence Scores from Language Models Fine-Tuned with Human Feedback},
  author={Tian, Katherine and Mitchell, Eric and Zhou, Allan and Sharma, Archit and Rafailov, Rafael and Yao, Huaxiu and Finn, Chelsea and Manning, Christopher D.},
  booktitle={Proceedings of the 2023 Conference on Empirical Methods in Natural Language Processing (EMNLP)},
  pages={5433--5442}, year={2023}, note={arXiv:2305.14975},
  doi={10.18653/v1/2023.emnlp-main.330}
}

@inproceedings{duan2024sar,
  title={Shifting Attention to Relevance: Towards the Predictive Uncertainty Quantification of Free-Form Large Language Models},
  author={Duan, Jinhao and Cheng, Hao and Wang, Shiqi and Zavalny, Alex and Wang, Chenan and Xu, Renjing and Kailkhura, Bhavya and Xu, Kaidi},
  booktitle={Proceedings of the 62nd Annual Meeting of the Association for Computational Linguistics (ACL)},
  pages={5050--5063}, year={2024}, note={arXiv:2307.01379},
  doi={10.18653/v1/2024.acl-long.276}
}

@inproceedings{wang2023selfconsistency,
  title={Self-Consistency Improves Chain of Thought Reasoning in Language Models},
  author={Wang, Xuezhi and Wei, Jason and Schuurmans, Dale and Le, Quoc and Chi, Ed and Narang, Sharan and Chowdhery, Aakanksha and Zhou, Denny},
  booktitle={International Conference on Learning Representations (ICLR)}, year={2023}, note={arXiv:2203.11171}
}

@inproceedings{chen2024bsdetector,
  title={Quantifying Uncertainty in Answers from any Language Model and Enhancing their Trustworthiness},
  author={Chen, Jiuhai and Mueller, Jonas},
  booktitle={Proceedings of the 62nd Annual Meeting of the Association for Computational Linguistics (ACL)},
  pages={5186--5200}, year={2024}, note={arXiv:2308.16175},
  doi={10.18653/v1/2024.acl-long.283}
}

@inproceedings{kuhn2023semantic,
  title={Semantic Uncertainty: Linguistic Invariances for Uncertainty Estimation in Natural Language Generation},
  author={Kuhn, Lorenz and Gal, Yarin and Farquhar, Sebastian},
  booktitle={International Conference on Learning Representations (ICLR)}, year={2023}, note={arXiv:2302.09664}
}

@article{farquhar2024detecting,
  title={Detecting Hallucinations in Large Language Models Using Semantic Entropy},
  author={Farquhar, Sebastian and Kossen, Jannik and Kuhn, Lorenz and Gal, Yarin},
  journal={Nature}, volume={630}, number={8017}, pages={625--630}, year={2024},
  doi={10.1038/s41586-024-07421-0}
}

@inproceedings{ulmer2024apricot,
  title={Calibrating Large Language Models Using Their Generations Only},
  author={Ulmer, Dennis and Gubri, Martin and Lee, Hwaran and Yun, Sangdoo and Oh, Seong Joon},
  booktitle={Proceedings of the 62nd Annual Meeting of the Association for Computational Linguistics (ACL)},
  pages={15440--15459}, year={2024}, note={arXiv:2403.05973},
  doi={10.18653/v1/2024.acl-long.824}
}

@inproceedings{shen2024thermometer,
  title={Thermometer: Towards Universal Calibration for Large Language Models},
  author={Shen, Maohao and Das, Subhro and Greenewald, Kristjan and Sattigeri, Prasanna and Wornell, Gregory and Ghosh, Soumya},
  booktitle={Proceedings of the 41st International Conference on Machine Learning (ICML)}, pages={44687--44711}, year={2024}, note={arXiv:2403.08819}
}

@inproceedings{mei2026reasoning,
  title={Reasoning about Uncertainty: Do Reasoning Models Know When They Don't Know?},
  author={Mei, Zhiting and Zhang, Christina and Yin, Tenny and Lidard, Justin and Shorinwa, Ola and Majumdar, Anirudha},
  booktitle={Findings of the Association for Computational Linguistics: EACL 2026}, pages={3408--3458}, year={2026}, note={arXiv:2506.18183},
  doi={10.18653/v1/2026.findings-eacl.178}
}

@inproceedings{halawi2024forecasting,
  title={Approaching Human-Level Forecasting with Language Models},
  author={Halawi, Danny and Zhang, Fred and Yueh-Han, Chen and Steinhardt, Jacob},
  booktitle={Advances in Neural Information Processing Systems (NeurIPS)}, year={2024}, note={arXiv:2402.18563}
}

@article{schoenegger2024silicon,
  title={Wisdom of the Silicon Crowd: {LLM} Ensemble Prediction Capabilities Rival Human Crowd Accuracy},
  author={Schoenegger, Philipp and Tuminauskaite, Indre and Park, Peter S. and Bastos, Rafael Valdece Sousa and Tetlock, Philip E.},
  journal={Science Advances}, volume={10}, number={45}, pages={eadp1528}, year={2024},
  doi={10.1126/sciadv.adp1528}, note={arXiv:2402.19379}
}

@inproceedings{naeini2015obtaining,
  title={Obtaining Well Calibrated Probabilities Using Bayesian Binning},
  author={Naeini, Mahdi Pakdaman and Cooper, Gregory F. and Hauskrecht, Milos},
  booktitle={Proceedings of the AAAI Conference on Artificial Intelligence}, pages={2901--2907}, year={2015}
}

@inproceedings{baniharouni2026rewarding,
  title={Rewarding Doubt: A Reinforcement Learning Approach to Calibrated Confidence Expression of Large Language Models},
  author={Bani-Harouni, David and Pellegrini, Chantal and Stangel, Paul and {\"O}zsoy, Ege and Zaripova, Kamilia and Navab, Nassir and Keicher, Matthias},
  booktitle={International Conference on Learning Representations (ICLR)}, year={2026}, note={arXiv:2503.02623}
}

@inproceedings{damani2026beyond,
  title={Beyond Binary Rewards: Training {LMs} to Reason About Their Uncertainty},
  author={Damani, Mehul and Puri, Isha and Slocum, Stewart and Shenfeld, Idan and Choshen, Leshem and Kim, Yoon and Andreas, Jacob},
  booktitle={International Conference on Learning Representations (ICLR)}, year={2026}, note={arXiv:2507.16806}
}

@article{puri2026reaching,
  title={Reaching Beyond the Mode: {RL} for Distributional Reasoning in Language Models},
  author={Puri, Isha and Damani, Mehul and Shenfeld, Idan and Ghassemi, Marzyeh and Andreas, Jacob and Kim, Yoon},
  journal={arXiv preprint arXiv:2603.24844}, year={2026}
}

@inproceedings{sprague2025cot,
  title={To {CoT} or not to {CoT}? {C}hain-of-thought helps mainly on math and symbolic reasoning},
  author={Sprague, Zayne and Yin, Fangcong and Rodriguez, Juan Diego and Jiang, Dongwei and Wadhwa, Manya and Singhal, Prasann and Zhao, Xinyu and Ye, Xi and Mahowald, Kyle and Durrett, Greg},
  booktitle={International Conference on Learning Representations (ICLR)}, year={2025}, note={arXiv:2409.12183}
}

@article{zhao2025dist2ill,
  title={{Dist2ill}: Distributional Distillation for One-Pass Uncertainty Estimation in Large Language Models},
  author={Zhao, Yicong and Tsang, King Yeung and Vejendla, Harshil and Shi, Haizhou and Li, Zhuohang and Hua, Zhigang and Xu, Qi and Zhang, Tunyu and Wang, Yi and Han, Ligong and Malin, Bradley A. and Wang, Hao},
  journal={arXiv preprint arXiv:2505.11731}, year={2025}
}

@article{lindley1956measure,
  title={On a Measure of the Information Provided by an Experiment},
  author={Lindley, Dennis V.},
  journal={The Annals of Mathematical Statistics}, volume={27}, number={4}, pages={986--1005}, year={1956}
}

@article{chaloner1995bayesian,
  title={Bayesian Experimental Design: A Review},
  author={Chaloner, Kathryn and Verdinelli, Isabella},
  journal={Statistical Science}, volume={10}, number={3}, pages={273--304}, year={1995}
}

@inproceedings{rainforth2018nesting,
  title={On Nesting Monte Carlo Estimators},
  author={Rainforth, Tom and Cornish, Robert and Yang, Hongseok and Warrington, Andrew and Wood, Frank},
  booktitle={Proceedings of the 35th International Conference on Machine Learning (ICML)}, pages={4267--4276}, year={2018}, note={arXiv:1709.06181}
}

@inproceedings{foster2019variational,
  title={Variational Bayesian Optimal Experimental Design},
  author={Foster, Adam and Jankowiak, Martin and Bingham, Eli and Horsfall, Paul and Teh, Yee Whye and Rainforth, Tom and Goodman, Noah},
  booktitle={Advances in Neural Information Processing Systems (NeurIPS)}, year={2019}, note={arXiv:1903.05480}
}

@inproceedings{poole2019variational,
  title={On Variational Bounds of Mutual Information},
  author={Poole, Ben and Ozair, Sherjil and van den Oord, Aaron and Alemi, Alexander A. and Tucker, George},
  booktitle={Proceedings of the 36th International Conference on Machine Learning (ICML)}, pages={5171--5180}, year={2019}, note={arXiv:1905.06922}
}

@inproceedings{foster2021deep,
  title={Deep Adaptive Design: Amortizing Sequential Bayesian Experimental Design},
  author={Foster, Adam and Ivanova, Desi R. and Malik, Ilyas and Rainforth, Tom},
  booktitle={Proceedings of the 38th International Conference on Machine Learning (ICML)}, pages={3384--3395}, year={2021}, note={arXiv:2103.02438}
}

@inproceedings{blau2022optimizing,
  title={Optimizing Sequential Experimental Design with Deep Reinforcement Learning},
  author={Blau, Tom and Bonilla, Edwin V. and Chades, Iadine and Dezfouli, Amir},
  booktitle={Proceedings of the 39th International Conference on Machine Learning (ICML)}, pages={2107--2128}, year={2022}, note={arXiv:2202.00821}
}

@inproceedings{mueller2022transformers,
  title={Transformers Can Do Bayesian Inference},
  author={M{\"u}ller, Samuel and Hollmann, Noah and Pineda Arango, Sebastian and Grabocka, Josif and Hutter, Frank},
  booktitle={International Conference on Learning Representations (ICLR)}, year={2022}, note={arXiv:2112.10510}
}

@inproceedings{hollmann2023tabpfn,
  title={{TabPFN}: A Transformer That Solves Small Tabular Classification Problems in a Second},
  author={Hollmann, Noah and M{\"u}ller, Samuel and Eggensperger, Katharina and Hutter, Frank},
  booktitle={International Conference on Learning Representations (ICLR)}, year={2023}, note={arXiv:2207.01848}
}

@article{hollmann2025tabpfn,
  title={Accurate predictions on small data with a tabular foundation model},
  author={Hollmann, Noah and M{\"u}ller, Samuel and Purucker, Lennart and Krishnakumar, Arjun and K{\"o}rfer, Max and Hoo, Shi Bin and Schirrmeister, Robin Tibor and Hutter, Frank},
  journal={Nature}, volume={637}, number={8045}, pages={319--326}, year={2025},
  doi={10.1038/s41586-024-08328-6}
}

@inproceedings{choudhury2026bedllm,
  title={{BED-LLM}: Intelligent Information Gathering with {LLMs} and Bayesian Experimental Design},
  author={Choudhury, Deepro and Williamson, Sinead and Goli{\'n}ski, Adam and Miao, Ning and Bickford Smith, Freddie and Kirchhof, Michael and Zhang, Yizhe and Rainforth, Tom},
  booktitle={International Conference on Learning Representations (ICLR)}, year={2026}, note={arXiv:2508.21184}
}

@inproceedings{hu2024uncertainty,
  title={Uncertainty of Thoughts: Uncertainty-Aware Planning Enhances Information Seeking in Large Language Models},
  author={Hu, Zhiyuan and Liu, Chumin and Feng, Xidong and Zhao, Yilun and Ng, See-Kiong and Luu, Anh Tuan and He, Junxian and Koh, Pang Wei and Hooi, Bryan},
  booktitle={Advances in Neural Information Processing Systems (NeurIPS)}, year={2024}, note={arXiv:2402.03271}
}

@inproceedings{mazzaccara2024learning,
  title={Learning to Ask Informative Questions: Enhancing LLMs with Preference Optimization and Expected Information Gain},
  author={Mazzaccara, Davide and Testoni, Alberto and Bernardi, Raffaella},
  booktitle={Findings of the Association for Computational Linguistics: EMNLP 2024}, pages={5064--5074}, year={2024}, note={arXiv:2406.17453},
  doi={10.18653/v1/2024.findings-emnlp.291}
}

@inproceedings{kobalczyk2025active,
  title={Active Task Disambiguation with {LLMs}},
  author={Kobalczyk, Katarzyna and Astorga, Nicol{\'a}s and Liu, Tennison and van der Schaar, Mihaela},
  booktitle={International Conference on Learning Representations (ICLR)}, year={2025}, note={arXiv:2502.04485}
}

@inproceedings{li2025eliciting,
  title={Eliciting Human Preferences with Language Models},
  author={Li, Belinda Z. and Tamkin, Alex and Goodman, Noah and Andreas, Jacob},
  booktitle={International Conference on Learning Representations (ICLR)}, year={2025}, note={arXiv:2310.11589}
}

@article{handa2024bayesian,
  title={Bayesian Preference Elicitation with Language Models},
  author={Handa, Kunal and Gal, Yarin and Pavlick, Ellie and Goodman, Noah and Andreas, Jacob and Tamkin, Alex and Li, Belinda Z.},
  journal={arXiv preprint arXiv:2403.05534}, year={2024}
}

@article{ramos2025bayesian,
  title={Bayesian Optimization of Catalysis with In-Context Learning},
  author={Ramos, Mayk Caldas and Michtavy, Shane S. and White, Andrew D. and Porosoff, Marc D.},
  journal={ACS Central Science}, volume={12}, number={5}, pages={599--615}, year={2026},
  doi={10.1021/acscentsci.5c02418}, note={arXiv:2304.05341}
}

@inproceedings{liu2024large,
  title={Large Language Models to Enhance Bayesian Optimization},
  author={Liu, Tennison and Astorga, Nicol{\'a}s and Seedat, Nabeel and van der Schaar, Mihaela},
  booktitle={International Conference on Learning Representations (ICLR)}, year={2024}, note={arXiv:2402.03921}
}

@inproceedings{huang2024sequential,
  title={Sequential Optimal Experimental Design of Perturbation Screens Guided by Multi-modal Priors},
  author={Huang, Kexin and Lopez, Romain and H{\"u}tter, Jan-Christian and Kudo, Takamasa and Rios, Antonio and Regev, Aviv},
  booktitle={Research in Computational Molecular Biology (RECOMB)}, series={Lecture Notes in Computer Science},
  pages={17--37}, year={2024}, doi={10.1007/978-1-0716-3989-4_2}
}

@inproceedings{hao2025perturboagent,
  title={{PerTurboAgent}: An {LLM}-based Agent for Designing Iterative {Perturb-Seq} Experiments},
  author={Hao, Minsheng and Lee, Yongju and Wang, Hanchen and Scalia, Gabriele and Regev, Aviv},
  booktitle={Proceedings of Machine Learning in Computational Biology (MLCB)},
  series={PMLR}, volume={311}, pages={44--64}, year={2025}
}

@inproceedings{grand2026shoot,
  title={Shoot First, Ask Questions Later? Building Rational Agents that Explore and Act Like People},
  author={Grand, Gabriel and Pepe, Valerio and Tenenbaum, Joshua B. and Andreas, Jacob},
  booktitle={International Conference on Learning Representations (ICLR)}, year={2026},
}

@inproceedings{li2024mediq,
  title={{MediQ}: Question-Asking LLMs and a Benchmark for Reliable Interactive Clinical Reasoning},
  author={Li, Shuyue Stella and Balachandran, Vidhisha and Feng, Shangbin and Ilgen, Jonathan S. and Pierson, Emma and Koh, Pang Wei and Tsvetkov, Yulia},
  booktitle={Advances in Neural Information Processing Systems (NeurIPS)}, year={2024}, note={arXiv:2406.00922}
}

@article{kosmos,
  title={Kosmos: An {AI} Scientist for Autonomous Discovery},
  author={Mitchener, Ludovico and Yiu, Angela and Chang, Benjamin and Bourdenx, Mathieu and Nadolski, Tyler and Sulovari, Arvis and Landsness, Eric C. and Barabasi, Daniel L. and Narayanan, Siddharth and Evans, Nicky and Reddy, Shriya and Foiani, Martha and Kamal, Aizad and Shriver, Leah P. and Cao, Fang and Wassie, Asmamaw T. and Laurent, Jon M. and Melville-Green, Edwin and Caldas, Mayk and Bou, Albert and Roberts, Kaleigh F. and Zagorac, Sladjana and Orr, Timothy C. and Orr, Miranda E. and Zwezdaryk, Kevin J. and Ghareeb, Ali E. and McCoy, Laurie and Gomes, Bruna and Ashley, Euan A. and Duff, Karen E. and Buonassisi, Tonio and Rainforth, Tom and Bateman, Randall J. and Skarlinski, Michael and Rodriques, Samuel G. and Hinks, Michaela M. and White, Andrew D.},
  journal={arXiv preprint arXiv:2511.02824}, year={2025}
}

@book{good1950probability,
  title={Probability and the Weighing of Evidence},
  author={Good, Irving John},
  year={1950},
  publisher={Charles Griffin}
}

@article{mantel1959statistical,
  title={Statistical aspects of the analysis of data from retrospective studies of disease},
  author={Mantel, Nathan and Haenszel, William},
  journal={Journal of the National Cancer Institute},
  volume={22},
  number={4},
  pages={719--748},
  year={1959}
}

@article{coibion2015information,
  title={Information Rigidity and the Expectations Formation Process: A Simple Framework and New Facts},
  author={Coibion, Olivier and Gorodnichenko, Yuriy},
  journal={American Economic Review},
  volume={105},
  number={8},
  pages={2644--2678},
  year={2015}
}

@article{bordalo2020overreaction,
  title={Overreaction in Macroeconomic Expectations},
  author={Bordalo, Pedro and Gennaioli, Nicola and Ma, Yueran and Shleifer, Andrei},
  journal={American Economic Review},
  volume={110},
  number={9},
  pages={2748--2782},
  year={2020}
}

@article{hawthorne2016belief,
  title={Belief is weak},
  author={Hawthorne, John and Rothschild, Daniel and Spectre, Levi},
  journal={Philosophical Studies},
  volume={173},
  number={5},
  pages={1393--1404},
  year={2016},
  publisher={Springer}
}

@article{shannon_mathematical_1948,
 author = {Shannon, C. E.},
 doi = {10.1002/j.1538-7305.1948.tb01338.x},
 issn = {0005-8580},
 journal = {The Bell System Technical Journal},
 number = {3},
 pages = {379--423},
 title = {A mathematical theory of communication},
 url = {https://ieeexplore.ieee.org/document/6773024},
 urldate = {2025-08-12},
 volume = {27},
 year = {1948}
}

@article{mackay1992information,
 author = {MacKay, David J. C.},
 doi = {10.1162/neco.1992.4.4.590},
 journal = {Neural Computation},
 number = {4},
 pages = {590--604},
 title = {Information-based objective functions for active data selection},
 volume = {4},
 year = {1992}
}

@inproceedings{wadden2020scifact,
  title={Fact or Fiction: Verifying Scientific Claims},
  author={Wadden, David and Lin, Shanchuan and Lo, Kyle and Wang, Lucy Lu and van Zuylen, Madeleine and Cohan, Arman and Hajishirzi, Hannaneh},
  booktitle={Proceedings of the 2020 Conference on Empirical Methods in Natural Language Processing (EMNLP)}, pages={7534--7550}, year={2020}, note={arXiv:2004.14974},
  doi={10.18653/v1/2020.emnlp-main.609}
}

@inproceedings{wadden2022scifactopen,
  title={{SciFact-Open}: Towards open-domain scientific claim verification},
  author={Wadden, David and Lo, Kyle and Kuehl, Bailey and Cohan, Arman and Beltagy, Iz and Wang, Lucy Lu and Hajishirzi, Hannaneh},
  booktitle={Findings of the Association for Computational Linguistics: EMNLP 2022}, pages={4719--4734}, year={2022}, note={arXiv:2210.13777},
  doi={10.18653/v1/2022.findings-emnlp.347}
}

@article{laurent2024labbench,
  title={{LAB-Bench}: Measuring Capabilities of Language Models for Biology Research},
  author={Laurent, Jon M. and Janizek, Joseph D. and Ruzo, Michael and Hinks, Michaela M. and Hammerling, Michael J. and Narayanan, Siddharth and Ponnapati, Manvitha and White, Andrew D. and Rodriques, Samuel G.},
  journal={arXiv preprint arXiv:2407.10362}, year={2024}
}

@article{mitchener2025bixbench,
  title={{BixBench}: a Comprehensive Benchmark for {LLM}-based Agents in Computational Biology},
  author={Mitchener, Ludovico and Laurent, Jon M. and Andonian, Alex and Tenmann, Benjamin and Narayanan, Siddharth and Wellawatte, Geemi P. and White, Andrew and Sani, Lorenzo and Rodriques, Samuel G.},
  journal={arXiv preprint arXiv:2503.00096}, year={2025}
}

@article{skarlinski2024paperqa2,
  title={Language agents achieve superhuman synthesis of scientific knowledge},
  author={Skarlinski, Michael D. and Cox, Sam and Laurent, Jon M. and Braza, James D. and Hinks, Michaela and Hammerling, Michael J. and Ponnapati, Manvitha and Rodriques, Samuel G. and White, Andrew D.},
  journal={arXiv preprint arXiv:2409.13740}, year={2024}
}

@article{laurent2026labbench2,
  title={{LABBench2}: An Improved Benchmark for {AI} Systems Performing Biology Research},
  author={Laurent, Jon M. and Bou, Albert and Pieler, Michael and Igoe, Conor and Andonian, Alex and Narayanan, Siddharth and Braza, James and Vassopoulos, Alexandros Sanchez and Steenwyk, Jacob L. and Lash, Blake and White, Andrew D. and Rodriques, Samuel G.},
  journal={arXiv preprint arXiv:2604.09554}, year={2026}
}

@article{koch2026bixbench3,
  title={{BixBench3}: Benchmarking {AI} agents on research-study-scale computational biology tasks},
  author={Koch, Zane and Wassie, Asmamaw T. and Valdes-Aleman, Javier and Lee, Jason and Hinks, Michaela M. and Rodriques, Samuel G. and White, Andrew D. and Laurent, Jon M.},
  journal={arXiv preprint arXiv:2608.25286}, year={2026}
}

@misc{almeida2026jev,
  title={Introducing {System One Models} \& {Jev}},
  author={Almeida, Diogo},
  howpublished={TypeSafe AI, \url{https://typesafe.ai/blog/introducing-system-one-models-and-jev}}, year={2026}
}

@inproceedings{rein2023gpqa,
  title={{GPQA}: A Graduate-Level {Google}-Proof {Q\&A} Benchmark},
  author={Rein, David and Hou, Betty Li and Stickland, Asa Cooper and Petty, Jackson and Pang, Richard Yuanzhe and Dirani, Julien and Michael, Julian and Bowman, Samuel R.},
  booktitle={Conference on Language Modeling (COLM)}, year={2024}, note={arXiv:2311.12022}
}

@inproceedings{tian2024scicode,
  title={{SciCode}: A Research Coding Benchmark Curated by Scientists},
  author={Tian, Minyang and Gao, Luyu and Zhang, Shizhuo Dylan and Chen, Xinan and Fan, Cunwei and Guo, Xuefei and Haas, Roland and Ji, Pan and Krongchon, Kittithat and Li, Yao and others},
  booktitle={Advances in Neural Information Processing Systems (NeurIPS), Datasets and Benchmarks Track}, year={2024}, note={arXiv:2407.13168}
}

@inproceedings{cui2025curie,
  title={{CURIE}: Evaluating {LLMs} On Multitask Scientific Long Context Understanding and Reasoning},
  author={Cui, Hao and Shamsi, Zahra and Cheon, Gowoon and Ma, Xuejian and Li, Shutong and Tikhanovskaya, Maria and Norgaard, Peter and Mudur, Nayantara and Plomecka, Martyna and Raccuglia, Paul and others},
  booktitle={International Conference on Learning Representations (ICLR)}, year={2025}, note={arXiv:2503.13517}
}

@inproceedings{wang2022scienceworld,
  title={{ScienceWorld}: Is your Agent Smarter than a 5th Grader?},
  author={Wang, Ruoyao and Jansen, Peter and C{\^o}t{\'e}, Marc-Alexandre and Ammanabrolu, Prithviraj},
  booktitle={Proceedings of the 2022 Conference on Empirical Methods in Natural Language Processing (EMNLP)}, pages={11279--11298}, year={2022}, note={arXiv:2203.07540},
  doi={10.18653/v1/2022.emnlp-main.775}
}

@inproceedings{jansen2024discoveryworld,
  title={{DiscoveryWorld}: A Virtual Environment for Developing and Evaluating Automated Scientific Discovery Agents},
  author={Jansen, Peter and C{\^o}t{\'e}, Marc-Alexandre and Khot, Tushar and Bransom, Erin and Mishra, Bhavana Dalvi and Majumder, Bodhisattwa Prasad and Tafjord, Oyvind and Clark, Peter},
  booktitle={Advances in Neural Information Processing Systems (NeurIPS), Datasets and Benchmarks Track}, year={2024}, note={arXiv:2406.06769}
}

@inproceedings{majumder2024discoverybench,
  title={{DiscoveryBench}: Towards Data-Driven Discovery with Large Language Models},
  author={Majumder, Bodhisattwa Prasad and Surana, Harshit and Agarwal, Dhruv and Mishra, Bhavana Dalvi and Meena, Abhijeetsingh and Prakhar, Aryan and Vora, Tirth and Khot, Tushar and Sabharwal, Ashish and Clark, Peter},
  booktitle={International Conference on Learning Representations (ICLR)}, year={2025}, note={arXiv:2407.01725}
}

@inproceedings{jansen2025matteroffact,
  title={Matter-of-Fact: A Benchmark for Verifying the Feasibility of Literature-Supported Claims in Materials Science},
  author={Jansen, Peter and Hassan, Samiah and Wang, Ruoyao},
  booktitle={Proceedings of the 2025 Conference on Empirical Methods in Natural Language Processing (EMNLP)}, pages={4090--4102}, year={2025}, note={arXiv:2506.04410},
  doi={10.18653/v1/2025.emnlp-main.203}
}

@inproceedings{chen2024scienceagentbench,
  title={{ScienceAgentBench}: Toward Rigorous Assessment of Language Agents for Data-Driven Scientific Discovery},
  author={Chen, Ziru and Chen, Shijie and Ning, Yuting and Zhang, Qianheng and Wang, Boshi and Yu, Botao and Li, Yifei and Liao, Zeyi and Wei, Chen and Lu, Zitong and Dey, Vishal and Xue, Mingyi and Baker, Frazier N. and Burns, Benjamin and Adu-Ampratwum, Daniel and Huang, Xuhui and Ning, Xia and Gao, Song and Su, Yu and Sun, Huan},
  booktitle={International Conference on Learning Representations (ICLR)}, year={2025}, note={arXiv:2410.05080}
}

@inproceedings{wang2023primacy,
  title={Primacy Effect of {ChatGPT}},
  author={Wang, Yiwei and Cai, Yujun and Chen, Muhao and Liang, Yuxuan and Hooi, Bryan},
  booktitle={Proceedings of the 2023 Conference on Empirical Methods in Natural Language Processing (EMNLP)}, pages={108--115}, year={2023}, note={arXiv:2310.13206},
  doi={10.18653/v1/2023.emnlp-main.8}
}

@article{liu2023lost,
  title={Lost in the Middle: How Language Models Use Long Contexts},
  author={Liu, Nelson F. and Lin, Kevin and Hewitt, John and Paranjape, Ashwin and Bevilacqua, Michele and Petroni, Fabio and Liang, Percy},
  journal={Transactions of the Association for Computational Linguistics (TACL)}, volume={12}, pages={157--173}, year={2024}, note={arXiv:2307.03172},
  doi={10.1162/tacl_a_00638}
}

@article{lou2024anchoring,
  title={Anchoring Bias in Large Language Models: An Experimental Study},
  author={Lou, Jiaxu and Sun, Yifan},
  journal={arXiv preprint arXiv:2412.06593}, year={2024}
}

@inproceedings{gupta2025coinflips,
  title={Enough Coin Flips Can Make {LLMs} Act {Bayesian}},
  author={Gupta, Ritwik and Corona, Rodolfo and Ge, Jiaxin and Wang, Eric and Klein, Dan and Darrell, Trevor and Chan, David M.},
  booktitle={Proceedings of the 63rd Annual Meeting of the Association for Computational Linguistics (ACL)}, pages={7634--7655}, year={2025}, note={arXiv:2503.04722},
  doi={10.18653/v1/2025.acl-long.377}
}

@inproceedings{kim2025evidence,
  title={From Evidence to Belief: A {Bayesian} Epistemology Approach to Language Models},
  author={Kim, Minsu and Kim, Sangryul and Thorne, James},
  booktitle={Proceedings of the 2025 Conference of the Nations of the Americas Chapter of the Association for Computational Linguistics (NAACL)}, pages={10578--10611}, year={2025}, note={arXiv:2504.19622},
  doi={10.18653/v1/2025.naacl-long.531}
}

@inproceedings{raimondi2025primacy,
  title={Exploiting Primacy Effect To Improve Large Language Models},
  author={Raimondi, Bianca and Gabbrielli, Maurizio},
  booktitle={Proceedings of the International Conference on Recent Advances in Natural Language Processing (RANLP)}, pages={989--997}, year={2025}, note={arXiv:2507.13949}
}

@article{zhang2025discounted,
  title={Large Language Models as Discounted {Bayesian} Filters},
  author={Zhang, Jensen and Yang, Jing and Wang, Keze},
  journal={arXiv preprint arXiv:2512.18489}, year={2025}
}

@article{sumita2024cognitive,
  title={Cognitive Biases in Large Language Models: A Survey and Mitigation Experiments},
  author={Sumita, Yasuaki and Takeuchi, Koh and Kashima, Hisashi},
  journal={arXiv preprint arXiv:2412.00323}, year={2024}
}

@article{kass1995bayes,
  title={{Bayes} Factors},
  author={Kass, Robert E. and Raftery, Adrian E.},
  journal={Journal of the American Statistical Association}, volume={90}, number={430}, pages={773--795}, year={1995},
  doi={10.1080/01621459.1995.10476572}
}

@article{good1979turing,
  title={Studies in the History of Probability and Statistics. {XXXVII} {A. M. Turing}'s statistical work in {World War II}},
  author={Good, I. J.},
  journal={Biometrika}, volume={66}, number={2}, pages={393--396}, year={1979},
  doi={10.1093/biomet/66.2.393}
}

@article{roohani2023gears,
  title={Predicting transcriptional outcomes of novel multigene perturbations with {GEARS}},
  author={Roohani, Yusuf and Huang, Kexin and Leskovec, Jure},
  journal={Nature Biotechnology}, volume={42}, number={6}, pages={927--935}, year={2024},
  doi={10.1038/s41587-023-01905-6}
}

@article{cui2024scgpt,
  title={{scGPT}: toward building a foundation model for single-cell multi-omics using generative {AI}},
  author={Cui, Haotian and Wang, Chloe and Maan, Hassaan and Pang, Kuan and Luo, Fengning and Duan, Nan and Wang, Bo},
  journal={Nature Methods}, volume={21}, number={8}, pages={1470--1480}, year={2024},
  doi={10.1038/s41592-024-02201-0}
}

@article{theodoris2023geneformer,
  title={Transfer learning enables predictions in network biology},
  author={Theodoris, Christina V. and Xiao, Ling and Chopra, Anant and Chaffin, Mark D. and Al Sayed, Zeina R. and Hill, Matthew C. and Mantineo, Helene and Brydon, Elizabeth M. and Zeng, Zexian and Liu, X. Shirley and Ellinor, Patrick T.},
  journal={Nature}, volume={618}, number={7965}, pages={616--624}, year={2023},
  doi={10.1038/s41586-023-06139-9}
}

@article{ahlmann2025linear,
  title={Deep-learning-based gene perturbation effect prediction does not yet outperform simple linear baselines},
  author={Ahlmann-Eltze, Constantin and Huber, Wolfgang and Anders, Simon},
  journal={Nature Methods}, volume={22}, number={8}, pages={1657--1661}, year={2025},
  doi={10.1038/s41592-025-02772-6}
}

@article{wei2026benchmarking,
  title={Benchmarking algorithms for generalizable single-cell perturbation response prediction},
  author={Wei, Zhiting and Wang, Yiheng and Gao, Yicheng and Wang, Shuguang and Li, Ping and Si, Duanmiao and Gao, Yuli and Wu, Siqi and Li, Danlu and Dong, Kejing and Yang, Xingbo and Tang, Chen and Fu, Shaliu and Chen, Xiaohan and Li, Wannian and You, Yuzhou and Zhang, Chen and Liang, Aibin and Chuai, Guohui and Liu, Qi},
  journal={Nature Methods}, volume={23}, number={2}, pages={451--464}, year={2026},
  doi={10.1038/s41592-025-02980-0}
}

@article{bunne2024virtualcell,
  title={How to build the virtual cell with artificial intelligence: Priorities and opportunities},
  author={Bunne, Charlotte and Roohani, Yusuf and Rosen, Yanay and Gupta, Ankit and Zhang, Xikun and Roed, Marcel and Alexandrov, Theo and AlQuraishi, Mohammed and Brennan, Patricia and Burkhardt, Daniel B. and others},
  journal={Cell}, volume={187}, number={25}, pages={7045--7063}, year={2024},
  doi={10.1016/j.cell.2024.11.015}
}

@article{roohani2025vcc,
  title={Virtual Cell Challenge: Toward a {Turing} test for the virtual cell},
  author={Roohani, Yusuf H. and Hua, Tony J. and Tung, Po-Yuan and Bounds, Lexi R. and Yu, Feiqiao B. and Dobin, Alexander and Teyssier, Noam and Adduri, Abhinav and Woodrow, Alden and Plosky, Brian S. and others},
  journal={Cell}, volume={188}, number={13}, pages={3370--3374}, year={2025},
  doi={10.1016/j.cell.2025.06.008}
}

@article{adduri2025state,
  title={Predicting cellular responses to perturbation across diverse contexts with {State}},
  author={Adduri, Abhinav K. and Gautam, Dhruv and Bevilacqua, Beatrice and Imran, Alishba and Shah, Rohan and Naghipourfar, Mohsen and Teyssier, Noam and Ilango, Rajesh and Nagaraj, Sanjay and Dong, Mingze and others},
  journal={bioRxiv}, year={2025},
  doi={10.1101/2025.06.26.661135}
}

@article{boiko2023coscientist,
  title={Autonomous chemical research with large language models},
  author={Boiko, Daniil A. and MacKnight, Robert and Kline, Ben and Gomes, Gabe},
  journal={Nature}, volume={624}, number={7992}, pages={570--578}, year={2023},
  doi={10.1038/s41586-023-06792-0}
}

@article{lu2024aiscientist,
  title={The {AI} Scientist: Towards Fully Automated Open-Ended Scientific Discovery},
  author={Lu, Chris and Lu, Cong and Lange, Robert Tjarko and Foerster, Jakob and Clune, Jeff and Ha, David},
  journal={arXiv preprint arXiv:2408.06292}, year={2024}
}

@article{yamada2025aiscientistv2,
  title={The {AI} Scientist-v2: Workshop-Level Automated Scientific Discovery via Agentic Tree Search},
  author={Yamada, Yutaro and Lange, Robert Tjarko and Lu, Cong and Hu, Shengran and Lu, Chris and Foerster, Jakob and Clune, Jeff and Ha, David},
  journal={arXiv preprint arXiv:2504.08066}, year={2025}
}

@article{swanson2025virtuallab,
  title={The Virtual Lab of {AI} agents designs new {SARS-CoV-2} nanobodies},
  author={Swanson, Kyle and Wu, Wesley and Bulaong, Nash L. and Pak, John E. and Zou, James},
  journal={Nature}, volume={646}, number={8085}, pages={716--723}, year={2025},
  doi={10.1038/s41586-025-09442-9}
}

@article{ghareeb2026robin,
  title={A multi-agent system for automating scientific discovery},
  author={Ghareeb, Ali E. and Chang, Benjamin and Mitchener, Ludovico and Yiu, Angela and Szostkiewicz, Caralyn J. and Shved, Dmytro and Gyimesi, Gavin J. and Laurent, Jon M. and Wright, Samantha M. and Razzak, Muhammed T. and White, Andrew D. and Finnemann, Silvia C. and Hinks, Michaela M. and Rodriques, Samuel G.},
  journal={Nature}, volume={655}, number={8122}, pages={497--505}, year={2026},
  doi={10.1038/s41586-026-10652-y}
}

@article{schulz1995empirical,
  title={Empirical evidence of bias: dimensions of methodological quality associated with estimates of treatment effects in controlled trials},
  author={Schulz, Kenneth F and Chalmers, Iain and Hayes, Richard J and Altman, Douglas G},
  journal={JAMA},
  volume={273},
  number={5},
  pages={408--412},
  year={1995},
  publisher={American Medical Association}
}

@article{gottweis2026coscientist,
  title   = {Accelerating Scientific Discovery with Co-Scientist},
  author  = {Gottweis, Juraj and Weng, Wei-Hung and Daryin, Alexander
             and Tu, Tao and Sirkovic, Petar and others},
  journal = {Nature},
  volume  = {655},
  number  = {8122},
  pages   = {487--496},
  year    = {2026},
  doi     = {10.1038/s41586-026-10644-y}
}

@article{ioannidis2005most,
  title={Why most published research findings are false},
  author={Ioannidis, John PA},
  journal={PLoS medicine},
  volume={2},
  number={8},
  pages={e124},
  year={2005},
  publisher={Public Library of Science}
}

@article{chan2004empirical,
  title={Empirical evidence for selective reporting of outcomes in randomized trials: comparison of protocols to published articles},
  author={Chan, An-Wen and Hr{\'o}bjartsson, Asbj{\o}rn and Haahr, Mette T and G{\o}tzsche, Peter C and Altman, Douglas G},
  journal={Jama},
  volume={291},
  number={20},
  pages={2457--2465},
  year={2004},
  publisher={American Medical Association}
}

@article{sterne2019rob,
  title={RoB 2: a revised tool for assessing risk of bias in randomised trials},
  author={Sterne, Jonathan AC and Savovi{\'c}, Jelena and Page, Matthew J and Elbers, Roy G and Blencowe, Natalie S and Boutron, Isabelle and Cates, Christopher J and Cheng, Hung-Yuan and Corbett, Mark S and Eldridge, Sandra M and others},
  journal={bmj},
  volume={366},
  year={2019},
  publisher={British Medical Journal Publishing Group}
}

@article{lundh2017industry,
  title={Industry sponsorship and research outcome},
  author={Lundh, Andreas and Lexchin, Joel and Mintzes, Barbara and Schroll, Jeppe B and Bero, Lisa},
  journal={Cochrane Database of Systematic Reviews},
  volume={2017},
  number={2},
  pages={MR000033},
  year={2017},
  doi={10.1002/14651858.MR000033.pub3}
}

@book{kuhn1997structure,
  title={The structure of scientific revolutions},
  author={Kuhn, Thomas S},
  volume={962},
  year={1997},
  publisher={University of Chicago press Chicago}
}
\end{document}